\documentclass[journal,10pt]{IEEEtran}
\usepackage{epigraph}
\usepackage{bbm}
\usepackage{amsmath}
\usepackage{acronym}  
\usepackage[dvips]{color}
\usepackage{epsf}
\usepackage{times}
\usepackage{epsfig}
\usepackage{notoccite}
\usepackage{graphicx}
\usepackage{epstopdf}
\usepackage{pstricks}
\usepackage{amssymb}
\usepackage{amsxtra}
\usepackage{here}
\usepackage{rawfonts}
\usepackage{float}
\usepackage{times}
\usepackage{url}
\usepackage{cite}
\usepackage{caption}
\usepackage{subcaption}
\usepackage{algorithm}
\usepackage{algpseudocode}
\usepackage{blindtext}
\usepackage{enumitem}
\usepackage{xcolor,cite,etoolbox}
\usepackage{relsize}
\usepackage{lipsum}
\usepackage{graphicx}
\usepackage{tabularx}
\usepackage{amsthm}

\usepackage{xparse}
\usepackage{array}
\newcolumntype{P}[1]{>{\centering\arraybackslash}p{#1}}
\usepackage{mleftright}

\usepackage{mathtools}

\usepackage{graphics}
\usepackage{physics}
\usepackage{amssymb}
\usepackage{siunitx}
\include{newcommands}
\usepackage{multicol}

\usepackage[nomain,acronym,shortcuts]{glossaries}
\makeglossaries
\newcommand*{\acro}[3][]{\newacronym[#1]{#2}{#2}{#3}}

\acro{QoNE}{quality-of-network experience}
\acro{D2D}{device-to-device}
\acro{ToM}{theory of mind}
\acro{JEPA}{joint embedding and predictive architecture}
\acro{NN}{neural network}
\acro{KPI}{key performance indicator}
\acro{KL}{Kullback-Leibler}
\acro{NeSy}{neuro-symbolic}

\acro{HDC}{hyper-dimensional computing}
\acro{SCM}{structural causal model}
\acro{CRL}{causal representation learning}
\acro{SIR}{signal-to-interference-ratio}
\acro{SINR}{signal-to-interference-plus-noise-ratio}
\acro{PCP}{Poisson cluster process}
\acro{CoMP}{coordinated multi-point}
\acro{BS}{base station} 
\acro{MD-CoMP}{macrodiversity CoMP transmission}
\acro{MAC}{medium-access-control}
\acro{JT-CoMP}{joint transmission CoMP}
\acro{CoMP-JT}{coordinated multipoint joint transmission}
\acro{SBS}{small base station}
\acro{MDSD}{multiple devices to a single device}
\acro{MDS}{maximum distance separable}
\acro{SCN}{small cell network}
\acro{PPP}{Poisson point process}
\acro{TCP}{Thomas cluster process}
\acro{CSI}{channel state information}
\acro{PDF}{probability distribution function}
\acro{PMF}{probability mass function}
\acro{RV}{random variable}
\acro{i.i.d.}{independently and identically distributed}
\acro{MBMS}{multimedia broadcasting multicasting service}
\acro{EE}{energy efficiency}
\acro{HCP}{hard-core placement}
\acro{CCDF}{complementary cumulative distribution function}
\acro{CDF}{cumulative distribution function}
\acro{PC}{probabilistic caching}
\acro{RC}{random caching}
\acro{CPF}{caching popular files} 
\acro{PGFL}{probability generating functional}
\acro{KKT}{Karush-Kuhn-Tucker}
\acro{PGF}{point generating function}
\acro{SCA}{successive convex approximation}
\acro{FHD}{full-high-definition}
\acro{UHD}{ultra-high-definition}
\acro{VR}{virtual reality}
\acro{AR}{augmented reality}
\acro{5G}{fifth-generation}
\acro{QoS}{quality-of-service}
\acro{QoE}{quality-of-experience}
\acro{IoT}{Internet of Things}
\acro{MHCPP}{Matern hardcore point process}
\acro{LoS}{line-of-sight}
\acro{NLoS}{non-line-of-sight}
\acro{PSD}{power spectral density}
\acro{MEC}{mobile edge computing}
\acro{E2E}{end-to-end}
\acro{THz}{terahertz}
\acro{CLT}{central limit theorem}
\acro{HQ}{High Quality}
\acro{eMBB}{enhanced mobile broadband}
\acro{URLLC}{ultra reliable low latency communications}
\acro{mmWave}{millimeter wave}
\acro{EVT}{extreme value theory}
\acro{GEV}{generalized extreme value}
\acro{LIS}{large intelligent surface}
\acro{RIS}{reconfigurable intelligent surface}
\acro{RF}{radio frequency}
\acro{UE}{user equipment}
\acro{MIMO}{multiple-input multiple-output}
\acro{EVaR}{entropic value-at-risk}
\acro{DNN}{deep neural network}
\acro{MDP}{Markov decision process}
\acro{RL}{reinforcement learning}
\acro{RNN}{recurrent neural network}
\acro{ANN}{artificial neural networks}
\acro{LSTM}{long short-term memory}
\acro{ReLu}{rectified linear unit}
\acro{VaR}{value-at-risk}
\acro{SNR}{signal-to-noise ratio}
\acro{AoSA}{array of subarray}
\acro{XR}{extended reality}
\acro{AoA}{angle of arrival}
\acro{ULA}{uniform linear array}
\acro{AoD}{angle of departure}
\acro{EM}{electromagnetic}
\acro{HRLLC}{s high-rate and high-reliability low latency communications}
\acro{6DoF}{six degrees of freedom}
\acro{MR}{mixed reality}
\acro{PAPR}{peak to average power ratio}
\acro{OFDM}{orthogonal frequency-division multiplexing}
\acro{OFDMA}{orthogonal frequency-division multiple access}
\acro{SC-FDM}{single carrier frequency-division multiplexing}
\acro{ToA}{time of arrival}
\acro{MUSIC}{multiple signal classification}
\acro{IoE}{Internet of Everything}
\acro{DT}{digital twin}
\acro{PT}{physical twin}
\acro{CT}{cyber twin}
\acro{DRL}{deep reinforcement learning}
\acro{FL}{federated learning}
\acro{DL}{deep learning}
\acro{CRAS}{connected robotics and autonomous system}
\acro{CL}{continual learning}
\acro{MSE}{mean squared error}
\acro{EWC}{elastic weight consolidation}
\acro{ML}{machine learning}
\acro{GD}{gradient descent}
\acro{MLP}{multi layer perceptron}
\acro{TL}{transfer learning}
\acro{AI}{artificial intelligence}
\acro{NFT}{non fungible token}
\acro{H2A}{human-to-avatar}
\acro{A2A}{avatar-to-avatar}
\acro{UAV}{unmanned aerial vehicle}
\acro{NTN}{non-terrestrial networks}
\acro{CIS}{connected intelligence system}
\acro{QoVE}{quality-of-virtual experience}
\acro{OOD}{out-of-distribution}
\acro{RAN}{radio access network}
\acro{PVD}{physical-virtual-digital}
\acro{Tx}{transmitter}
\acro{Rx}{receiver}
\acro{RRM}{radio resource management}
\acro{AGI}{artificial general intelligence}
\acro{LLM}{large language model}
\acro{FM}{foundation model}
\acro{QoDE}{quality-of-digital experience}
\acro{QoPE}{quality-of-physical experience}
\acro{POMDP}{partially observable Markov decision process}
\acro{HD}{hyper-dimensional}
\acro{IIT}{integrated information theory}
\acro{DAG}{directed acyclic graph}
\acro{GNN}{graph neural network}
\acro{XAI}{explainable learning}

\usepackage{datetime}
\usepackage{amssymb}
\usepackage{subcaption}
\usepackage{verbatim}

\newtheorem{definition}{\bf Definition}

\IEEEoverridecommandlockouts

\newcommand{\xdownarrow}[1]{%
  {\left\downarrow\vbox to #1{}\right.\kern-\nulldelimiterspace}
}

\newcommand{\mU}{{\mathcal U}}
\newcommand{\mV}{{\mathcal V}}
\newcommand{\mF}{{\mathcal F}}

\newcommand{\mX}{{\mathcal X}}

\newcommand{\mS}{{\mathcal S}}
\newcommand{\mA}{{\mathcal A}}
\newcommand{\mL}{{\mathcal L}}
\newcommand{\mwL}{{\widehat{\mathcal L}}}
\newcommand{\mH}{{\mathcal H}}
\newcommand{\mW}{{\mathcal W}}

\newcommand{\mG}{{\mathcal G}}
\newcommand{\bmU}{{\boldsymbol U}}

\DeclareMathOperator*{\argmax}{arg\,max}
\DeclareMathOperator*{\argmin}{arg\,min}

\newcommand{\bmy}{{\boldsymbol y}}
\newcommand{\bmV}{{\boldsymbol V}}
\newcommand{\bmQ}{{\boldsymbol Q}}
\newcommand{\bmK}{{\boldsymbol K}}
\newcommand{\bmA}{{\boldsymbol A}}
\newcommand{\bmx}{{\boldsymbol x}}
\newcommand{\bmh}{{\boldsymbol h}}
\newcommand{\bmb}{{\boldsymbol b}}
\newcommand{\bmM}{{\boldsymbol M}}
\newcommand{\bmz}{{\boldsymbol z}}
\newcommand{\bmzh}{{\widehat \bmz}}

\newcommand{\bmW}{{\boldsymbol W}}
\newcommand{\bmu}{{\boldsymbol u}}
\newcommand{\bmv}{{\boldsymbol v}}

\newcommand{\bms}{{\boldsymbol s}}
\newcommand{\bma}{{\boldsymbol a}}
\newcommand{\mbI}{{\mathbb I}}
\newcommand{\beq}{{\begin{equation}}}
\newcommand{\eeq}{{\end{equation}}}

\begin{document}
\title{Artificial General Intelligence (AGI)-Native Wireless Systems: A Journey Beyond 6G
\thanks{ W. Saad, O. Hashash, and C. K. Thomas are with Wireless@VT, Bradley Department of Electrical and Computer Engineering, Virginia Tech, Arlington, VA, USA.
E-mails: \protect{walids@vt.edu, omarnh@vt.edu, christokt@vt.edu}.}
\thanks{C. Chaccour is with Ericsson, Inc., Plano, Texas, USA. Email: \protect{christina.chaccour@ericsson.com}.}
\thanks{M. Debbah is with is with Khalifa University of Science and Technology, Abu Dhabi 127788, United Arab Emirates, and also with the CentraleSupelec, University ParisSaclay, 91192 Gif-sur-Yvette, France. E-mail: \protect{merouane.debbah@ku.ac.ae}.}
\thanks{N. Mandayam is with the Wireless Information Network Laboratory (WINLAB), Department of Electrical and Computer Engineering, Rutgers University, New Brunswick, NJ 08902 USA. E-mail: \protect{narayan@winlab.rutgers.edu}.}
\thanks{Z. Han is with the Department of Electrical and Computer Engineering, University of Houston, Houston, TX 77004 USA, and also with the Department of Computer Science and Engineering, Kyung Hee University, Seoul 446-701, South Korea. E-mail: \protect{zhan2@uh.edu}.}}%

\author{Walid~Saad,~\IEEEmembership{Fellow,~IEEE,} 
Omar~Hashash,~\IEEEmembership{Graduate~Student~Member,~IEEE,}
Christo~Kurisummoottil~Thomas,~\IEEEmembership{Member,~IEEE,}
Christina~Chaccour,~\IEEEmembership{Member,~IEEE,}
M{\'e}rouane~Debbah,~\IEEEmembership{Fellow,~IEEE,}
Narayan~Mandayam,~\IEEEmembership{Fellow,~IEEE,}
and~Zhu~Han,~\IEEEmembership{Fellow,~IEEE} \vspace{-0.2cm} }%
\maketitle

\begin{abstract}

Building next-generation wireless systems that could support metaverse services like digital twins (DTs) and holographic teleportation is challenging to achieve exclusively through incremental advances to conventional wireless technologies like meta-surfaces or holographic antennas. While the 6G concept of artificial intelligence (AI)-native networks promises to overcome some of the limitations of existing wireless technologies, current developments of AI-native wireless systems rely mostly on conventional AI tools like auto-encoders and off-the-shelf artificial neural networks. However, those tools struggle to manage and cope with the complex, non-trivial scenarios appearing in the network environment and the growing quality-of-experience requirements of the aforementioned, emerging wireless use cases. In contrast, in this paper, we propose to fundamentally revisit the concept of AI-native wireless systems, equipping them with the \emph{common sense} necessary to transform them into \emph{artificial general intelligence (AGI)-native systems}. Our envisioned AGI-native wireless systems acquire common sense by exploiting different cognitive abilities such as perception, analogy, and reasoning, that can enable them to effectively generalize and deal with unforeseen scenarios. The proposed AGI-native wireless system is mainly founded on three fundamental components: A perception module, a world model, and an action-planning component. Collectively, these three fundamental components enable the four pillars of common sense that include dealing with unforeseen scenarios through horizontal generalizability, capturing intuitive physics, performing analogical reasoning, and filling in the blanks.
Towards developing these components, we start by showing how the perception module can be built through abstracting real-world elements into generalizable representations. These representations are then used to create a \emph{world model}, founded on principles of causality and hyper-dimensional (HD) computing.
Specifically, we propose a concrete definition of a world model, viewing it as an HD causal vector space that aligns with the intuitive physics of the real world -- a cornerstone of common sense.
In addition, we discuss how this proposed world model can enable analogical reasoning and manipulation of the abstract representations. Then, we show how the world model can drive an action-planning feature of the AGI-native network. In particular, we explain how brain-inspired methods such as integrated information theory and hierarchical abstractions play a crucial role in the proposed intent-driven and objective-driven planning methods that maneuver the AGI-native network to plan its actions. Next, we discuss how an AGI-native network can be further exploited to enable three use cases related to human users and autonomous agents applications: a) analogical reasoning for next-generation DTs, b) synchronized and resilient experiences for cognitive avatars, and c) brain-level metaverse experiences exemplified by holographic teleportation.
Finally, we conclude with a set of recommendations to ignite the quest for AGI-native systems. Ultimately, we envision this paper as a roadmap for the next-generation of wireless systems beyond 6G.

\end{abstract}

\section{Introduction} 

In the next decade, novel wireless use cases, such as the metaverse and holographic societies, are anticipated. Those use cases will largely strain the communication limits of modern-day wireless systems due to their unique performance requirements, which are quite different from conventional use cases like smartphone-centric services or intelligent transportation, that were the key drivers for 5G and early 6G systems~\cite{saad2019vision}. For instance, the metaverse will blend the physical-virtual-digital dimensions. Herein, supporting metaverse components such as avatars and \acp{DT} in their ultimate versions over future wireless networks requires meeting novel communication, computing, sensing, and \ac{AI} challenges. For example, endowing avatars with cognitive abilities to faithfully embody \ac{XR} users will require achieving a stringent \ac{E2E} synchronization. Meanwhile, real-world \acp{DT} will require real-time physical-digital interactions and human-like decisions to enable a seamless digital world experience~\cite{hashash2023seven}.
\begin{figure*}
	\centering
	\includegraphics[width=\linewidth]{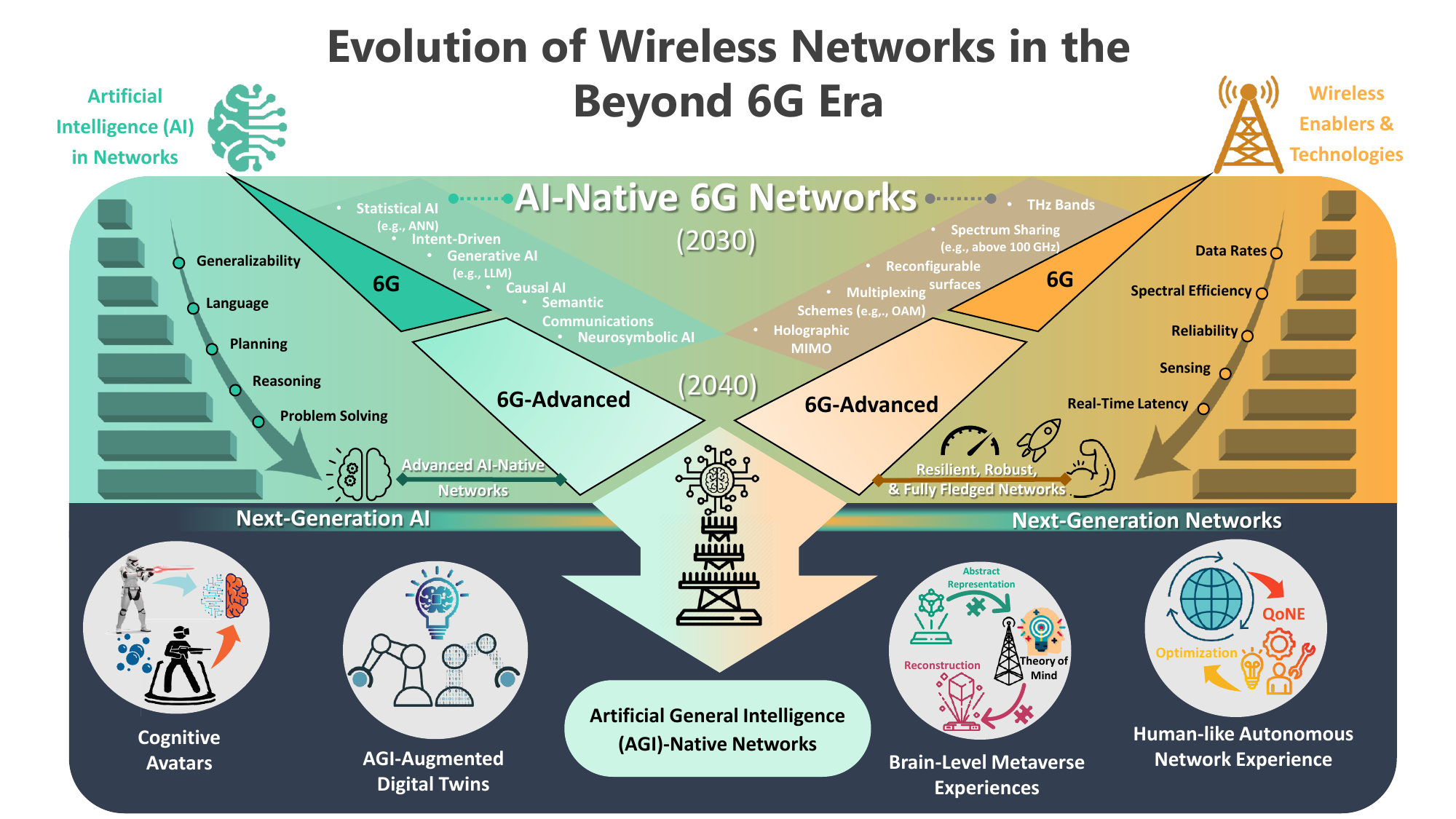}
	\caption{\centering \small{Overview of the evolution of wireless networks from 6G to the beyond 6G era converging towards our envisioned, next-generation AGI-native networks and their corresponding use cases.}}
	\label{evolution}
\end{figure*}
This, in turn, constrains the underlying network with an evolved set of unprecedented requirements that include real-time latency, extreme reliability, and advanced \ac{AI} capabilities. Clearly, on their own, incremental extensions to conventional communication technologies that have driven the evolution from 4G to 6G (e.g., exploiting larger antenna arrays, enhancing multiplexing schemes, etc.) are simply not sufficient to meet the aforementioned challenges of forthcoming wireless services. This is because physical enablers, technologies, and resources are closely approaching their fundamental limits.

Thus, it is natural to ask, what is the next game-changing technology that can potentially help wireless systems overcome the limitations of traditional enablers -- a question that we will deeply explore in this paper. 
For example, one can imagine an avatar experience in the metaverse as one exciting next-generation application, and think about the new type of wireless technologies that we should invest in to fully embody the human in the avatar. For instance, will it be sufficient to explicitly factor in the physics and underlying electromagnetic principles of multi-antenna technologies, or perhaps just exploit sensing modalities as is being done today for 6G? The answer is likely no, because those technologies alone, while important, will still be limited by traditional constraints, like interference, susceptibility to blockage, high propagation losses, and physical network capacity. This, in turn, will most likely keep their prospective performance gains insufficient for supporting the unimaginable new use cases brought forward by the metaverse and its derivatives. 

\begin{figure}
	\centering
	\includegraphics[width=\linewidth]{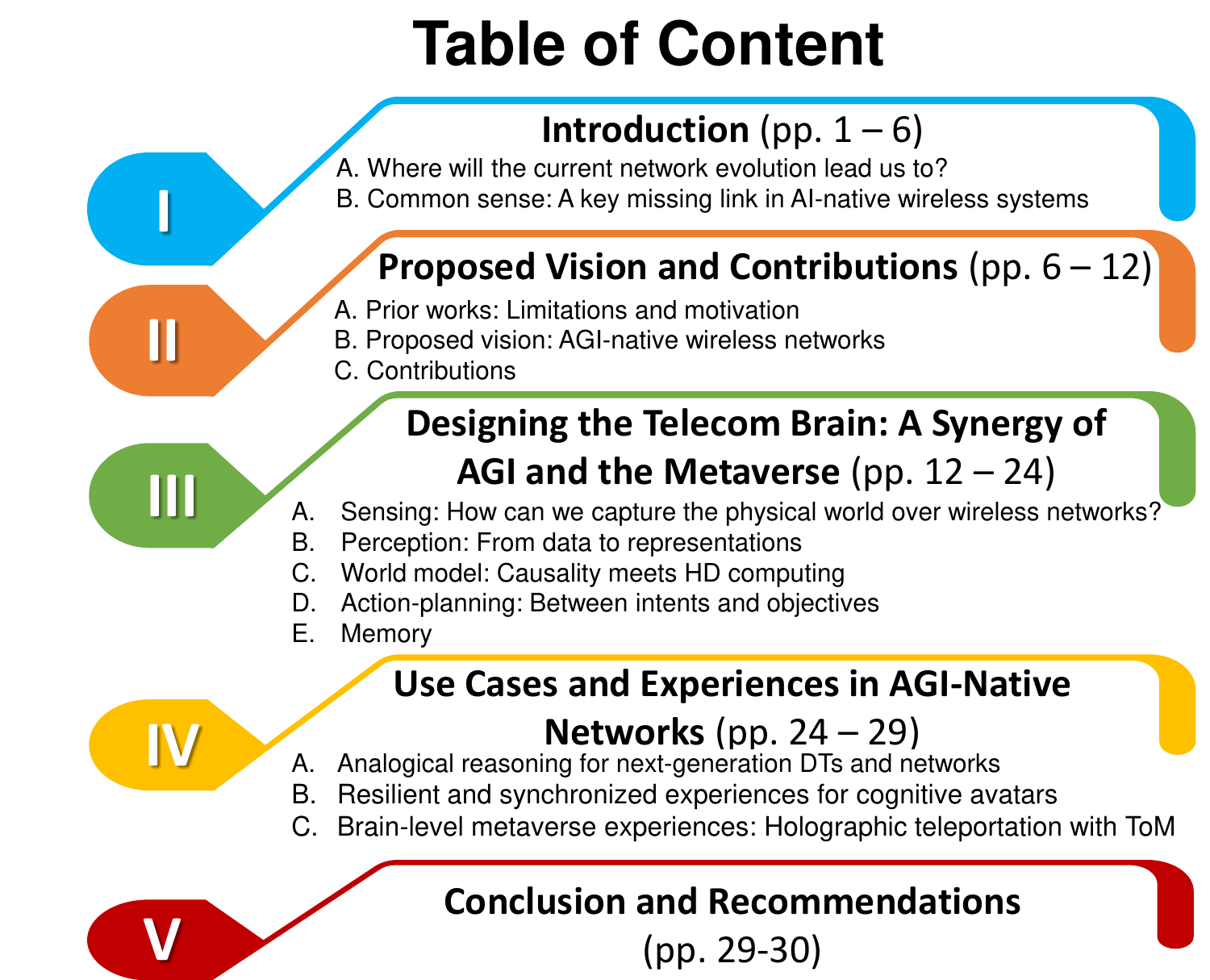}
	\caption{\centering \small{Organization of the sections in this paper.}}
	\label{ToC}
\end{figure}

To answer the aforementioned question, we envision a journey from 6G systems towards their next generation, while passing through the different milestones and technologies that are explained in the various sections of this paper, as summarized in Fig.~\ref{ToC}.

\subsection{Where will the current network evolution lead us to?}

As shown in Fig.~\ref{evolution}, the 6G-era evolution of wireless cellular networks has been mainly driven by two major routes: 1) the evolution of conventional communication technologies, and 2) the exploration of the role of AI systems (culminating in the concept of \emph{\ac{AI}-native systems} for 6G). Moreover, this evolution can be further broken down into two network phases: 6G and 6G-advanced networks.

\begin{figure}
	\centering
	\includegraphics[width=\linewidth]{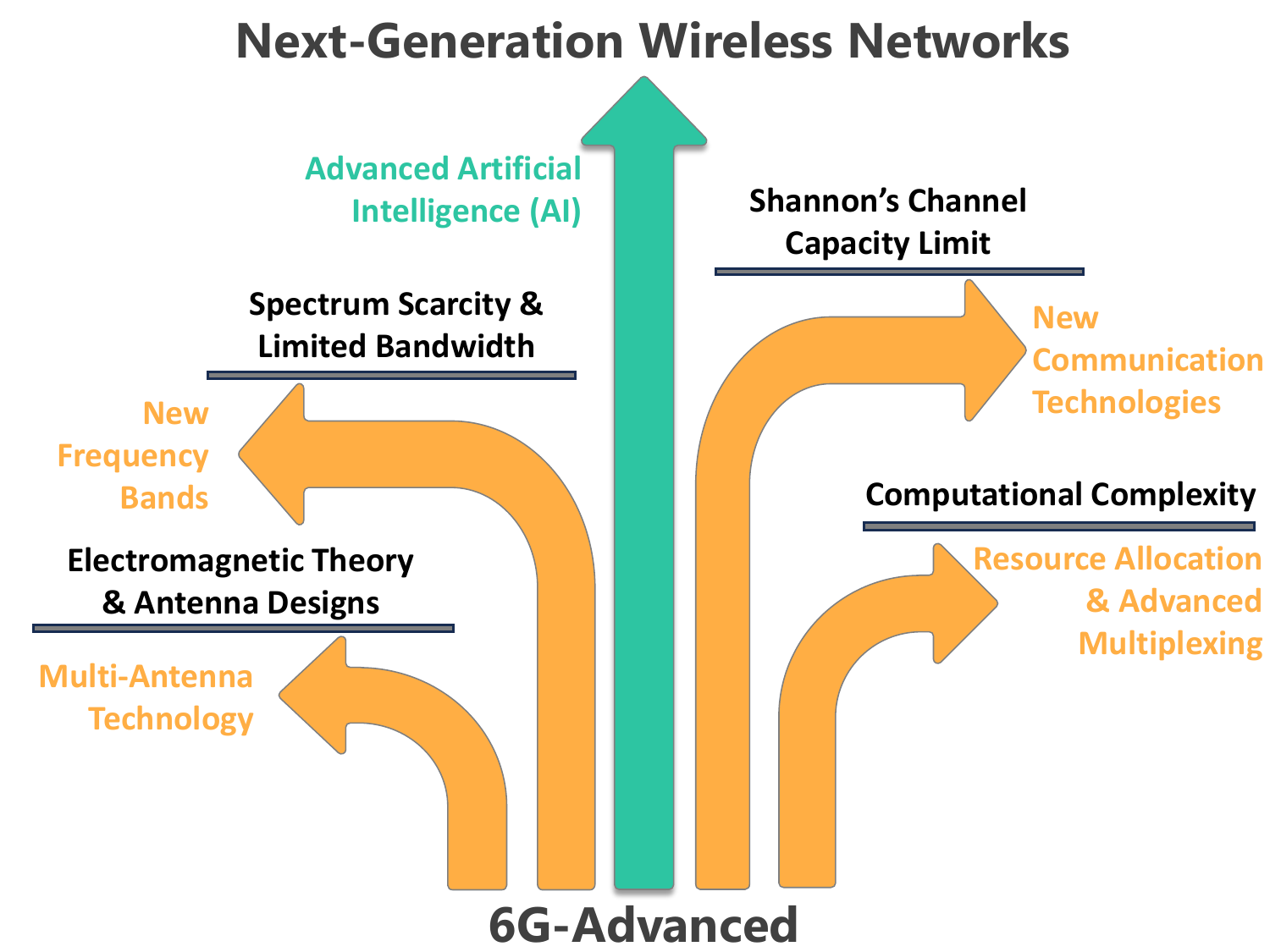}
	\caption{\centering \small{Illustrative figure showcasing the physical constraints facing wireless enablers in the evolution from 6G-Advanced towards the next-generation of wireless networks. Note that this figure is illustrative only and does not purport to showcase exact quantitative numbers.}\vspace{-0.2cm}}
	\label{From_6GAdvanced_To_NextGeneration}
\end{figure}

Conventionally, every generation of wireless networks since 2G has been defined by new multi-antenna and communication technologies (e.g., holographic \ac{MIMO}, \acp{RIS}, etc.), efficient resource allocation and advanced multiplexing schemes (e.g., orbital angular momentum), and the opening of new frequency bands in quest for additional bandwidth (e.g., \ac{mmWave} and \ac{THz} bands). While this path has been effective in leading us to 6G, its limitations are rapidly becoming apparent. For example, while stacking multi-layer meta-surfaces can enable the convergence of communication and computing\cite{an2023stacked}, it cannot inherently address challenges related to antenna impedance matching. Moreover, exploiting holographic \ac{MIMO} technologies can significantly increase communication capacity~\cite{huang2020holographic}, however, 
it will not be able to overcome the fundamental limitations posed by channel conditions and resulting near-field propagation environments. Meanwhile, revisiting electromagnetic information theory~\cite{zhu2024electromagnetic} to overcome challenges like antenna coupling will likely help in optimizing the energy efficiency of communication systems; however, it cannot deal with the degraded performance of wireless systems when the assumed channel models fail to accurately represent the real-world propagation characteristics.
As evident from the previous examples, incremental extensions to conventional technologies are not a sustainable path towards a truly disruptive paradigm shift in wireless networking. The culprit is that, despite some impactful innovations, these technologies will remain limited by the different laws of electromagnetic theory and antenna designs (e.g., antenna-wavelength spacing, radiating aperture size, etc.), as well as by hard constraints like the Shannon capacity limit and the spectrum scarcity. In addition, achieving granular advancements in these communication technologies is accompanied with computationally intensive and complex solutions that may include impractical assumptions (e.g., perfect channel conditions). Hence, asymptotically, under practical considerations, and without discounting advances in the aforementioned fields, we cannot solely rely on such solutions\footnote{We acknowledge that in the current state of wireless research, there is a need for both: (a) developing mature technologies to meet the direct short-term requirements of society~\cite{pegararo-6g}, and (b) developing research visions and roadmaps to shape the long-term evolution of the wireless landscape. Clearly, this work falls into category (b) while naturally building on the state-of-the-art activities of \ac{AI} in 6G recommendations~\cite{itu-r-m2160-0}, that will play an instrumental role to achieve this vision.}, as illustrated in Fig.~\ref{From_6GAdvanced_To_NextGeneration}.
Considering these limitations, the era of 6G becomes a central moment to question this unsustainable evolution towards next-generation wireless systems, by asking a rather existential question: 
\begin{quote}
    ``\emph{What innovation could truly disrupt wireless technologies, allowing them to autonomously and intelligently manage their physical communication constraints?" } 
\end{quote}

The answer to this question could potentially lie in the second, \ac{AI} path that wireless evolution has taken, starting in the 6G era. Indeed, as shown in Fig.~\ref{evolution}, 6G ignited the alternative route of AI-native wireless systems, that focuses on embedding an \ac{AI}-based infrastructure across the various layers and functions of a wireless network. In \ac{AI}-native systems, \ac{AI} becomes a central component for deploying, optimizing, and operating communication networks throughout their lifecycle~\cite{britto2023telecom}. \ac{AI}-native systems can learn and improve their performance by exploiting advanced learning techniques that enable wireless networks to gain system knowledge and expand it into different scenarios.
For instance, the possible adoption of \ac{AI} into the radio interface \cite{hoydis2021toward} could boost its performance transforming it into a largely autonomous, adaptive, and generalizable system that can handle different settings and scenarios of operation~\cite{chen20235g}. 
For example, consider a downlink multi-user scenario in which the air interface optimizes the beamforming configuration at the \ac{BS} for a certain network environment~\cite{yuan2020transfer}. Due to the dynamic nature of wireless environments, a distribution shift (e.g., in the channel gain) can lead to a mismatch with the trained \ac{AI} model of the air interface. This, in turn, can possibly deteriorate the \ac{SINR} at the user side. To mitigate this phenomenon, this air interface can then directly leverage the knowledge it has attained from one environment to swiftly adapt its model for executing its beamforming strategy in this new environment. 
This generalization can be done with \ac{ML} techniques like meta-learning and transfer learning \cite{akrout2023domain}. 

Despite their promising potential for solving such wireless problems, these classical \ac{AI} solutions suffer from multiple drawbacks that can limit their applicability.
In particular, current \ac{ML} models often rely on \acp{NN} that tend to capture \emph{highly non-linear, statistical} relationships and, thus, remain greatly influenced by their training data. Indeed, \acp{NN} often require frequent re-training and adaptation of their underlying models with every domain variation. Moreover, these models tend to lose their effectiveness (i.e., by becoming either rigid or plastic \ac{AI} models~\cite{hashash2022edge}) after multiple updates. In addition, their acquired knowledge diminishes rapidly when the respective testing domains heavily differ (statistically) from those of the initial training phase. 
Beyond constraining their generalization capabilities, this continuous stream of model updates will also result in significant communication and computing resource drainage. Hence, relying on such statistical, black-box models prevents the wireless network from reaching full generalizability and effectively accumulating its knowledge -- two features that are necessary to create truly autonomous wireless systems.

One possible way to overcome this challenge is through the incorporation of cognitive features in the design of \ac{AI} systems. By doing so, one can build \emph{advanced AI-native systems} that can better adapt to dynamic network conditions, improve contextual awareness, and enhance decision-making capabilities, leading to more efficient and reliable network operations~\cite{aref2022impact}. In the identified 6G-advanced era, these advanced \ac{AI}-native networks are envisioned to adopt rule-based solutions that go beyond statistical \ac{AI} models.
This approach yields new possibilities for maneuvering the network to enhance its generalizability and ensure its trustworthiness. One way to achieve this is by embedding reasoning capabilities into the network's nodes (i.e., \ac{Tx} and \ac{Rx}). \emph{Reasoning} essentially means allowing the network to make sense of information and using inference to reach conclusions from its acquired knowledge.
In particular, \emph{causal reasoning}~\cite{thomas2023causal} is one notable form of reasoning that has been proposed for enabling wireless networks to uncover cause-effect relationships existing within the network data and extrapolate a myriad of logical results in the form of interventional and counterfactual operations. This is particularly important in certain domains such as \ac{THz} beam training~\cite{thomas2023reliable} and semantic communications~\cite{ChristoTWCArxiv2022}. In the \ac{THz} regime, due to the high susceptibility of the signals to blockages that limit the \ac{LoS}, the channel response can exhibit highly dynamic changes.
Hence, leveraging statistical \ac{ML} models at the air interface level may not be sufficient to carry out the beam selection in such a scenario. This is mainly because classical \ac{ML} models purely rely on capturing the correlations between variables. As such, the severe fluctuations in the channel response arising at \ac{THz} bands makes the correlation between the channel variables and corresponding beam index (as an output label) highly variable. A possible solution here could be to adopt causal \ac{AI} schemes that rely on capturing causal relationships in the data rather than correlations. In fact, we have demonstrated in our previous work~\cite{thomas2023reliable} that leveraging causal \ac{AI} can help reduce the amount of \ac{AI} re-training needed by capturing a more robust, generalizable relationship in the data. Thereby, empowering wireless networks with reasoning capabilities, through concepts like causality, represents a major block on the path towards realizing advanced \ac{AI}-native networks.

Nevertheless, advanced \ac{AI} and cognitive capabilities span more than cause-effect relationships and reasoning faculties. In fact, the set of human cognitive skills does include several other functions, such as \emph{planning}. For instance, the planning ability represents the essence of the problem solving skills attributed to humans~\cite{biundo2011advanced}. In particular, planning is the process of instantiating a sequence of actions attempting to achieve a particular goal with minimal cost. Similar to humans closing in on a goal through coherent actions, \ac{AI} systems can also be driven to formulate plans with intermediate steps to fulfill given objectives. This is of particular importance for the emerging concept of intent-driven networks~\cite{LeivadeasCST2022}, defined as networks that must navigate and precisely control their resources to fulfill overarching intents. One simplistic example of an intent could be to define a goal of minimizing network energy consumption by 5\%, while still guaranteeing a certain \ac{QoE} for the \ac{UE}~\cite{zou2023wireless}. In fact, such intent or objective-based planning could possibly be handled and incorporated with generative \ac{AI} tools like \acp{LLM}~\cite{shen2024large} and \cite{xu2024large}. 
In our example, a sequential planning of steps may include the design of efficient precoding schemes, followed by optimizing the response of the \acp{RIS} in the network, and subsequent optimization of downlink communication resources. 

Eventually, the convergence of intent-based networking, reasoning, and planning aims to establish \emph{fully autonomous zero touch networks} driven by their intents and objectives. This can facilitate automating the network deployment and adaptation on a dynamic basis. In such a case, the network must continuously adapt its real-time performance with limited human intervention, and in a standalone fashion. Thus, it is anticipated that these autonomous networks can exhibit intelligent responses and decisions that resemble those of humans.
As shown in Fig.~\ref{evolution}, as we continue to equip the network with more reasoning and planning capabilities in the beyond 6G era, we will approach a plateau of advanced intelligence levels that must drive the autonomous operations of 6G-advanced networks.

From the above discussion, it is evident that the incorporation of cognitive abilities such as reasoning and planning represents a stepping stone to evolve \ac{AI}-native wireless systems and help them meet the challenges of future services.
Yet, while valuable, these cognitive abilities alone do not adequately equip a communication system to completely curb and tame the dynamic nature of the wireless \ac{RAN} and its complex environment. Indeed, networks that lack full generalization capabilities cannot become fully autonomous, and they will not be sufficient to create a new generation of communication systems.
Hence, a fundamental question arises here: ``\emph{How can we design intelligent wireless systems with new cognitive abilities that can become fully autonomous and potentially usher in a new ``G" of networks?}''


This is the core question that this paper will seek to answer, and in the path to do so, we reflect back to an inspirational quote:\vspace{-0.3cm}

 \epigraph{``\emph{Every generation imagines itself to be more intelligent than the one that went before it, and wiser than the one that comes after it.}"}{George Orwell} 
Next, we explain what is the missing link from current \ac{AI}-native networks that must be addressed for reaching the next-generation of wireless systems.


\subsection{Common sense: A key missing link in \ac{AI}-native wireless systems}
\label{common sense in AI native}

Although reasoning and planning constitute an important part of cognitive abilities, their current forms remain insufficient for a network to become fully autonomous, driven by its intents, similar to humans.
On the one hand, while causal reasoning can help in generalization, it remains task specific and limited to the in-domain (i.e., out-of-distribution) context. In other words, current reasoning frameworks~\cite{jiang2021edge, sloman1996empirical, pearl2009causal}, like causal \ac{AI}, on their own, may struggle with \emph{out-of-domain} generalization to unfamiliar, ``\emph{corner cases}", that have never been witnessed before.

On the other hand, exploring state-of-the-art solutions like \acp{LLM} to perform the planning steps in a wireless system will be susceptible to hallucinations that can initiate non-logical steps. This is due to the fact that \acp{LLM} possess limited reasoning power and lack experience and general knowledge about the world, while missing the fundamental elements of problem solving: decisions, objectives, and transition models of the problem~\cite{valmeekam2022large}.
Although there has been some attempts to equip foundation models with reasoning capabilities through causality~\cite{zhang2023towards} and chain-of-thought reasoning~\cite{zhang2023igniting}, such solutions primarily reduce hallucinations over the training data, rather than bolstering generalization to new scenarios. Meanwhile, although some recent solutions~\cite{kirk2023understanding} develop \acp{LLM} that can generalize to out-of-distribution scenarios (i.e., cope with distribution shift), \ac{LLM} approaches cannot handle out-of-domain scenarios beyond their training data~\cite{xu2024large}. Consequently, this can hinder planning in real-world situations that are full of corner cases and continuously confronted with unfamiliar situations. In other words, to be truly autonomous, wireless systems should know how to plan even in novel situations.

\ac{AI}-based planning has also been closely tied to \ac{ML} techniques like \ac{RL}~\cite{moerland2023model}. Although such frameworks can possibly plan and progress towards an intended goal, they can only do so in a closed environment that consists of limited action/state spaces. 
This can limit the generalizability of these frameworks and hinder planning in unforeseen, out-of-domain scenarios, different tasks, and non-stationary environments beyond this limited space. Hence, \ac{RL} and its variants eventually tend to learn and memorize, rather than to solve problems in scenarios with open possibilities.
Therefore, the current line of reasoning and planning in \ac{AI}-native networks, (e.g.,~\cite{thomas2023reliable, zou2023wireless, xu2024large}, and~\cite{tarkoma2023ai}) is largely constrained by their limited capabilities of generalization to unfamiliar scenarios~\cite{valmeekam2022large}. This is possibly due to their lack of adequate knowledge of the basic principles about the world around them.

\begin{figure}
	\centering
	\includegraphics[width=\linewidth]{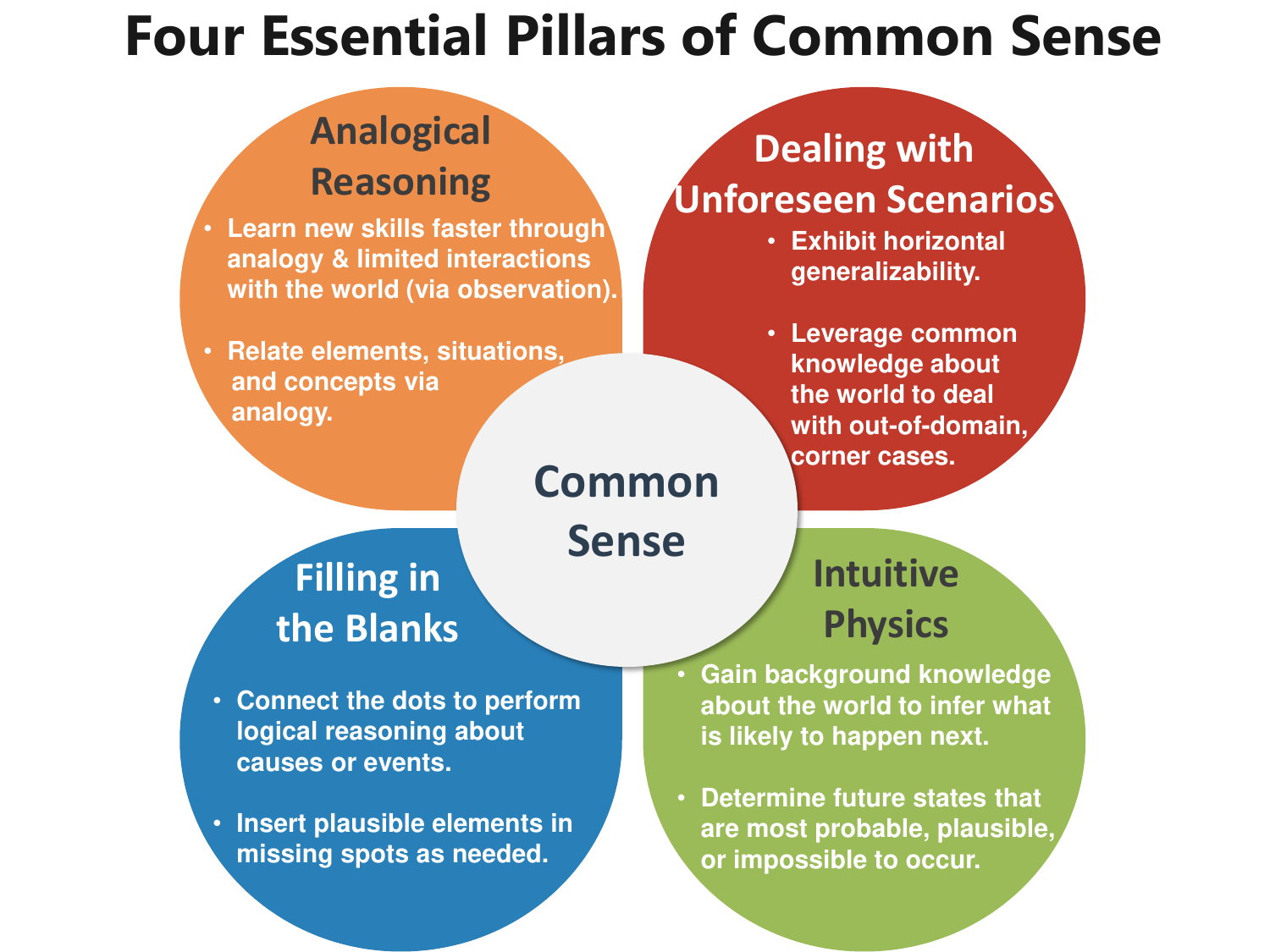}
	\caption{\centering \small{Illustrative figure showcasing the four essential pillars of common sense.}}
	\label{Common Sense}
\end{figure}

Evidently, the design of a generalizable \ac{AI} system that can adapt to diverse and dynamic network conditions remains a persistent challenge for autonomous networks.
To address this issue, attempts have initially focused on training statistical \ac{AI} models with massive data samples and millions of \ac{RL} trials~\cite{kasgari2020experienced}. This aims to expose \ac{AI} systems to every plausible scenario that they can possibly encounter.
Nevertheless, this solution cannot effectively generalize and deal with specific, risky, and rare case scenarios. 
Although causal \ac{AI} can generalize the relation between the cause and effect (i.e., vertical generalizability),
yet, as stated by Y. LeCun~\cite{LeCun2022OR}, we still do not have an \ac{AI} system that can deal with new, unfamiliar, and out-of-domain scenarios. This is because \ac{AI} systems lack \emph{horizontal generalizability}, which is the missing component preventing them from becoming fully autonomous and independent~\cite{scholkopf2021toward}. Horizontal generalizability deals with the ability to generalize to out-of-domain distributions. This lack of horizontal generalizability mainly pertains to the absence of \emph{common sense} in \ac{AI}. Common sense is a cognitive trait that can be majorly defined by the four key technical pillars that we have concretely defined in Fig.~\ref{Common Sense}.
Basically, common sense carries humans out of trouble when dealing with the endless unforeseen scenarios that they encounter on a daily basis in the real-world. It also enables humans to relate concepts and learn much faster, by \emph{analogy} (i.e., \emph{analogical reasoning}). In addition, it helps them connect the dots to reach logical deductions, and fill in plausible, missing elements as needed. In short, common sense is the background knowledge about the world that enables individuals to infer what is likely to happen next. This definition considers the basic context of common sense which broadly refers to the core skills of intuitive physics (i.e., object navigation and manipulation)\footnote{While intuitive psychology (e.g., social cognition) is also related to common sense, it is less considered in the scope of this work.}. These skills involve innate concepts and principles that humans grasp by understanding the physical behaviors in the world. For instance, such basic principles include intuitively knowing that a ball will fall to the ground when it is dropped because of its gravitational weight. 

In unforeseen scenarios, humans rely on their common sense to reasonably navigate out of difficult situations, whereas \ac{AI} systems lacking this ability encounter major challenges in doing the same. For example, a wireless network may fail to deliver the \ac{QoE} required by proper resource management when the application domains changes.
As a very simple example, the network can be trained to properly allocate a beam for \ac{XR} users, yet, it fails to handle the beam assignment for a setting with autonomous vehicles.
In general, this challenge will persist as long as \ac{AI} systems remain rule-based systems that tend to just extract patterns from their training data, and capture the underlying correlation and causal relationships hidden within, without relying on common sense. As a matter of fact, current \ac{AI} systems lack this common sense because they \emph{learn from data and not from the world itself}. In other words, today's \ac{AI} systems do not understand how the world works. 
In concert with~\cite{yu2023explainable}, we posit that this common sense can only be acquired by grasping the ability to \emph{learn world models}.

Once \ac{AI} systems are equipped with world models, they can engage in the adequate reasoning and planning that would allow them to actually become autonomous. On the one hand, reasoning via common sense can provide ways to generalize (e.g., via analogical reasoning) and deal with unforeseen scenarios. On the other hand, planning by leveraging a world model brings forth a rigorous approach for action-planning, whereby planning is merged with reasoning about the general knowledge of the world.

Therefore, acquiring common sense through a world model plays an instrumental role in the path to designing advanced \ac{AI} systems with human-like cognitive abilities. Fundamentally, a core, fundamental element of human intelligence pertains to building the capability to simulate the physical world~\cite{li2020causal}. Toward this end, a world model enables predicting the different plausible future states resulting from the actions that could be performed. Hence, \ac{AI} systems that understand their underlying world can foresee the consequence of their actions if they were to be executed, similar to what humans do. 
That is, humans mainly learn through observation and limited interactions with the world in a \emph{task-independent, unsupervised way}~\cite{LeCun2022OR}, rather than through a large volume of labeled data samples and numerous expensive trials of \ac{RL}. This observation of the world is followed by simulating the specific scenarios that incorporate their background knowledge, before they attempt to act. Thus, the aim of this simulation in \ac{AI} systems is to solve problems and plan actions, upon emulating the abilities of humans to \emph{think} and \emph{imagine} beforehand. Therefore, \emph{common sense drives in new human-like cognitive abilities for thinking about actions and imagining the world}.

Nevertheless, simulating the physical world would not just require attaining a world model, but it also requires accurate \emph{perception} of its real-time status. In fact, perception is another crucial ability missing in most of today's \ac{AI} systems. Perception typically relies on estimating the state of world and representing it in the form of \emph{abstractions}~\cite{bengio2013representation}. These abstractions play a crucial role in the \ac{AI} system's ability to think about the world elements and their relationships~\cite{nam2022learning}, and they are the key to carry out analogy between these elements.

Clearly, there is a need to integrate more advanced cognitive abilities into \ac{AI}-native wireless networks, primarily common sense, in order to achieve true levels of intelligence and generalization. Once these generalization and intelligence levels are reached, wireless networks can then deal with unforeseen scenarios during reasoning and planning, thereby enabling truly autonomous networks. Hence, the answer to our earlier question on the design of new \ac{AI}-native networks with cognitive abilities lies in the integration of common sense. Indeed, to unleash a new ``\emph{G}", wireless networks must operate with advanced human-like cognitive abilities. A key byproduct of this common sense integration will be a much anticipated transition from \ac{AI} towards \emph{\ac{AGI}}, whose ultimate goal is essentially to replicate the broad range of human cognitive abilities~\cite{goertzel2014artificial}.
As will be evident from subsequent sections, this paper will design a new generation of wireless networks with \ac{AGI} abilities by equipping them with the common sense necessary to facilitate other crucial cognitive abilities such as imagination, thinking, and perception, along with reasoning and planning.
\vspace{-0.2cm}

\section{Proposed Vision and Contributions}

\subsection{Prior works: Limitations and motivation}
\label{State_of_the_art}

Designing a wireless system with \ac{AGI} abilities has not been studied in prior works to date. However, in some recent works like~\cite{bariah2023ai}, there has been some ``hints'' towards the interplay between \ac{AGI} and wireless networks. For instance, in~\cite{bariah2023ai}, the authors discuss the use of embodiment to present some form of \ac{AGI} in 6G networks. While this prior work discusses \ac{AGI} as a concept, it does not have a framework to truly achieve \ac{AGI} over the network, but instead, it relies on the principle of \ac{AI} embodiment that grants \ac{AI} systems the abilities to interact with the world. Moreover, the work in~\cite{bariah2023ai} is impractical because learning in the physical environment itself can incur irreducible, risky costs for \ac{AI} systems, similar to \ac{RL} that learns through trial and error in the real-world. In addition, this work does not highlight the specific role of wireless functionalities in the interaction and perception processes. 

Furthermore, since the world can be largely explained in terms of cause and effect~\cite{sloman2005causal}, there has been a number of works that looked at the use of causal structures for designing~\cite{gupta2024essential} or reasoning over world models e.g.,~\cite{yu2023explainable} and~\cite{li2020causal}. 
Indeed, the work in~\cite{gupta2024essential} leverages a causal foundation model to model the world for embodied \ac{AI} interactions. Nevertheless, the solution of~\cite{gupta2024essential} cannot perceive generalizable abstractions\footnote{Herein, the term ``generalizable" refers to obtaining a common general denominator between abstractions of similar real-world elements. This is necessary to proficiently approach unforeseen elements. Moreover, this encompasses the generalizabilty of the representation itself by remaining invariant to out-of-distribution shifts.} 
of the world, and it lacks true transparency since it still relies on black-box foundation models. 
Moreover, the work in~\cite{kipf2019contrastive} leverages contrastive learning to disentangle and perceive abstract representations of objects in world models. However, this prior work does not take into account the crucial role that analogical reasoning plays in the perception of unforeseen objects and how to relate them to generalizable abstract representations.
Indeed, relating unforeseen objects to similar, real-world elements is largely overlooked in~\cite{gupta2024essential} and~\cite{kipf2019contrastive}. 
Alternatively, in more transparent models like that of~\cite{yu2023explainable}, a world model is designed as a \ac{SCM} to assist in explainable \ac{RL} decisions. In~\cite{li2020causal}, a world model is constructed in a causal partially observable Markov
decision process to give an autonomous agent the abilities of imagination for physical reasoning. Considerably, the presented models in both~\cite{yu2023explainable} and~\cite{li2020causal} are confined to a closed environment comprising a limited set of probabilistic action/state spaces that do not encompass representations. As such, intuitive physics operations for object manipulation and navigation in the real-world -- the core of common sense -- is not captured in the solutions of \cite{yu2023explainable} and \cite{li2020causal}. Although some recent works such as~\cite{bubeck2023sparks} provide evidence for common sense emerging in \acp{LLM}, they do not build a real, physically-consistent world model. Hence, while \ac{LLM} designs like those in~\cite{bubeck2023sparks} may answer questions that require some common sense, such answers remain tied to textual knowledge and lack grounding in the physical world. Indeed, common sense acquired from text does not provide sufficient means for generalization to out-of-domain scenarios.



One of the most prominent and comprehensive visions towards \ac{AGI} was articulated by Y. LeCun in~\cite{LeCun2022OR}. For achieving \ac{AGI}, \cite{LeCun2022OR} envisions a modular \emph{cognitive brain architecture} that comprises six different modules representing cognitive abilities: perception, world model, actor, cost, short-term memory, and configurator. As will be evident in the next section, our vision of wireless systems with \ac{AGI} abilities will align with those modules. However, it is not a straightforward application of this prior vision~\cite{LeCun2022OR}. For instance, the AGI view of \cite{LeCun2022OR} faces different challenges that limit its adoption into wireless systems. These challenges stem from the intention of \emph{granting \ac{AGI} abilities directly to individual agents} (e.g., autonomous vehicle, robot, etc.) through this cognitive architecture, an idea that has some key shortcomings:
\begin{enumerate}
    \item \textbf{Independent worlds:} The framework of \cite{LeCun2022OR} considers a single agent scenario and attributes the physical world exclusively to this agent. In reality, there exists different agents that share the physical world as they interact with each other. This can have direct implications on building world models. First, granting individual \ac{AGI} agents the ability to build their own worlds independently will not necessarily lead to having \emph{consistent models of the same physical world}, even if those agents share some information. 
    This is because \ac{AGI} agents build their world models according to their individual experiences and knowledge. Consequently, we cannot guarantee that the predicted ``futures" or planned actions by individual \ac{AGI} agents comply with one another. 
    Second, the interaction between \ac{AGI} agents in a shared space would still require a coordination of their actions. For instance, consider two vehicles at a crossroad. It is natural to ask which vehicle will pass first, even if both can leverage intuitive physics to foresee the possibility of an accident if they do not decelerate. While that may seem very basic and intuitive to human nature, because they lack coordination, \ac{AGI} agents built on \cite{LeCun2022OR}'s approach could stumble in such situations.
    This stems from the lack of wisdom and ethical motives in such agents, that are necessary to navigate in such situations, particularly in the absence of effective means for reliable coordination.
    Alternatively, a possible idea could be to divide the world between agents. Accordingly, each agent would control only their limited part of the world and plan their actions, however, this does not reflect the interactive reality of the world. Finally, predicting the future of a common space would require time synchronization between the \ac{AGI} agents to become effective. Nevertheless, this synchronization is not guaranteed by scattering \ac{AGI} individually across agents, as foreseen in \cite{LeCun2022OR}.

    \item \textbf{Limitless perception:} Perceiving the world and then focusing on the limited part and details relevant to the task (or objective) in hand can have multiple challenges. On the one hand, inferring which parts of the world are most relevant to the task is typically beyond the capabilities of \ac{AGI} agents. In~\cite{LeCun2022OR}, a configurator component is defined for this purpose, however, its elements remain undefined. 
    On the other hand, it is challenging to define the physical limits of perception for an autonomous agent. For instance, consider an autonomous airplane; it is imperative to clarify the limits of the world that it should perceive and what it should focus on. One may argue that it should perceive just what it is able to detect based on its abilities. Another argument may be to perceive the whole world all the way from the Earth up to space as it is relevant to its specific task. In addition, many autonomous lightweight agents (e.g., drones) are often constrained by limited sensing, computing, and storage capabilities, which can be impractical for acquiring common sense (i.e., building a world model) or \ac{AGI}.
    All these aspects related to perception are not addressed in~\cite{LeCun2022OR}.

    \item \textbf{General-purpose agents:} In general, autonomous agents are typically designed to be aware of their numerous, narrow tasks. Hence, they are not necessarily reconfigurable to achieve any arbitrary task as the design implies in~\cite{LeCun2022OR}.
    In other words, equipping autonomous agents with a cognitive architecture, such as in~\cite{LeCun2022OR}, implies that those systems must be able to deal with any objective. Nevertheless, in reality, an autonomous agent may fail to achieve a given objective if it happens to fall outside its scope. Practically, an autonomous agent does not need to perform every task (i.e., general-purpose), but it is rather confined to a defined set of germane tasks (i.e., multi-purpose). For instance, an autonomous vehicle must know the operations needed in the scope of driving, but it will never be oriented to ``fly" like an airplane. This is different from learning a new skill or being directed to fulfill a new objective within its defined scope. Instead, an autonomous vehicle must know how to act in corner cases that appear upon performing its defined narrow tasks. Although agents having \ac{AGI} as in~\cite{LeCun2022OR} can, in an ideal case, solve the majority of these corner case situations, this can become an expensive solution given the massive numbers of autonomous agents that are expected to proliferate over next-generation networks. Instead, it may be more desirable to find sustainable, concise, and steerable solutions that keep \ac{AGI} controllable, while still granting autonomous agents the ability to deal with corner cases, as needed.

\end{enumerate}

\begin{figure*}
	\centering
	\includegraphics[width=\linewidth]{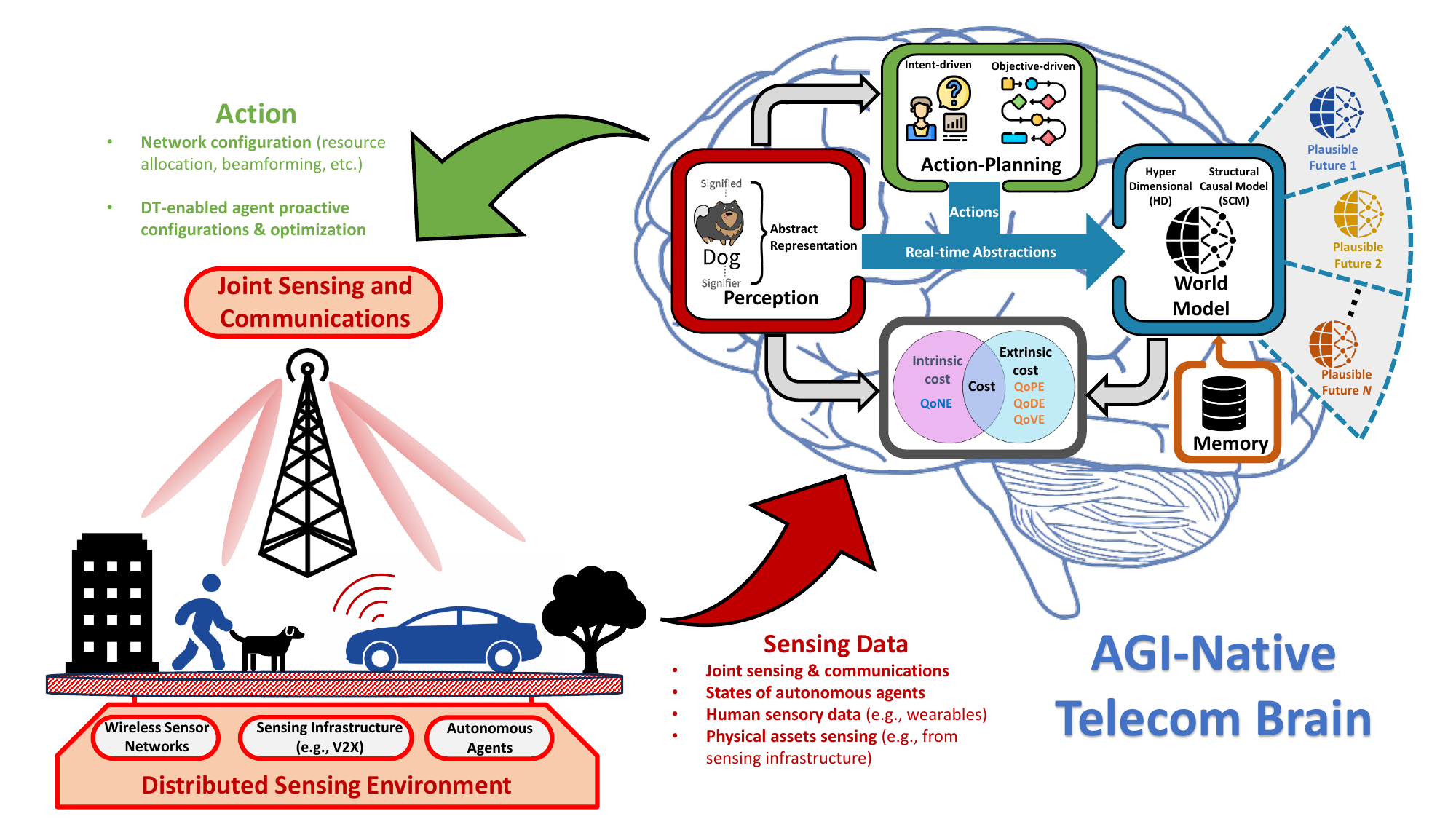}
	\caption{\centering \small{Illustrative figure showcasing the operation of an \ac{AGI}-native telecom brain and its different modules. This design is inspired from the \ac{AGI} architecture in~\cite{LeCun2022OR}, but it refines it for our communication network purposes.}}
	\label{Telecom_Brain}
\end{figure*}

\vspace{-0.1cm}
Despite the above drawbacks, the vision for \ac{AGI} presented in~\cite{LeCun2022OR} provides us with a valuable basis for wireless networks to progress towards human-level \ac{AI}. In a nutshell, the previously discussed works, like~\cite{LeCun2022OR, bariah2023ai, datta2023development} do not consider how wireless networks can reach \ac{AGI} levels, whereby both the network and its agents can operate with \ac{AGI}. Evidently, wireless networks will need to start by building world models of the physical world. To do so, one can exploit the emerging concepts of the metaverse and \acp{DT} because they provide means of replicating the real-world through the lens of the wireless network~\cite{hashash2023seven}. Herein, this can be a promising solution for the network to procure common sense and the autonomous agents to acquire \ac{AGI}. Accordingly, the intersection of the metaverse with future wireless systems might possibly offer a gateway towards providing \ac{AGI} abilities to the network and its autonomous agents. To shed light on this promising avenue, we next present one of the first visions that explores the design of a new generation of wireless system with \ac{AGI} capabilities.

 \vspace{-0.1cm}
\subsection{Proposed vision: \ac{AGI}-native wireless networks}
\label{Vision}

We envision a new breed of wireless systems with \ac{AGI} abilities, that can reason, plan, imagine, think, and have common sense, operating with a novel \emph{cognitive brain architecture} that we call the \emph{telecom brain}, as shown in Fig.~\ref{Telecom_Brain}. Some of the key concepts related to this architecture are summarized in Table~\ref{tab:lexicon}. This architecture tailored to wireless systems comprises three main modules related to the cognitive abilities that we have discussed:
\begin{itemize}
    \item \textbf{Perception:} A perception module allows the wireless network to capture \emph{generalizable abstract representations} from the physical world through a fusion of contrastive learning and causal representation learning. These representations should exhibit an optimal level of complexity that balances between causality and generalizability.
    \item \textbf{World model:} The envisioned world model couples the causal aspect of the world and the transparency of \acp{SCM} with \emph{\ac{HD} computing}~\cite{ThomasJAIR2024}. Thus, our envisioned world model can manipulate the representations in the form of \ac{HD} vectors that are compatible with the \emph{intuitive physics} operations of common sense and suitable for \emph{analogical reasoning}. 
    \item \textbf{Action-planning:} This module considers two main strategies to plan the actions of autonomous wireless systems: a) \emph{intent-driven} planning and, b) \emph{objective-driven} planning. These strategies build on brain-inspired methods such as \ac{IIT}~\cite{tononi2016integrated} and hierarchical abstractions.
\end{itemize}

\begin{figure*}
	\centering
	\includegraphics[width=\linewidth]{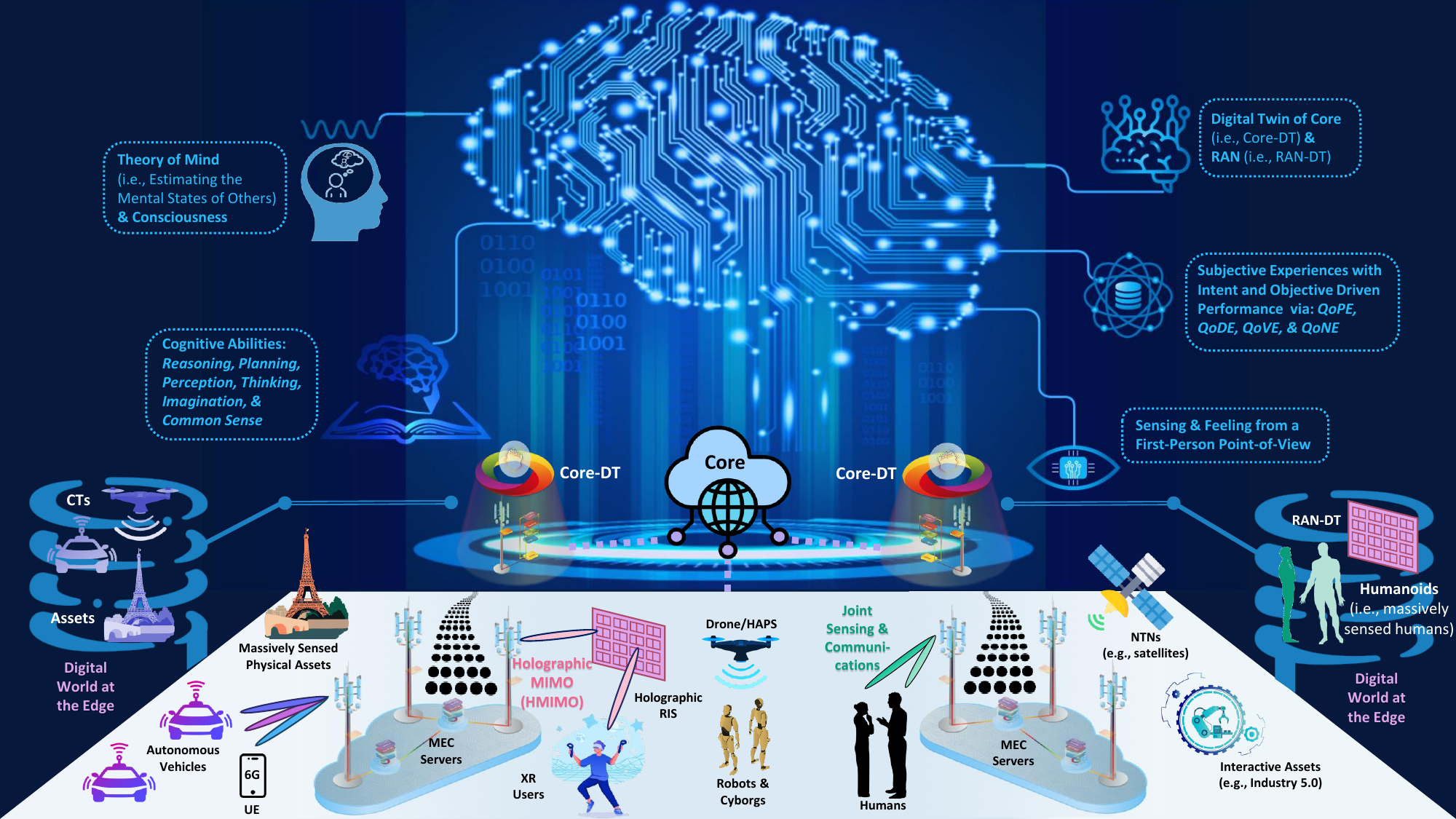}
	\caption{\centering \small{Illustrative figure showcasing the harmonization and synergy between the \ac{AGI}-native telecom brain and the different use cases over next-generation \ac{AGI}-native wireless networks.}}
	\label{System_Model}
 \vspace{-0.15cm}
\end{figure*}

These main three modules also rely on interconnections with a cost module that operates based on various network \ac{QoE} indicators (which are explained and summarized in Table~\ref{tab:lexicon}) as well as with a memory module, as shown in Fig.~\ref{Telecom_Brain}. This envisioned cognitive architecture is anticipated to bring forth unprecedented levels of intelligence, which can transform the wireless network from an \ac{AI}-native system into an \emph{\ac{AGI}-native} system. With common sense, this new generation of networks could achieve a leap in generalization to unforeseen scenarios and autonomous abilities by operating at \ac{AGI} levels.
Thus, we will explore how the cognitive architecture in Fig.~\ref{Telecom_Brain}, with the emergence of the metaverse, will bring in new levels of general intelligence\footnote{We acknowledge that the term \ac{AGI}, when used to refer to actual human-level intelligence, could be misleading since complete, fully-fledged human-level intelligence may never be attained by AI. However, we use this commonly adopted term to refer to an AI system that can have common sense. Although we call this instance of intelligence ``general", we consider it specialized to a multitude of specific domains or tasks. Thus, \ac{AGI} refers to being task-independent, with distinct generalization performance could outperform narrow \ac{AI}. This stems from the fact that even humans are intelligent within specific domains and not in every domain.} into the network. Next, we provide a concise summary of the operation of our three main modules that drives in \ac{AGI} into the network.



\textbf{Perception. }
As evident from Section~\ref{State_of_the_art}, in order to design a wireless system with \ac{AGI} capabilities, we must endow the network with the ability to perceive the physical world. Here, perceiving the state of the world can be equivalent to providing a synchronized, real-time digital replica of it. Remarkably, this is exactly the role of the digital world of the metaverse that captures this replica, while encompassing its different physical constituents (see Fig.~\ref{System_Model})~\cite{hashash2023seven}. These constituents include humans (digitally represented as so-called humanoids), autonomous agents, physical assets (e.g., buildings, infrastructure, etc.), and the network itself (i.e., \ac{RAN} and core). 
Notably, autonomous agents are \ac{DT}-enabled applications, that have their \acp{PT} replicated into the digital world as \acp{DT}~\cite{hashash2023seven}. 
Clearly, autonomous agents that require common sense will now be  perceived as \acp{DT} by the network. 
This is a crucial angle that is surprisingly neglected in works that deal with autonomous agents (e.g., vehicles, drones, etc.), such as~\cite{LeCun2022OR, bariah2023ai}, and~\cite{datta2023development}. In essence, \acp{DT} are bi-directional \ac{AI} models that enable the proactive configuration and performance optimization of autonomous agents~\cite{yu2023internet}. To facilitate their aforementioned roles, the \acp{DT} must acquire their proactive abilities from the world model of the network.

\vspace{-0.1cm}

\textbf{World model and action-planning.}
Integrating the perceived digital world with a world model can allow predicting the plausible future states of the network, including those of the \acp{DT} of the autonomous agents. This can be done by representing the perceived abstractions as \ac{HD} vectors and manipulating them with the actions from the action-planning module. On the one hand, simulating the plausible reality worlds can enable planning the optimal actions to be executed by the \ac{AGI}-native network. As discussed earlier in Section~\ref{common sense in AI native}, this is the essence of \ac{AGI}. On the other hand, the network will now acquire an additional degree of freedom to optimize the future states of the \acp{DT}. For this purpose, the \ac{DT} configuration feedback is passed to the \acp{PT} in the physical world. This feedback includes the configurations needed for the \ac{PT} to reach this optimal (predicted) future state. As such, this feedback can account for any unforeseen scenario that could be encountered by the \ac{PT} in the physical world. Consequently, the \ac{PT} operates as if it has acquired common sense. By leveraging \acp{DT} and an \ac{AGI}-native network, autonomous agents do not need to acquire \ac{AGI} directly as discussed in prior works~\cite{LeCun2022OR}.
In contrast, autonomous agents become \ac{AGI}-augmented \acp{DT} that are endowed with common sense from the network. Therefore, an \ac{AGI}-native network can enable general intelligence on both the network and agent levels, simultaneously. That said, the potential of \ac{AGI}-native networks also extends to enable other use cases beyond revolutionizing autonomous agents.

\begin{figure*}
	\centering
	\includegraphics[width=\linewidth]{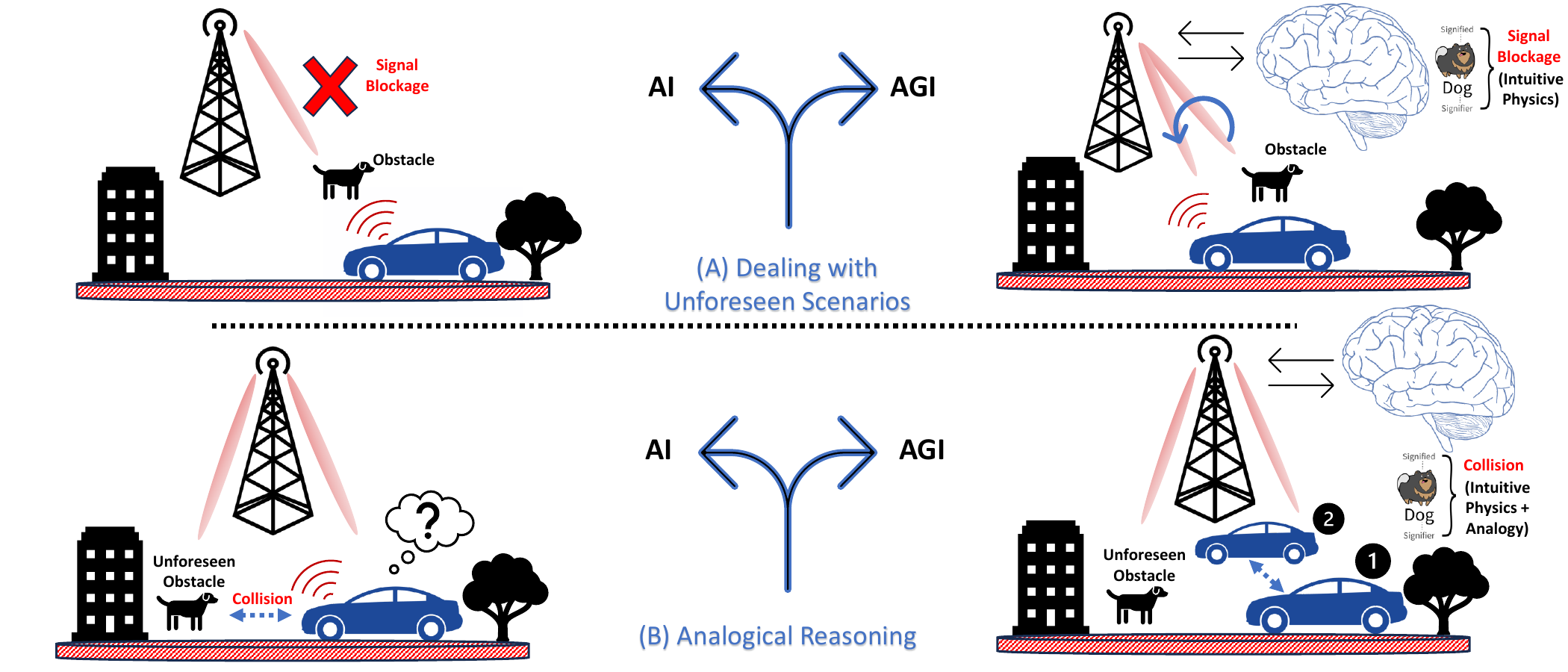}
	\caption{\centering \small{Illustrative figure showcasing simple, direct examples of \ac{AGI}-native networks that consider: (A)~dealing with unforeseen scenarios and (B)~analogical reasoning. Naturally, these examples are provided for illustrative purposes, and the proposed \ac{AGI}-native network will be able to deal with more complex and large-scale use cases.}}
 \vspace{-0.2cm}
	\label{Two_Examples}
\end{figure*}

\textbf{Use cases of \ac{AGI}-native networks.} Evidently, the digital world considers twinning the core and \ac{RAN} elements of the network, such as holographic \acp{RIS} (see Fig.~\ref{System_Model}), on the network itself (i.e., \ac{RAN}-\ac{DT} and core-\ac{DT}). In fact, the network can rely on its twin to plan its actions. Hence, the \ac{AGI}-native network will be driven by its telecom brain, that determines its actions and orchestrates its resources, as shown in Fig.~\ref{Telecom_Brain}.
Similarly, the emergence of \ac{AGI}-native networks is anticipated to revolutionize human-centric applications and experiences of the metaverse as well. In particular, an \ac{AGI}-native network can enable \ac{AI}-driven cognitive avatars that require common sense to faithfully embody and immerse \ac{XR} users over the network. Moreover, an \ac{AGI}-native network can leverage its common sense to estimate the states of network users, which can be vital in enabling novel metaverse applications such as holographic teleportation. For example, these new abilities can play a role in reliably teleporting the interactive assets of industry $5.0$ applications over the network~\cite{hashash2023seven}. 
In Fig.~\ref{System_Model}, we provide an illustration that shows the blend of the telecom brain with human-centric use cases and constituents (see Table~\ref{tab:lexicon} for the definition of constituents).

\vspace{-0.1cm}

\textbf{Examples of analogical reasoning and dealing with unforeseen scenarios.} To further exemplify the abilities of an \ac{AGI}-native network, we can directly extend our previous example on causal reasoning for \ac{THz} beamforming. Let us consider an autonomous vehicle navigating the real-world when suddenly an object appears in its proximity, as shown in Fig.~\ref{Two_Examples}. For the discussion purposes, let us consider this object to be a dog in this case. Next, we will consider two examples to show the vital impact of \ac{AGI} and how the network and vehicle may fail without it. 
In the first example, this object acts an obstacle that blocks the beam from the \ac{BS} to the vehicle. In this case, the \ac{AI} model in the air interface is trained to specify the beam according to the causal \ac{AI} solution (i.e., based on channel response and location of the vehicle). As a result, the network is not trained to this unforeseen scenario and can fail to adjust the beamforming. In contrast, when endowed with \ac{AGI}, the network can identify that the beam would be blocked (i.e., through intuitive physics) and can modify its configuration accordingly to provide an alternative beam.
In the second example, we assume the object crosses in front of the vehicle, and the vehicle has never encountered this unforeseen obstacle. In this case, under a classical \ac{AI}-native system, the action of the vehicle is undetermined as it has never been trained to deal with this object. In contrast, if the vehicle is endowed with \ac{AGI} from the network, then the network could identify this unfamiliar object as an obstacle (e.g., through analogical reasoning). Accordingly, the network can maneuver the vehicle away from this object (e.g., through intuitive physics) to avoid a potential crash.
In both examples, an \ac{AGI}-native network can further deal with these unforeseen situations and objects by assigning the vehicle to a different beam.

Evidently, the envisioned \ac{AGI}-native network can overcome the limitations of task-defined models that have constrained \ac{AI}-native networks, to date, and, instead provide a task-independent model for general intelligence. 
Henceforth, \ac{AGI}-native networks can leverage such new abilities to deliver the \ac{QoPE}, \ac{QoDE}, and \ac{QoVE} of immersive \ac{XR} users (see Table~\ref{tab:lexicon} for the definitions of these metrics), \ac{DT}-enabled autonomous systems (e.g., autonomous vehicles), and the cognitive avatars, respectively. These metrics constitute the extrinsic reward (cost) of the telecom brain. In addition, this reward must be optimized along with the telecom brain's own \ac{QoNE} that guides its autonomous operations in planning its actions (see Table~\ref{tab:lexicon} and Fig.~\ref{Telecom_Brain}), where \ac{QoNE} constitutes the intrinsic reward of the telecom brain.
Henceforth, empowered with the ability to deal with unforeseen scenarios and generalize, this new breed of networks can potentially enable a new set of unprecedented experiences.

\begin{table*}
\centering
\caption{\small Lexicon of the index terms used in AGI-native wireless systems}
\label{tab:lexicon}
\begin{tabular}{p{5cm} p{12.2cm}}
\hline
\multicolumn{1}{c}{\textbf{Index Terms}} &
  \multicolumn{1}{c}{\textbf{Definition}} \\ \hline
\textbf{Reasoning} &
  The ability to draw conclusions from acquired knowledge and perform decision-making as exhibited by humans. \\ \hline
\textbf{Planning} &
  \begin{tabular}[c]{@{}l@{}}
\parbox[t]{12.2cm}{
\begin{itemize} \vspace{-0.2cm}
\item The process of anticipating the right actions to reach a specific goal.
\item The ability to think about the future.
\end{itemize}}
  \end{tabular}\\ \hline
\textbf{Common Sense} &
  \begin{tabular}[c]{@{}l@{}}
  \parbox[t]{12.2cm}{
  \begin{itemize} \vspace{-0.2cm}
\item A cognitive trait pertaining to the acquired background knowledge about the world that can be leveraged to deal with unfamiliar scenarios and reach reasonable conclusions. 
\item Common sense encompasses the general understanding of intuitive physics and intuitive psychology shared by (i.e., common to) humans. 
\item Common sense includes the ability to foresee the consequences of actions and identifying the probable, plausible, and impossible scenarios that can take place in the observed world.
  \end{itemize}}
  \end{tabular} \\ \hline
 \textbf{Analogical Reasoning} &
  The ability to relate and generalize between instances and scenarios to cope with unforeseen conditions. \\ \\ \hline
\textbf{Intuitive Physics} &
  The basic core skills of physical object manipulation and navigation. \\ \\ \hline
  \textbf{Vertical Generalizability} &
  The ability to generalize to out-of-distribution shifts in the data. \\ \\\hline
\textbf{Horizontal Generalizability} &
  The ability to generalize by leveraging the common knowledge about the world to deal with out-of-domain scenarios and corner cases. \\ \hline
\textbf{Perception} &
  The cognitive ability of acquiring an abstract representation of the state of the real-world constituents. \\ \\\hline
\textbf{AGI-Native Telecom Brain} &
  \begin{tabular}[c]{@{}l@{}}An AGI system encompassing an interconnected architecture of cognitive modules that can autonomously\\ control and orchestrate a wireless network (see Fig. 3).\end{tabular} \\ \hline
\textbf{Digital World} &
  \begin{tabular}[c]{@{}l@{}}An alternative synchronized digital reality that replicates the physical world and its constituents in the\\ form of real-time abstractions.\end{tabular} \\ \hline
\textbf{Cognitive Avatars} &
  \begin{tabular}[c]{@{}l@{}}Next-generation of \ac{AI}-driven avatars that can:\\ 
  \parbox[t]{12.2cm}{
  \begin{itemize} \vspace{-0.2cm}
 \item Learn how to map sensory and tracking inputs of XR users to movements and actions at the avatar. 
 \item Apply reasoning abilities to deduct the sensory feedback and actuations that are passed from the avatar to the XR user.
  \end{itemize}}\end{tabular} \\ \hline
\textbf{AGI-Augmented Digital Twins (DTs)} &
  \begin{tabular}[c]{@{}l@{}}Bidirectional operational AI models that can proactively optimize and configure the states of autonomous\\ systems with common sense feedback endowed through an AGI-native network.\end{tabular} \\ \hline
\textbf{Physical Assets} &
  Unidirectional (digital) simulation streams of massively sensed physical elements (e.g., Eiffel Tower). \\ \\\hline
\textbf{Humanoids} &
  Massively sensed matterless human representations that capture the human presence in the digital world. \\ \\\hline
  \textbf{Quality-of-Network Experience (QoNE)} &
  \begin{tabular}[c]{@{}l@{}} 
  \parbox[t]{12.2cm}{
  \begin{itemize} \vspace{-0.2cm}
 \item A novel metric that captures the quality of the autonomous operation in an \ac{AGI}-native network to independently achieve its own demands (e.g., guarantee sustainability, satisfy intents, etc.).
 \item An \ac{AGI}-native network with a \ac{QoNE} can pass through subjective experiences to learn and understand the world, similar to a human being's way of building up their knowledge from around themselves.
 \item It reflects the ``relief" or discomfort of the network from its own first-person point-of-view, and can possibly incorporate other conscious abilities~\cite{butlin2023consciousness}. 
  \end{itemize}}\end{tabular} \\ \hline
\textbf{Quality-of-Physical Experience (QoPE)} &
  \begin{tabular}[c]{@{}l@{}} A metric to assess the \ac{QoE} delivered to users in the physical world (e.g., \ac{XR} users). Its dimensions\\ include the rate, reliability, latency, etc. demanded by these users.  
  \end{tabular} \\\hline
  \textbf{Quality-of-Digital Experience (QoDE)} &
  \begin{tabular}[c]{@{}l@{}} A metric to assess the \ac{QoE} of DT-enabled autonomous agents (e.g., vehicles). Its dimensions could include \\satisfying trustworthiness of the \ac{DT} configurations, synchronization between the \ac{PT} and \ac{DT}, abiding by\\ guardrails, etc.  
  \end{tabular}  \\\hline
    \textbf{Quality-of-Virtual Experience (QoVE)} &
  \begin{tabular}[c]{@{}l@{}} A metric to assess the \ac{QoE} of cognitive avatars in the virtual world. Its dimensions could include\\ synchronization, fidelity, and accuracy in replicating the actions between the \ac{XR} and avatar.  
  \end{tabular}  \\\hline
\end{tabular}
\end{table*}

\vspace{-0.2cm}
\subsection{Contributions}
The main contribution of this paper is a holistic, forward-looking vision of \emph{\ac{AGI}-native wireless networks}, as articulated in Section~\ref{Vision}. This vision advocates for a disruptive paradigm shift in the traditional evolution of wireless networks that is asymptotically capped by the different physical limitations of conventional communication enablers, illustrated in Fig.~\ref{From_6GAdvanced_To_NextGeneration}. In particular, we envision that the metaverse will play a crucial role in pushing towards a new \ac{AI}-based revolution for networks. On the one hand, the metaverse with its digital world can enable a real-time perception of the physical world, which is an essential factor to enable \ac{AGI}-native networks. 
On the other hand, the metaverse brings forth novel use cases and applications such as cognitive avatars and \ac{AGI}-enabled \acp{DT} that require common sense abilities. 
To the best of our knowledge, this is the first work that explores the design of wireless systems with common sense as a pathway for the emergence of next-generation \ac{AGI}-native networks that bring forth a revolutionary set of capabilities, users, and experiences.
In summary, our key contributions include:
 \begin{itemize}
    \item We propose the \emph{first vision of an \ac{AGI}-native wireless system}, that promises the \emph{next revolution towards a new ``G" of networks}. Unlike its previous generations, we envision this network to be driven by a \emph{telecom brain architecture}, as shown in Fig.~\ref{Telecom_Brain}. In addition, we articulate how this \ac{AGI}-native network can bring forth a new generation of human-centric applications. In our vision, we advocate for the network to become the main entity that acquires common sense to reach \ac{AGI} levels, rather than the individual autonomous agents, as assumed in~\cite{LeCun2022OR}. In contrast, individual agents will become \emph{\ac{AGI}-augmented \ac{DT} applications} that are endowed with common sense from the \ac{AGI}-native network.

    \item We concretely define the pillars of common sense, as per Fig.~\ref{Common Sense}. Then, we investigate the crucial role that common sense plays in \ac{AGI}-native networks, highlighting that its integration into wireless networks can pave the way towards their fully autonomous operation. Here, we envision common sense to be the cornerstone for generalizable reasoning and planning abilities in networks. In particular, we foresee these abilities as the turning point for the network to truly deal with all possible corner cases that it can face along with its autonomous agents. 
    
    \item We envision that an \ac{AGI}-native network acquires common sense by building a hypothetical world model, rather than by learning from the real world itself, as assumed in~\cite{bariah2023ai}. To perceive this real world, we leverage the scalability of the network in capturing a synchronized digital world in the metaverse. Effectively, scaling the real world is facilitated by a decentralized digital world architecture over the network, that can bypass the need for the configurator module, a largely undefined element in the design of \ac{AGI} systems~\cite{LeCun2022OR}. Moreover, this scalable approach considers predicting the world in a concise, synchronized, and well coordinated manner, in contrast to randomly predicting individual futures at the level of individual agents.
    
    \item We propose capturing generalizable forms of abstractions of real-world elements by disentangling their semantic content through the fusion of \ac{ML} techniques like contrastive learning with causal representation learning. Subsequently, we show how this step is crucial to enable analogical reasoning between elements and effectively deal with unforeseen scenarios in \ac{AGI}-native networks. 

    \item We propose a first physics-based, causal world model in the literature. The proposed model merges the transparency of \acp{SCM} with the higher order vector representations of \ac{HD} computing~\cite{ThomasJAIR2024} to effectively manipulate abstractions in a brain-inspired fashion, while capturing rich causal relationships and representing the intuitive physics operations pertaining to common sense. To convert these abstract representations into the \ac{HD} space, we leverage the mathematical underpinnings of category theory~\cite{barr1990category} that can facilitate this transformation.

    \item To guide the autonomous decision-making operations of \ac{AGI}-native networks, we design two action-planning methods driven by intents and objectives. Inspired from neuroscience, we leverage concepts such as \ac{IIT}~\cite{tononi2008consciousness} for the design of the intent-driven and objective-driven planning strategies.
    

    \item We discuss how \ac{AGI}-native networks can provide resilient and synchronized avatar experiences to faithfully immerse and embody \ac{XR} users in the metaverse. Moreover, we show how an \ac{AGI}-native network can leverage its intuitive psychology capabilities pertaining to the \ac{ToM}~\cite{carlson2013theory} to enable brain-level metaverse experiences such as holographic teleportation.
    
    \item We conclude with a sequel of recommendations on how to evolve towards \ac{AGI}-native wireless networks in the beyond 6G era.
\end{itemize}

The rest of the paper is organized as follows. In Section III, we showcase how to design the telecom brain architecture of an \ac{AGI}-native network including its different modules shown in Fig.~\ref{Telecom_Brain}. In Section IV, we present the different use cases and experiences that an \ac{AGI}-native network can bring forth for humans and autonomous agents. Finally, we conclude with a set of recommendations that arise along the path to enable \ac{AGI}-native networks in Section V. A summary of this organization is shown in Fig.~\ref{ToC}.





\section{Designing the Telecom Brain: A Synergy of \ac{AGI} and the Metaverse}
\label{Telecomverse}

We begin our design of \ac{AGI}-native networks by shedding light on the path to construct the telecom brain of \ac{AGI}-native wireless systems. In particular, we provide a comprehensive discussion of the various modules (shown in Fig.~\ref{Telecom_Brain}) that appear in the design of the telecom brain. This includes sequential steps initiated with sensing the physical world and perceiving it in the form of abstractions over the network. This is followed by efficient representation of these abstractions as \ac{HD} vectors that can be manipulated with the intuitive physics operations of common sense. Accordingly, this will allow the network to infer the next plausible states and plan the corresponding network actions.

\subsection{Sensing: How can we capture the physical world over wireless networks?}
\label{Sensing}

To establish a real-time, digital replica of the physical world over the network, we first must capture the real-time sensory data of the different physical constituents before feeding them to the perception module of Fig.~\ref{Telecom_Brain}. This includes the data collected/generated by \ac{DT}-enabled autonomous agents, humans, and physical assets (e.g., Eiffel Tower, Statue of Liberty, etc). To facilitate this process, it is necessary to integrate diverse sensing technologies in 6G and beyond networks that can range from joint sensing and communications (e.g., in the sub-\ac{THz} bands~\cite{chaccour2023joint}) to wireless sensor networks, along with other sensing infrastructure (e.g., vehicle-to-everything (V2X)).
This integration can help create a collective view of the physical world from multiple angles, close any sensing gaps, and ensure a faithful replication process.

Nevertheless, attempting to replicate the real-world in a centralized, cloud-based manner over the network can result in significant communication delays that can jeopardize the synchronization between the physical and digital worlds. To address this issue, a decentralized, edge-enabled digital world is necessary. However, this requires establishing effective modeling techniques of the physical world that can capture the states of its constituents (e.g., assets, autonomous agents, etc.), while allowing efficient decomposition of the replication process over the edge. For instance, these techniques must consider the different computing and communication resources at each network edge to preserve the maximum synchronization between the physical and digital counterparts of these constituents. In our previous work~\cite{hashash2022towards}, we have demonstrated that the optimal approach to achieving this digital reality involves decentralizing the digital world into so-called ``\emph{sub-metaverses}'' -- digital counterparts of physical world spaces. These sub-metaverses are orchestrated, along with their components (e.g., assets, \acp{DT}, etc.) at the wireless edge to preserve the highest levels of synchronization.
On the one hand, this orchestration aims to conserve the \emph{inter-synchronization} between the physical and digital worlds and ensure upmost levels of real-time replication. On the other hand, this is complemented by minimizing the delay gap between the sub-metaverses so as to preserve the \emph{intra-synchronization} between the distributed parts of the digital world. In this case, the digital world can conserve its overall homogeneity as a collective structure. For \ac{AGI}-native wireless systems, this synchronization is necessary as it will allow the telecom brain to predict concise future states that truly reflect the real-state of the physical world. This, in turn, can allow taking the proper network actions and enabling reliable coordination of the \ac{DT}-enabled autonomous agents. This solution differs from the approach outlined in~\cite{LeCun2022OR} that allows individual agents to predict the future states individually, which lacks synchronization and coordination between agents in the prediction process and can possibly lead to chaos in the physical world, as explained in Section~\ref{State_of_the_art}.

	\begin{figure}
		\begin{minipage}{0.99\linewidth}
			\centering
			\includegraphics[scale=0.4]{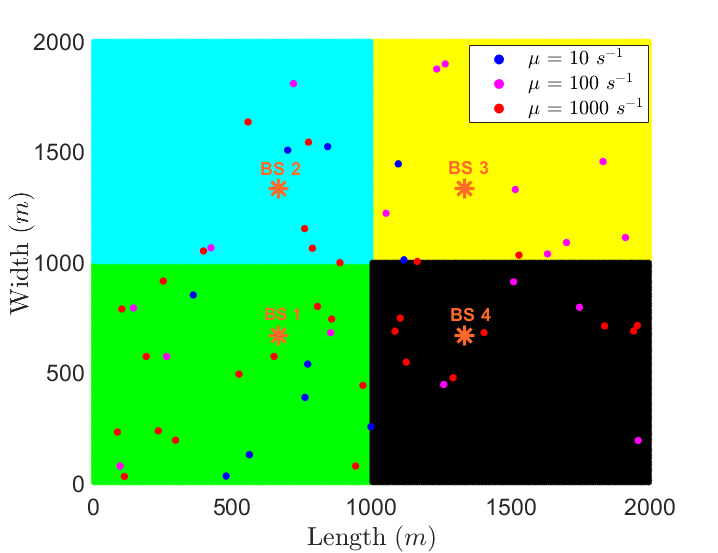}
			\subcaption{}    \label{SNR}
		\end{minipage}
		\begin{minipage}{0.99\linewidth}
			\centering
			\includegraphics[scale=0.4]{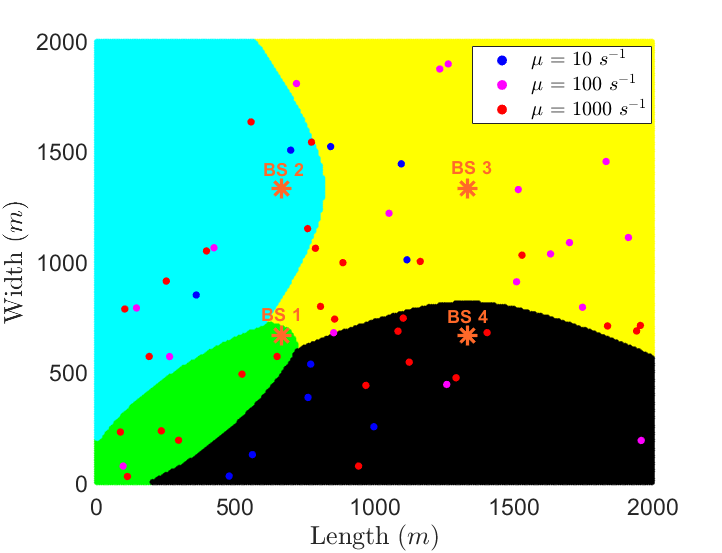}
			\subcaption{}    \label{OT}
		\end{minipage}
		\caption{\small{Region partitioning and DT association according to the (a) SNR method and (b) proposed optimal transport method}~\cite{hashash2022towards}.}  \label{partitioning}
		\vspace{-0.1cm}
        \label{optimal transport}
	\end{figure}

Thus, as per our work in~\cite{hashash2022towards}, we can model the physical world through a probabilistic approach with a continuous distribution of sensors that capture the states of physical objects and assets. In this model, we incorporate two essential metrics: a) volumetric sensing density ($\mathrm{bps/m^3}$), and b) spatial distribution of sensors. Here, the volumetric sensing density represents the amount of data being produced from each spatial position in the physical world. Moreover, the spatial distribution describes the likelihood of the sensors being located around the 3D assets in the physical world. Thus, it is the fusion of both metrics that provides a reflection of the effective data flowing from the physical world. This perspective is aligned with the view that future wireless systems can be seen as \emph{massive sensing or imaging devices} and not just mere communication systems~\cite{saad2019vision}.

In addition, our proposed solution in~\cite{hashash2022towards} provides an effective technique for distributing the digital world through an iterative algorithm that can guarantee the maximum synchronization between the physical and digital worlds. As shown in Fig.~\ref{partitioning} from~\cite{hashash2022towards}, our solution can provide a non-uniform distribution and association of the physical world and its \acp{PT}, as sub-metaverses with their corresponding \acp{DT}, respectively, at the edge. Unlike the uniform \ac{SNR}-based association scheme, our distribution method also considers the synchronization intensity $\mu$ which represents the tolerable threshold for the different \ac{DT} applications to replicate their \acp{PT}, and the computing and communication resources associated to each edge. Hence, our method provides a comprehensive solution to determine the optimal association of sub-metaverses and \acp{DT} at the edge.
In fact, our results show that this non-uniform distribution appears due to an optimal tradeoff between sub-metaverses and \acp{DT} associations at the edge that can ensure the highest inter-synchronization is achieved and the synchronization intensity requirements of \ac{DT} applications are met. 

\textbf{Open Problems.} Although sensing the physical world over wireless networks has been instantiated with prior works such as~\cite{hashash2022towards}, there remain open problems that require further investigation, such as:
\begin{itemize}
    \item \textbf{Replicating the \ac{RAN} and core:} Replicating the physical world is not just exclusive to mirroring the wireless users, but it also encompasses a replica of the network. That said, the \ac{RAN} and core components must be replicated in a distributed manner over the network to ensure the scalability in providing a synchronized twin of the wireless system. In this case, it is necessary to investigate how to distribute the replication of \ac{RAN} elements close to the network edge to preserve their synchronization with the physical counterparts, while hierarchically replicating the other components of the network as we move closer to the core. Hence, it is challenging to set the boundaries and designate the precise orchestration of the \ac{RAN}-\ac{DT} and core-\ac{DT} over an \ac{AGI}-native network.
    
    \item \textbf{Designing efficient collaborative sensing schemes:} Upon replicating the physical world, multiple modes of sensing data are gathered to describe the physical elements (e.g., LiDAR, \ac{IoT} sensors, etc). Clearly, sensing data may include redundant information from multiple modalities. Hence, it is necessary to design collaborative sensing frameworks that can combine the distributed sensing inputs to efficiently utilize the communication resources. That said, it is also necessary to consider methods such as the value of information to reduce the rounds of sensing updates on the network.

    \item \textbf{Joint sensing and communications:} Naturally, for creating a massive replica of the world, it will be important to design joint sensing and communication schemes that can exploit emerging wireless technologies (like \ac{THz} bands) to get an image of the real world and create parts of the digital world. Indeed, using the communication signal to perform sensing and imaging is an important open problem here. Hence, the design of low-cost, effective joint sensing and communication systems is an interesting direction for research in this component of our \ac{AGI} vision.

\end{itemize}
 

After distributing the sensing process of the world over the network, the following step is to perceive the world in the form of real-time abstract representations. In fact, replicating the real world into its digital counterpart is facilitated by this perception process.
Creating such representations is essential for determining the plausible future states of the world and its elements (e.g., assets, autonomous agents, etc.). Moreover, these abstract representations will be the key to carry out analogical reasoning and generalizing in unforeseen scenarios. Next, we will show how such generalizable abstractions can be uncovered through disentangling the ``semantic representations" that exist in the sensory data coming from the physical world.

\subsection{Perception: From data to representations}
\label{perception}

Perception is one of the primary cognitive abilities that should exist at the frontier of the telecom brain, as observed from our proposed framework in Fig.~\ref{Telecom_Brain}. In essence, perception is the cognitive ability that allows the computation of an abstract representation of a real-world element. An \emph{abstract representation} refers to a simplified structure of an element that captures its essential features while omitting irrelevant details.
Such representations can be created by simply embedding the different meanings, properties, and the functions of real-world elements, in an abstract form~\cite{guo2022deep}. Nevertheless, this simple approach to build abstract representations can be insufficient for unleashing the full capabilities of an \ac{AGI}-native network. 
An \ac{AGI}-native network must further exploit these abstractions to make future predictions and analogy with the unfamiliar elements it can encounter in the physical world.
Therefore, to facilitate these functionalities, the representations in an \ac{AGI}-native network must exhibit certain characteristics beyond just abstraction.
In particular, the telecom brain in an \ac{AGI}-native network must carefully encode the abstract forms into representations that i) sufficiently hold their essential characteristics, ii) uncover the relations with other representations, and iii) maintain a common generalizable form that allows carrying out analogical reasoning in unforeseen scenarios. 



On the path towards capturing such representations from the physical world, the network must start by understanding the contextual meaning of the real-world elements from their sensory data. In other words, the telecom brain must unravel the \emph{semantic content elements}~\cite{chaccour2022less} pertaining to each physical element. Here, the ``semantic" aspect broadly refers to the meaning inside the data. As such, a semantic content element therefore refers to the meaning of a physical element that is present within the captured data points of this element.
This can help abstract the essential features of each physical element to further encode them into corresponding representations. In fact, the process of abstraction and representation is the cornerstone of replicating the physical world into its corresponding version of the digital world. However, encoding these representations cannot take place by embedding the underlying meaning in the semantic content element in a \emph{minimally sufficient} manner, as is the case in the field of semantic communications~\cite{chaccour2022less}. In contrast, encoding in an \ac{AGI}-native network requires an advanced level of representation complexity to express the aforementioned requirements about maintaining the essential characteristics, uncovering the relations, and remaining generalizable.
On the one hand, these representations should unfold their inherent causal relations, to faithfully predict future states of the world and accurately plan the actions of the telecom brain. This is expected to minimize the error between the predicted abstractions and the real-world. This error minimization is also expected to drive lower costs\footnote{Here, the cost (or reward) is captured by the \ac{QoPE}, \ac{QoDE}, \ac{QoVE}, and \ac{QoNE}, shown in Fig.~\ref{Telecom_Brain}.} (or higher rewards) for the telecom brain.
On the other hand, as the complexity of these representations increases, the representations start to overfit the semantic content. Therefore, the representations must conserve a generalizable form for analogical reasoning. This generalizable form makes representations relatable to identify unforeseen elements.

However, before discussing how abstract representations should be designed in \ac{AGI}-native wireless systems, we must clarify the distinctions and synergies between our proposed methodology and that of semantic communications \cite{chaccour2022less, lu2023semantics}. Similar to semantic communication systems that consider capturing representations of the different content elements in the data, we leverage semantic representations to capture the real-world elements and identify their real-time status from the sensor data. Nevertheless, there exist key fundamental differences: 
\begin{itemize}
    \item In general, semantic communications leverage abstractions to enhance link level efficiency and minimize communication resources. However, \ac{AGI}-native networks must exploit abstractions to create a complete, and holistic understanding of both the physical world and the network. Hence, abstractions in \ac{AGI}-native networks will play a different role than the one they play in semantic communications. They will also have to possess other distinct properties.
    Indeed, abstract representations have a role central to the different modules of the telecom brain architecture shown in Fig.~\ref{Telecom_Brain}. Hence, this role extends beyond the role of transmitting meaning, that is the focus in semantic communications, to computing and controlling the physical world. This is particularly reflected in their aforementioned characteristics that need to i) hold their essential characteristics, ii) uncover the essential relations, and iii) maintain a generalizable form.
  

    \item In essence, semantic communication exclusively deals with the reconstruction of a \ac{Tx}'s message at the \ac{Rx} side. While that may be possible by exploiting minimally sufficient representations of real-world elements, leveraging those same representations to predict the future states of these elements is inadequate for \ac{AGI}-native networks. That is, minimally sufficient representations may fall short in uncovering the entire causal relations between representations. Consequently, this will directly degrade the faithful predictions of the telecom brain along with the anticipated rewards gained from its actions. Finally, it is important to note that \ac{AGI}-native networks use abstract representations to deal with problems beyond just \ac{Rx} reconstruction.
 \end{itemize}

Now that we have distinctly pinpointed the uniqueness of abstractions in \ac{AGI}-native networks, we describe how we can capture the semantic content elements from the sensing data.
This procedure is illustrated in Fig.~\ref{Perception_Disentangling_Causal_Generalizable} and explained in the following two steps: 

\begin{figure*}
	\centering
	\includegraphics[width=0.95\linewidth]{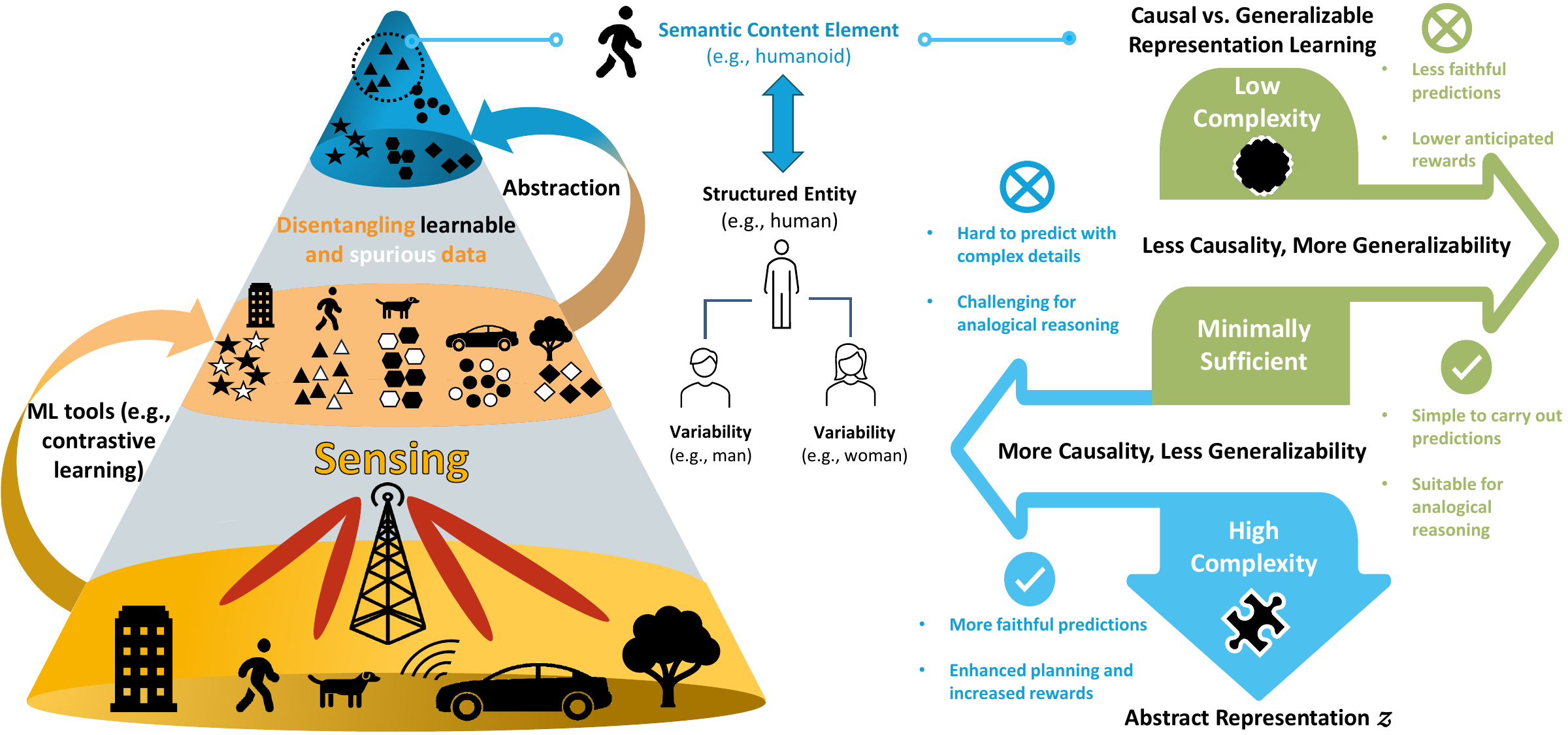}
	\caption{\centering \small{Illustrative figure showcasing the process of perception in the telecom brain that includes: i) Disentangling of learnable and spurious data to capture the semantic content elements contributing to the abstractions of real-world elements and, ii) Encoding these abstractions into representations by optimizing their complexity levels to balance between causality and generalizability.}}
	\label{Perception_Disentangling_Causal_Generalizable}
\end{figure*}

\subsubsection{Disentangling learnable and spurious data} The first step to capture abstractions is to disentangle well-structured points within the data from those that are weakly structured. On the one hand, well-structured data points are those that are rich in meaning and express a consistent format. On the other hand, weakly structured data points pertain to spurious or even very specific instances that lack any format, and do not necessarily map to the essential characteristics of a semantic content element. Subsequently, this step aims to categorize the data into two streams: a) learnable and b) spurious. Effectively, it is the learnable data that will contribute to the abstract representations in \ac{AGI}-native networks.
In this case, the spurious data can be neglected as it lacks any underlying structure and does not represent specialized semantic attributes of the elements that can further contribute to the representation\footnote{An abstract representation is an invariant, well-defined, reduced structure of an element. Therefore, the contributions of spurious data points, that lack these crucial features, to abstractions is minimal and can be further neglected. This is a key difference from semantic communications that still needs to transmit this type of data stream to reconstruct the elements back at the \ac{Rx}.}. This is the first step to abstract and differentiate between the elements in the data.
Here, one promising technique that can facilitate extracting these structured representations from the data is contrastive learning~\cite{chen2020simple}. As shown in Fig.~\ref{Perception_Disentangling_Causal_Generalizable}, the telecom brain can disentangle the learnable and spurious data of each real-world element through contrastive learning. In fact, our work in~\cite{chaccour2022disentangling} demonstrated how we can properly disentangle and structure the data, to efficiently transmit it over wireless networks by adopting a semantic language of these representations. Also, it is worthwhile noting that multi-modal sensing requires the fusion of learnable data structures from each modality into a single representation.

Here, we can also note that once we have acquired structured, learnable data, the \ac{AGI}-native network can further decompose the data into two components: \emph{i)~structured entity} and, \emph{ii)~variability}~\cite{chaccour2022less}. The structured entity represents the general form of the representation that is shared among different real-world elements, while the variability refers to specific information that is exclusive to the specific element among others. 
For instance, a learnable data related to a humanoid (see Table~\ref{tab:lexicon} for definition) can be decomposed into a structured entity of a human and the specific data points that differentiate between a man and woman are the those of the variability, as shown in Fig.~\ref{Perception_Disentangling_Causal_Generalizable}.
As such, maintaining a generalizable representation will depend on the structured entity that can be shared with similar real-world elements.

Thus far, we have captured the learnable structures of the elements that will contribute to their abstract representations. The next necessary step is to encode these abstract forms into the corresponding representations. As mentioned earlier, these representations must maintain a certain level of complexity, beyond the minimalism of semantic communications, that can uncover rich causal relations while preserving a generalizable structure for similar real-world elements. 
In other words, the telecom brain must further optimize the complexity level of the representations to balance between a) capturing the causal relations necessary to have faithful predictions and accurate planning, and b) maintaining a generalizable structure of representations for analogical reasoning. 
Next, we will show how to optimize the complexity of these representations so as to balance between causality and generalizability.

\subsubsection{Causal vs. generalizable representation learning for abstractions}
Representing the captured semantic content elements in abstract form requires encoding them into proper symbols, as shown in Fig.~\ref{Perception_Disentangling_Causal_Generalizable}. Here, these symbols are related through causal relationships. 
Hence, these symbols must be designed to facilitate the discovery of the rich causal relations between them by the telecom brain. With the proper design of such symbols, the telecom brain can faithfully predict the future states of the real-world elements and accurately plan the actions of the network.

Nevertheless, uncovering the majority of causal relations between the representations requires encoding them into detailed, complex symbols. Hence, it will be challenging for the telecom brain to faithfully predict the future states of the elements with such granular details in the symbols. Moreover, the additional degrees of complexity introduced may not drive in similar advancements in terms of reward (or cost) for the actions of the telecom brain. 
Moreover, having more complex symbols will favor the variability in the representations over the structure. This is due to the fact that including more granular details in the symbols will reduce the similarity between the symbols of similar real-world elements. Consequently, this can hinder effective analogical reasoning.
For instance, encoding the semantic content element of a humanoid in a complex form will capture each of its minute details (see the low complexity and high complexity icons in Fig.~\ref{Perception_Disentangling_Causal_Generalizable}). Subsequently, such encoding will capture much more rich and exact causal relations between the symbols corresponding to the different world elements.

Therefore, the telecom brain must optimize the complexity level at which it encodes the semantic content elements, so as to balance between the expected rewards, prediction error, and similarity of abstractions. Moreover, we should also note that the complexity herein is related to the level of details to which we encode the semantic content elements. This is different from the complexity of a semantic language that characterizes the difficulty of identifying and learning the semantic content elements~\cite{chaccour2022less}.

Furthermore, it is crucial for the telecom brain to maintain an \ac{SCM} to represent the causal relations between these symbols. Here, one standard approach to represent an \ac{SCM} can be in the form of a \ac{DAG} that captures the causal relationships between the different symbols of the semantic content elements. This \ac{SCM} structure can be generally defined as follows.

\begin{definition}
\label{SCM-Defintion}
An \emph{\ac{SCM}} is a collection of elements $<\mU, \mV, \mF, P(\bmU)>$, where $\mV$ and $\mU$ represent endogenous and exogenous variables, respectively. For an \ac{AGI}-native network, $\mV$ represents the encoded symbols of semantic content elements and $\mU$ represents latent random variables.
These exogenous variables captured in a vector (or matrix) $\bmU$ represent the stochastic and random side of the world that makes it partially predictable.
For any index $i$, any endogenous variable $\bmv_i \subset \mV$ can be determined by the structural functions $f_i \in \mF$ and modeled as $f_i(\textrm{PA}_i, \bmu_i)$, where $\textrm{PA}_i \subset \mV$ (in the graph) are the sets of its parents and  $\bmu_i \subset \mU$ are exogenous inputs. 
The exogenous distribution $P(\bmu_i)$ determines the values of $\bmu_i$, and thus the distribution of endogenous variables $\mV.$

\end{definition}


The network seeks to optimize the complexity level of the representations. This optimal complexity can balance between the needed causality and the level of generalization to perfectly encode the semantic content elements in an \ac{AGI}-native network. Clearly, this optimization will impact the structure of the \ac{SCM} and the causal discovery process between the symbols~\cite{pearl2009causal}. 
Thus, we seek to find the optimal symbol representation that minimizes: i) the cost (maximizes the reward) of the telecom brain, ii) the prediction error between the future representation and the captured real-world outcome, and iii) the generalization mismatch between similar real-world symbols.  
This can be formulated as follows:
\begin{equation}
\label{complexity}
    \left[\bmz^{*}\right] =  \argmax\limits_{\bmz} \quad \underbrace{\alpha \mathbb{L}(\bmz)}_{\mathrm{Reward}} -  \underbrace{\beta \mathbb{G}(\bmz, \tilde{\bmz})}_{\substack{\mathrm{Prediction} \\ \mathrm{Error}}} -  \hspace{-0.2cm} \underbrace{\gamma \mathbb{J}(\bmz)}_{\substack{\mathrm{Generalization} \\ \mathrm{Error}}},
\end{equation}
where $\bmz \in \mV$ is the symbol representation in the \ac{SCM} and $\tilde{\bmz}$ is the real-world outcome of the representation.
Moreover, $\mathbb{L}(\bmz)$ is the reward (cost) function of the telecom brain that is captured through intrinsic cost \ac{QoNE} and extrinsic costs of \ac{QoPE}, \ac{QoDE}, and \ac{QoVE}. In addition,  $\mathbb{G}(\bmz, \tilde{\bmz})$ is the prediction error between the representations of the predicted future $\bmz$ and the real-world outcome $\tilde{\bmz}$. $\mathbb{J}(\bmz)$ is the generalization error that captures the mismatch in the structure between representations of similar real-world elements.
Furthermore, $\alpha$, $\beta$, and $\gamma$ are hyper-parameters determined by the telecom brain to control the tradeoff.
In essence, finding $\bmz^{*}$ can be seen as the equivalent of  acquiring rich and generalizable symbol representations that are closely related to neighboring symbols in a semantic space and resilient to the semantic noise distortion from other representations.


\textbf{Open Problems.} In the context of perceiving real-world elements, there is a number of existing challenges that should be particularly addressed to enable the full functionality of the perception module in the telecom brain: 
\begin{itemize}

    \item \textbf{Dimension collapse in contrastive learning:} \ac{AI}/\ac{ML} techniques such as constrastive learning disentangle the learnable and spurious data streams. Prior to that, such methods distinguish between the different content elements in the sensing data. Although constrastive learning considers a high-dimensional embedding space to differentiate between the semantic content elements, it may still face the problem of dimensional collapse~\cite{jing2021understanding}. That is, data points of different semantic content elements can become indistinguishable or collapse into a lower-dimensional space. For example, this can happen as a result of the training loss in contrastive learning that might encourage learning how to differentiate between the elements, while missing to capture their high-dimensional structure present in the data. Here, there is a need for novel approaches to contrastive learning that can overcome this issue. Alternatively, one can investigate other approaches beyond contrastive learning, such as energy-based models~\cite{kim2022energy}. 

    \item \textbf{Symbol representation of \acp{DT}:} It is imperative to differentiate the symbol representations of autonomous applications from other physical world elements. This is because these applications are enabled by \acp{DT} that are essentially \ac{AI} models driven by sensory data from the world. Evidently, one crucial angle of perception is that the \acp{DT} of these applications must be integrated by the telecom brain as abstract representations into the digital world. Hence, these representations take part in a \emph{slow} thinking process for planning the optimal actions of the telecom brain. Simultaneously, the \acp{DT} should respond \emph{fast} to the large amounts of sensory data to synchronize with the \acp{PT}. Thus, in response to these integral roles of \acp{DT}, they must be modeled through a hybrid approach that can capture both fast and slow modes of thinking~\cite{kahneman2011thinking}. One solution can be to define \acp{DT} as \emph{neuro-symbolic \ac{AI}} systems that can capture both of the aforementioned aspects~\cite{garcez2023neurosymbolic}. This is because neuro-symbolic \ac{AI} is an approach that merges the rule-based and logic capabilities offered by symbolic \ac{AI} to represent knowledge and reason, with \ac{NN}-based learning that excels in detecting patterns within data. This hybrid approach can strengthen \ac{AI} systems with both approaches. 
    On the one hand, these symbols can play a role in the slow thinking and reasoning process of the telecom brain. On the other hand, leveraging \acp{NN} can provide a swift response and fast actions by the \ac{DT}.

    \item \textbf{Perceiving the network:} Evidently, the network itself is perceived in terms of the \ac{RAN}-\ac{DT} and core-\ac{DT}. Hence, this will require abstracting its elements (e.g., \ac{RAN}, core, channels, etc.) and functionalities (e.g., beamforming, resources, etc). Although there has been some recent works that consider semantics and representations of a communication network to enhance its efficiency (e.g., for \ac{CSI} feedback~\cite{cao2023adaptive}), a key open challenge is the need for new approaches to abstract the network and build the core-DT and \ac{RAN}-DT in terms of representations that can be exploited to initiate actions such as beam steering (as we will discuss in Section~\ref{action-planning}).  As such, this will require encoding the abstractions of the network into proper symbols as defined in~\eqref{complexity}. In this case, an important open problem is to adequately determine the complexity of the encoded symbols of the network. This due to the fact that this complexity will depend on discovering how the actions of the network are related to one another as well as to real-world elements.

    \item \textbf{Categorizing representations of similar instances:} 
    Capturing abstract representations of real-world elements is a dynamic real-time process. Hence, it is necessary for the telecom brain to identify the abstracted element as a new or a previously identified element.
    This requires the telecom brain to categorize the symbols that largely hold the same semantics into a single space dedicated to the same representation. For example, if the telecom brain identifies a humanoid with some new additional variabilities, it must consider this previously identified humanoid and relate it directly to its acquired representation. Indeed, real-world elements cannot be identified as new instances once a slight modification occurs to their representation.   
    To address this problem, a promising approach can be to explore the concept of \emph{persistent homology} from the field of topology \cite{Hatcher2002}. In fact, our prior work~\cite{ChristoJSAITArxiv2022} discusses the use of  persistent homology to design the semantic space inherent for each representation. Thus, one can consider various semantic content elements within the data to form a simplicial complex. Accordingly, a simplicial complex comprises a finite assembly of simplices, such that each $k$-dimensional simplex being an affine combination of $k+1$ semantic content elements. Techniques such as filtration within persistent homology offer rigorous capabilities in organizing disparate semantic content elements. Hence, these elements can be categorized and clustered according to their similarity between representations or the requisite level of abstraction.

    \item \textbf{Migration of \acp{DT} and its effect on \acp{SCM}:} In general, the \acp{PT} can move around the physical world. This would require the \ac{DT} to transition from one edge to the other to remain synchronized to the \ac{PT}. This can lead to multiple challenges. First, the \acp{DT} can be present over one edge, however, they can have causal relations with elements from another edge. Hence, it is challenging to determine how an \ac{SCM} can be formed to model this relation between elements from different edges. In addition, as one \ac{DT} migrates to a new edge, a key open question would be to determine how the \ac{SCM} that establishes the world model at this edge can be efficiently updated to include the migrating \ac{DT}.

\end{itemize}

Given that we have now perceived the real-world elements in terms of abstract representations, it is crucial to manipulate those symbols of abstract nature with the principles of intuitive physics that govern the real-world as well as common sense. In addition, it is likewise important to perform analogical reasoning with such abstract symbols. Therefore, it is necessary to \emph{ground these symbols} within a world model that is compatible with the nature of physics, facilitating their manipulation and enabling analogical reasoning.

To achieve the above goal, we propose to transform these representations from symbols to vectors in an \ac{HD} space. In essence, leveraging \ac{HD} vectors and spaces~\cite{ThomasJAIR2024} allows the telecom brain to efficiently manipulate its abstract representations with intuitive physics operations. This approach will enable the telecom brain to predict the plausible future states of the world as described in Fig.~\ref{Common Sense}, and plan the optimal actions of the \ac{AGI}-native network accordingly. Moreover, the structure of vectors in an \ac{HD} space foster the abilities of the telecom brain to perform efficient analogical reasoning. However, there is a need to find a mapping technique that can transform these representations from the symbol space to the vector space. For that purpose, we propose the use of \emph{category theory}~\cite{barr1990category}, from the fields of abstract algebra and topology, as a rigorous tool that can facilitate this mapping from the category of symbols to vectors, as described next.

\subsection{World model: Causality meets \ac{HD} computing}
\label{HD causal world model} 
The world model is one of the most intricate components of the \ac{AGI}-native brain architecture. Its responsibilities encompass two strategic purposes that are the cornerstone of common sense. Firstly, it must estimate the information that was missed upon perceiving the elements from the real-world, thereby enabling the prediction of the natural progression of real-world events. Secondly, it plays a crucial role in simulating the plausible future states of the world that can result from endogenous and exogenous contributions. Hence, without a physically-grounded world model that can manipulate symbols to make analogy and predict, there is no possibility to acquire general intelligence. 

The design of a world model has been already touted in the \ac{AI} literature such as~\cite{LeCun2022OR} and  \cite{li2020causal} as the cornerstone of \ac{AGI} and its derivatives. However, remarkably, to date, there are no world models that can permit autonomous agents to manipulate representations so that they can predict the future and perform analogy between real-world elements. 
Interestingly, because the telecom brain has access to a scalable replica of the world, it provides the missing link needed to overcome these persistent challenges and bring in a new design of world models.

Although the idea of a causal world model as an \ac{SCM} between real-world variables has been proposed previously e.g.,~in~\cite{thomas2023causal} and~\cite{zevcevic2021causal}, the design of a world model that permits physical interactions (i.e., object manipulation and navigation) with abstract representations is still largely under-explored. In fact, the design of such a world model should be influenced by the cognitive mechanism by which the brain performs mental computations over its representations. This is accompanied by the need to address two crucial limitations of \acp{SCM}: 
\begin{itemize}
    \item Difficulty in representing symbols with a multitude of distinct features as a single variable in an \ac{SCM}.
    \item Limited scalability in modeling the causal relations between the features of real-world elements in an \ac{SCM}.
\end{itemize} 

To address these challenges, in our \ac{AGI}-native wireless system, we propose to couple causal world models with \emph{\ac{HD} computing}~\cite{KanervaCS2009}. The inspiration for \ac{HD} computing comes in part from the study of human cognition that addresses how the brain processes information and perceives the world with all of its different variations. In particular, the perception of information in the brain is represented by the activation of numerous neurons, that fire in a certain sequence, to signal a specific concept or element. Hence, the same neurons, when activated differently, can represent completely different elements. Therefore, information is represented as a combination of activated neurons, sharing the same basis. Analogously, the key to \ac{HD} computing is representing the information of a certain element or concept as a combination of feature vectors in an \ac{HD} space. In essence, these feature vectors essentially represent the different characteristics of real-world elements.
Thus, this notion of \ac{HD} vectors is compatible with the representations captured in the perception module of our \ac{AGI}-native wireless system, which in turn, are a composite of different key features that make up a representation. While \ac{HD} computing has been used in the \ac{AI} literature previously, e.g.,~\cite{morris2019comphd} and~\cite{hassan2021hyper}, those prior works are limited to certain applications such as lightweight classification in resource constrained systems.
Nevertheless, in general, these works do not account for the intuitive physics operations and analogical reasoning of common sense. In contrast, we consider the vectorial nature of \ac{HD} as an enabler to manipulate the abstract representations with the actions of the telecom brain and facilitate the interaction between different representations.

Thus, to transform the representations from the symbol space to the desired vector space, we propose leveraging category theory, building on our prior work~\cite{ChristoTWCArxiv2022}. Category theory (see \cite[Appendix A]{ChristoTWCArxiv2022} for category theory preliminaries) deals with interrelated abstract representations and provides certain algebraic structural properties, facilitating the grouping of elements within a category and capturing the relations between the elements, as well as between the different categories. We next define basic terminologies in category theory that are useful for our purpose.

\begin{definition}
A \emph{category} is defined as a mathematical structure that comprises a set of objects and morphisms.
\end{definition}
\begin{definition}
A \emph{morphism} is a directed relation from object $\boldsymbol{w}$ to object $\bmy$ in a category $\Psi$ that indicates whether $\boldsymbol{w}$ can cause $\bmy$ or $\bmy$ is a property of $\boldsymbol{w}$.
\end{definition}

\begin{definition}
A \emph{functor} $F$ is a mathematical object that maps between categories in a way that preserves the structure of those categories.
\end{definition}

\begin{figure}
	\centering
	\includegraphics[width=\linewidth]{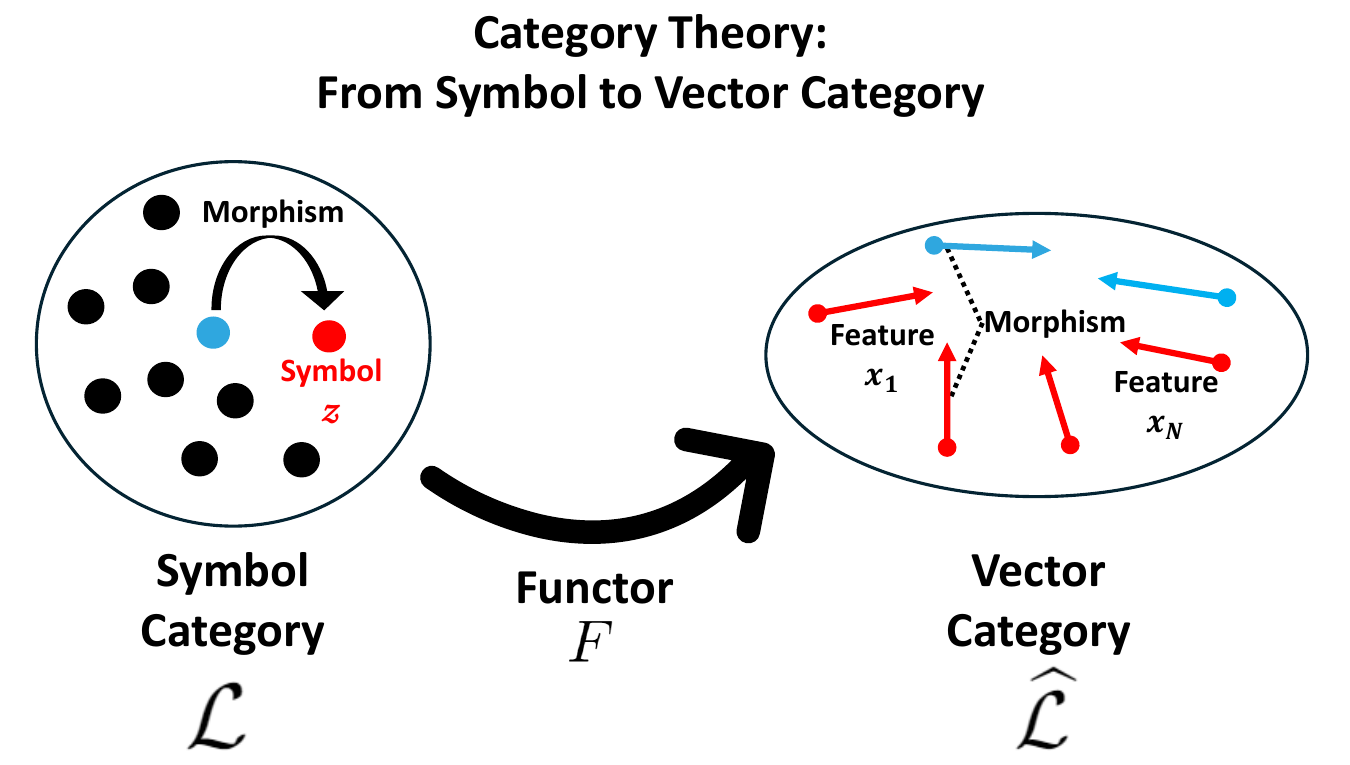}
	\caption{\centering \small{Illustrative figure showcasing the use of category theory in the transformation from the symbol space to the vector space. Here, every symbol $\bmz$ in the symbol category $\mL$ is decomposed into $N$ feature vectors  $\bmx_1, \ldots, \bmx_N$ in the vector category $\mwL$ through functor $F$. In addition, morphisms between symbols in $\mL$ are transformed into morphisms between feature vectors in $\mwL$.}}
	\label{category theory}
 \vspace{-0.3cm}
\end{figure}

As shown in Fig.~\ref{category theory}, the symbols in the semantic space form a \emph{symbol category} $\mL$ and the extracted causal relations between these symbols can be represented as morphisms. To enable this transformation of space, the symbol category $\mL$ is mapped via a functor $F$ into a \emph{vector category} $\mwL$, where $F:\mL \rightarrow \mwL$. Here, the resulting category $ \mwL$ is formed of vectors that represent the features of these symbols. In particular, the symbol $\bmz$ is decomposed into its features vectors $\bmx_1, \ldots, \bmx_N$, as shown in Fig.~\ref{category theory}.
Evidently, while an \ac{AGI}-native telecom brain can identify the set of symbol representations and their causal relations, it still needs to identify the proper functor $F$ that can facilitate the mapping from the symbol to vector category; which is an interesting open problem. 
After transforming the abstract representation $\bmz$ from the symbol into the vector space, the telecom brain must manipulate these abstractions to predict the future states of the world. Hence, the telecom brain must build \ac{HD} representations from the objects in the vector space.

Toward this end, we explain the foundations of mathematical operations in \ac{HD} computing that can be leveraged to transform the symbols encoded by the telecom brain into the \ac{HD} space. In other words, we explain how abstract representations can be expressed as \ac{HD} vectors.
This solution, based on \ac{HD} computing, provides a scalable and efficient approach to represent elements with numerous features, while capturing the causality between their features. Moreover, through the use of vectors, this approach can provide a foundation for handling basic physics operations (e.g., addition, subtraction, translation, etc.) and object manipulation by altering the entries of the vector separately. As such, these operations are essential for common sense in \ac{AGI}-native networks.
In particular, these operations are necessary for the telecom brain because it has to manipulate these vectors to predict the future states, and reason over them, to plan its actions. Hence, for the proposed \ac{AGI}-native wireless systems, a world model is concretely defined as an \ac{HD} space of vectors that represent the symbols of the telecom brain. These \ac{HD} vectors are further connected with an \ac{SCM} between their entries to model the causal relations between their features. Thus, this process culminates in an \ac{HD}-enabled \ac{SCM} of the world. Subsequently, we explain the facets of \emph{\ac{HD} causal world models} that build on \acp{SCM} as their underlying basis.

\begin{figure}
	\centering
	\includegraphics[width=\linewidth]{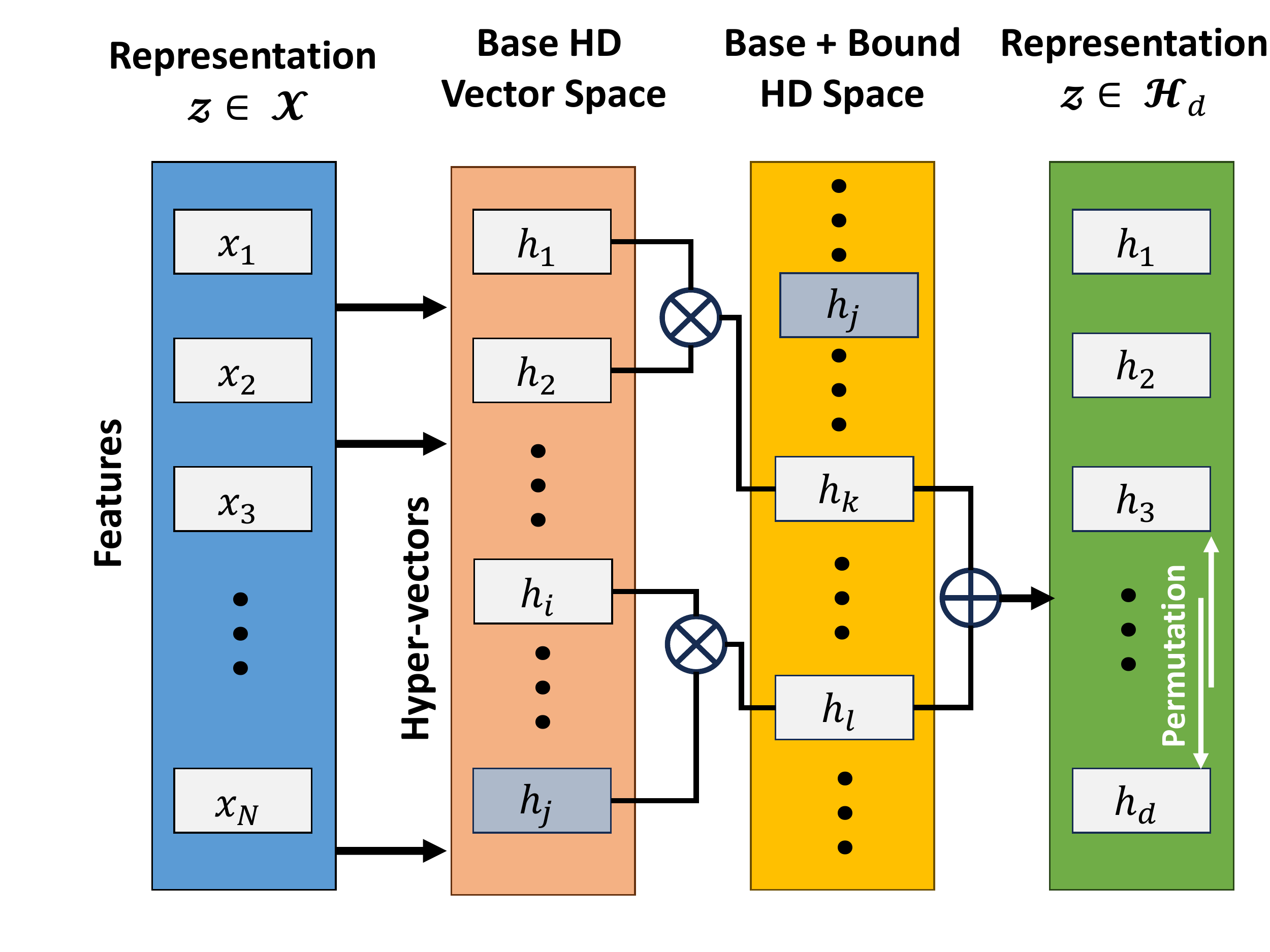}
	\caption{\centering \small{Mathematical operations of \ac{HD} computing to capture an abstract representation as a vector in the \ac{HD} space.}}
	\label{HD}
 \vspace{-0.3cm}
\end{figure}

A fundamental block in \ac{HD} computing is the encoder $f:\mX \rightarrow \mH_d$, where $\mH_d$ represents a $d-$dimensional HD space. In this context, the representation $\bmz \in \mX$ may have a dimension of $N$ features, while $\bmh = [h_1,\cdots,h_d]$ represents an \ac{HD} vector with $d \gg N$. Hence, each dimension $h_k, k \in \{1,\ldots,d\}$ of a vector in an \ac{HD} space refers to either a feature or its corresponding value (not necessarily numerical) that are unraveled from the lower-dimensional space containing the representation $\bmz$. For instance, a feature can be the color and the value can be red (non-numerical) or its hexadecimal value (numerical). To initiate a representation in an \ac{HD} space, each feature must be combined to its values as a vector. Then, the combination of these different vectors defines the representation.
Given this mapping, we discuss how the perceived abstract representations can be represented as \ac{HD} vectors. This process is illustrated in~Fig.~\ref{HD} and includes the following sequel of mathematical vector operations:

\begin{itemize}
\item \textbf{Binding (multiplication)}: The binding operation $\boldsymbol{h}_i\otimes \boldsymbol{h}_j$ combines two hyper vectors $\boldsymbol{h}_i, \boldsymbol{h}_j \in \mH_d$ into a new ``\emph{bound}" hyper vector in the same space that represents them as a pair. Hence, a binding operation is equivalent to coordinate-wise multiplication that combines ideas. In general, it is the main operation that binds features to their values. For instance, consider having a feature vector for the human that represents ``direction of movement" and another vector that represents the direction ``right". Thus, the resulting bound vector is nearly orthogonal to both vectors and represents ``direction of movement is right" (See Fig.~\ref{world model}). Broadly, each binding operation will result in a new orthogonal basis and an entry in the \ac{HD} space.
In addition, it is worth noting that if we were to consider binding the vector basis $\boldsymbol{h}_i$ that symbolize the different features, then the resulting \ac{HD} vector $\bigotimes\limits_{i=1}^{N} \boldsymbol{h}_i$ can effectively represent a generalizable structure entity of the representation $\bmz$.

\item \textbf{Bundling (addition/aggregation)}: The bundling operator $\boldsymbol{h}_k \oplus \boldsymbol{h}_l$ involves taking a set of hyper vectors, usually bound vectors, and aggregating them into a hyper vector that represents their \emph{superposition}. A standard technique here is to implement the bundling as an addition operation for real/complex valued vectors (quantitative) and a XOR operation for binary (qualitative) \ac{HD} vectors. For instance, consider a bound vector ``height is tall" that is superpositioned with ``direction of movement is right" to represent a humanoid that is both tall and moving to the right.

\item \textbf{Permutation (ordering)}: This operation involves re-arranging the individual elements of the vectors. This is an efficient way to represent the order of occurrence between the bounds and deal with sequences, particularly, upon superposing bound vectors while aiming to preserve their order. Permutation provides an effective technique to perform \emph{temporal reasoning} about events that occur sequentially.
\end{itemize}

Now that we have transformed the real-time representations into \ac{HD} vectors, the world model will then need to predict the next states of the world. This will require predictions to take place based across multiple factors, as shown in Fig.~\ref{world model}. Initially, the world model must predict the natural evolution of some representations based on previous encounters with such structure (fetched from a memory module that will be explained in Section~\ref{memory}) and a grasp of intuitive physics. Consequently, this change of state is reflected within the entries of the vector. 
For example, a humanoid moving to the right will usually continue in the same direction (to the right) in order to reach their destination, and will not start moving upwards. Hence, the feature of ``direction" will remain constant in the predicted evolution of the humanoid. Also, these predicted states are contingent on other factors. Notably, the future states are impacted by the causal relationships that exist between the bounds of different \ac{HD} vector representations (see Fig.~\ref{HD}). In essence, these relationships are captured within this \ac{HD}-based \ac{SCM}. Hence, once a bound of a certain \ac{HD} vector changes, it will affect the causally related bounds of \ac{HD} vectors pertaining to other real-world elements. In addition, \acp{SCM} account for the random, stochastic variables from the world as stated in Definition~\ref{SCM-Defintion}. Indeed, this random factor is inherently embedded within the \ac{SCM} and considered in the prediction process.


Furthermore, the predicted future states of the real-world elements are affected by the actions of an \ac{AGI}-native network. Hence, the telecom-brain must carefully think of its optimal action sequence before it takes its actions. In particular, the telecom brain must choose a sequence of actions that can bring it closer towards achieving its desired goals and fulfill its intents.
In an \ac{AGI}-native network, these actions can be in the form of beamforming designs, resource allocations, or any network optimization and management functionality. As such, this functionality includes the configurations passed to a \ac{PT} of an autonomous agent to augment it with common sense. Therefore, predicting the future states of these real-world elements will require considering: 1) the natural evolution of the representations, 2) the corresponding cause-effect relationships between the representations, and 3) the effect impinging from the actions of the telecom brain.

\textbf{Open Problems.} When dealing with our vision of \ac{HD} causal world models, there is a need to still address a number of key challenges:
\begin{itemize}
    \item \textbf{Capturing intuitive physics:} Although real-world elements can be represented as \ac{HD} vectors, there is still a need to manipulate these vectors to predict the future states of the world.
    To do so, it is necessary to incorporate intuitive physics and impinge the effect of its operations on the representations. 
    Hence, integrating intuitive physics into the world model requires representing basic physical actions (e.g., motion, force, gravitational weight, collision, etc.) as \ac{HD} vectors. These actions are fundamental physics phenomena such as momentum and friction that humans and agents encounter in the physical world.
    In addition, intuitive physics must allow manipulating the representations of different elements as they interact with each other. For instance, consider the representation of a humanoid and an autonomous vehicle. The telecom brain should be able to infer that the collision of both representations will have a negative impact that increases the cost. Hence, it should avoid this risk of accident as it may reduce the \ac{QoE} of the autonomous vehicle and can have undesirable effects. Here, one possible solution for \ac{AGI}-native networks to capture these forces is through learning physics principles from real-world situations. Indeed, emerging \ac{AI} models such as \ac{JEPA}~\cite{assran2023self} that learn how to map abstract representations between different time instances can be a promising solution. For instance, \ac{JEPA} models can be exploited to extract forces from data and learn how these forces affect the representations and bounds. Moreover, transforming the perceived forces into \ac{HD} vectors requires a more elaborate analysis of category theory to determine the functor objects. 

\item \textbf{Learning functors from symbol to vector category:} 
The transformation from symbols to \ac{HD} vectors through category theory can be conceptualized via a functor mapping. A simple, yet efficient approach to represent the functors ($F$) between category $\mL$ and $\mwL$ can be through linear transformations ($\bmV_F$) as follows:
\begin{equation}
\begin{aligned}
    \hspace{5mm}&\mL \hspace{1.1cm} \bmu \xrightarrow[\hspace*{1.8cm}]{\bmK_f} \hspace{2mm}\bmv \\ &\xdownarrow{0.5cm}F \hspace{0.8cm} \xdownarrow{0.5cm}{V_{F}}\hspace{2cm}\xdownarrow{0.5cm}{V_{F}} \\ &\mwL\hspace{0.7cm} \bmV_F \bmu \xrightarrow[\hspace*{1.5cm}]{\bmK_{F(f)}} \bmV_F\bmv
\label{eq_structural}
\end{aligned} 
\end{equation}  
The resulting transformation matrices $\bmV_F$ should be chosen such that it obeys the structural properties $\bmK_f \bmV_F = \bmV_F\bmK_{F(f)}$ (ensuring the preservation of morphisms between $\mL$ and $\mwL$).
Physically, this implies that the functor maintains the meaning or interpretation of relationships among objects within category $\mL$, even after its transformation to $\mwL$. Nevertheless, the transformation of a single symbol from $\mL$ to $\mwL$ decomposes the symbol into multiple feature vectors. One key challenge is to learn this one-to-many trasformation in category theory in an unsupervised fashion. Here, one promising approach can be through functorial learning \cite{SheshmaniMLST2021}. More generally, the development of fundamental techniques grounded in category theory for building the world model is an important open area for research in this space. 

\item \textbf{Training and updating the world model:} Analogous to humans that learn and enhance their world models as they progress over time, the telecom brain must learn and update its world model with new encounters and scenarios. This update encompasses discovering new real-world elements and updating the causal relations.
Hence, the world model should be differentiable to allow for this update. Effectively, the \ac{SCM} must optimize the complexity of the symbols and update the causal relations, in a gradient-inspired fashion, to maximize the reward of the telecom brain. Nevertheless, pinpointing the threshold at which to update the world model is still an important challenge. Furthermore, updating the world model could also require simulating alternative realities that were not encountered in the real-world and what could have happened if other actions were to take place. In other words, updating a world model should be induced through imagination. Imagination is done through evoking hypothetical scenarios of reality by dynamically altering the features of real-world elements. Here, as our model integrates causal relationships into its system, we conjecture that counterfactuals and interventions can be leveraged to simulate these alternative realities and update the world model~\cite{pearl2009causal}.

\begin{figure}
	\centering
	\includegraphics[width=\linewidth]{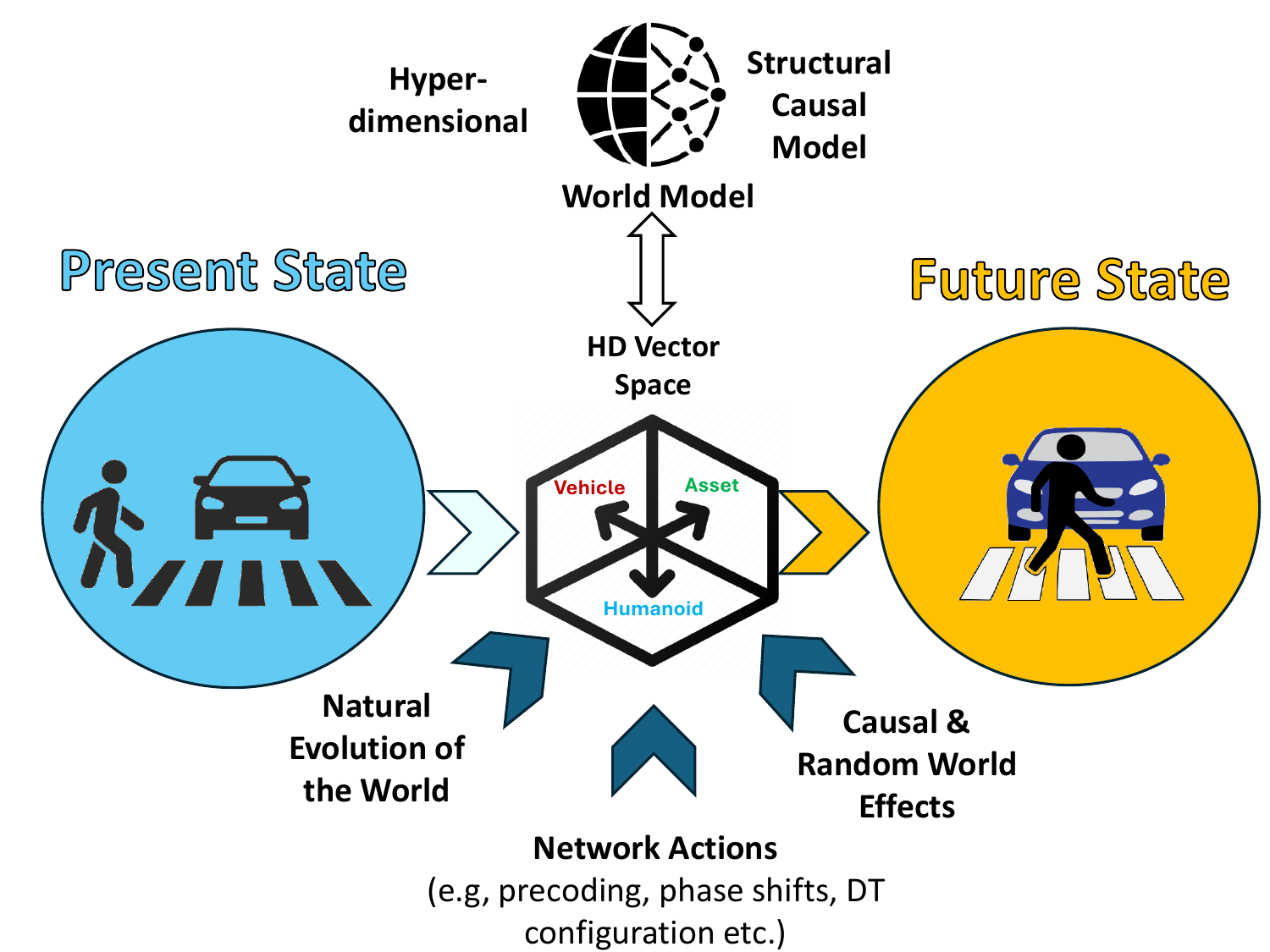}
	\caption{\centering \small{Illustrative figure showcasing how an \ac{HD} vector space of a world model can facilitate the prediction of future world states.}}
	\label{world model}
\end{figure}

\end{itemize}

Once the world model is built, the telecom brain will have to plan its actions in attempt to achieve its objectives or fulfill its intents. To do so, the telecom brain can engage in planning methods to perfectly choose these optimal set of network actions (e.g., resource allocations, beamforming, etc).
Next, we will define the intent-driven and objective-driven planning methods of the action-planning module in our vision in Fig.~\ref{Telecom_Brain}.

\subsection{Action-planning: Intents vs. objectives}
\label{action-planning}


For an \ac{AGI}-native network to plan its actions, it has to imagine the plausible future states of the world as a function of these possible actions. As such, the telecom brain must choose the actions that will minimize its costs (or maximize its rewards), and bring it closer towards its goals i.e., objective or intent. This can be facilitated through two main planning methods: i) \emph{intent-driven} planning and, ii) \emph{objective-driven} planning. Here, intent-driven planning refers to the network strategy of determining actions to fulfill intents that do not necessarily incorporate a particular end-goal or objective. In contrast, objective-driven planning refers to the network strategy to drive its actions towards achieving an objective or goal. Indeed, the distinction between intent and objective is that objective-driven planning encompasses an end-goal that the network should attain, whereas intent-driven planning does not necessarily incorporate such a goal. Next, we describe in more detail how these two planning methods can be developed.

\subsubsection{Intent-driven planning}
In general, the concept of intent refers to the purpose or aim behind a certain action. In other terms, it signifies the motivation that drives one to act in a certain way. Similarly, the motivation behind intent-driven planning in \ac{AGI}-native wireless networks is typically to drive in a reduced cost (enhanced reward) for the telecom brain, or more broadly the wireless network. This cost (reward) can involve multiple factors, and it can be defined in terms of the intrinsic and extrinsic costs of the telecom brain, i.e., \ac{QoNE}, \ac{QoPE}, \ac{QoDE}, and \ac{QoVE}. For instance, an intent in an \ac{AGI}-native network can be to: ``Satisfy the users' \ac{QoE} requirements while minimizing the power consumption of the network".
As such, intent-driven planning refers to finding the set of actions that are incorporated into the \ac{HD} causal world model that can bring the telecom brain closer to fulfilling its intent. 
This is distinct from objective-driven planning as it does not account for the existence of a measurable goal that the network can come close to accomplish with a set of sequential actions. 

Nevertheless, planning with a causal world model considers a form of interrelation between the different world states. Therefore, intent-driven planning is contingent on the causal information present in the world model.
For instance, planning over time depends on the degree to which we can reliably imagine the world states up ahead. Hence, imagining the future states with confidence is a major aspect to be considered in intent-driven planning so as to ensure the fidelity of the planned actions. Accordingly, we propose to quantify this causal dependence between the world states to extract the information about the future states through a brain-inspired approach. In particular, this approach builds on \emph{\ac{IIT}}~\cite{tononi2016integrated} from the field of neuroscience which can possibly quantify the number of planning steps the telecom brain can perform to reliably imagine into the future.

In essence, capturing the causal relationships between abstract representations (or their bounds) is a cognitive aspect of the brain that requires a state of consciousness~\cite{freeman1999consciousness}. This consciousness is then reflected in the sequential states of the world that appear over time. One attempt to capture this consciousness can be through \ac{IIT}. In fact, \ac{IIT} states that consciousness is embedded in the amount of \emph{integrated information generated by a system}~\cite{tononi2008consciousness}. Integrated information refers to the extent to which the information within a system is unified and cannot be subdivided into independent parts. In particular, \ac{IIT} can provide an analytical solution to quantify the information conveyed collectively within the sequential states of the causal world model. This metric can be leveraged to assess the common sense of the telecom brain. On the one hand, it provides a technique to capture the causality between the states of the representations. In consequence, this technique inherently incorporates intuitive physics. On the other hand, it can provide an overall assessment of the causal relations that exist between the different abstract representations.
As such, \ac{IIT} characterizes information as both \emph{causal and intrinsic} based on the influence of the current states on the likelihood of its past and future states. Hence, \ac{IIT} can play a crucial role in capturing the depth of the planning steps that the telecom brain can reliably perform. 
Clearly, \ac{IIT} represents a shift from traditional information theory that statistically captures the mutual information $\mathbb{I}(X;Y)$ between two random variables and overlooks the causal dependency between them, i.e., $\mathbb{I}(X;Y) = \mathbb{I}(Y;X)$. In contrast, \ac{IIT} is inherently tailored towards causal relationships and can capture the causal information conveyed by the states of the world model. Next, we present a primer on quantifying \ac{IIT} to capture the information in the world perceived by our \ac{AGI}-native network.

The planning methodology can be defined by options representing sequences of actions in a structured manner \cite{MachadoJMLR2023}. An option $\omega \in \Omega$ is a tuple $\omega = \langle I_{\omega}, \pi_{\omega}, \beta_{\omega}\rangle$, where $I_{\omega} \subset \mS$ is the option's initiation state, with $\mS$ representing the set of states of the representation, $\pi_{\omega}:(\mS\times \mA)^T \rightarrow \left[0,1\right]$ is the planning strategy (over $T$ time steps) that describes the causal sequence of states and actions, and $\beta_{\omega}$ is the goal state to be reached. To compute the optimal planning steps in $\pi_{\omega}$, it is crucial to quantify the information conveyed by the causal transition from $\bms^0_i$ to $\bms^T_i$. Here, $s_i^t$ represents the state of the representation $i$ at time $t$.
The sequence of causal states $\mathcal{S}_i$ includes the causal state transitions, and can be defined as $\mathcal{S}_i = \left\{\bms^0_i,\bms^1_i,\ldots,\bms^{T-1}_i,\bms^T_i\right\}$. To capture the information conveyed by this set of causal states, we consider the intrinsic and integrated information. As such, the intrinsic and integrated information can be leveraged to quantify the information conveyed by each abstract representation and  integrated in the world model, respectively, as follows~\cite{OizumiPLOS2014}:

\begin{itemize}
    \item \textbf{Intrinsic information for abstract representation}:
The intrinsic information refers to the inherent cause-and-effect structure related to an abstract representation that
produces a particular set of observed states and transitions. 
This is conveyed by any state $\bms_i^t$ as follows: 
\begin{equation}
\mathbb{I}(\bms_i^t) = \min \{\mbI_c(\bms_i^{t-1}\mid \bms_i^t),\,\mbI_e(\bms_i^{t+1}\mid\bms_i^t)\},
\end{equation}
where $\mbI_c(\bms_i^{t-1}\mid \bms_i^t) = \mathbb{D}\left(p\left(\bms_i^{t-1}\mid \bms_i^t\right ) || p\left(\bms_i^{t-1}\right)\right)$ is the \emph{cause information} that the current state $\bms_i^t$ 
specifies about the past, $\mbI_e(\bms_i^{t+1}\mid\bms_i^t) = \mathbb{D}\left(p\left(\bms_i^{t+1}\mid \bms_i^t\right) || p\left(\bms_i^{t+1}\right)\right)$ is the \emph{effect information} that $\bms_i^t$ specifies about the future, $\mathbb{D}$ is a distance measure between probability distributions of each representation (e.g., Wasserstein distance, \ac{KL} divergence, etc.), and $p$ is the probability distribution of the state of each representation $i$.
Although the intrinsic information may convey the information captured by an abstract representation, it is still necessary to integrate this information with that of other abstractions to convey the information represented collectively by the world model, which is defined next.
\item \textbf{Integrated information for world model}:
The integrated information represents the information generated
by a world at a certain state, beyond the information generated by its individual representations. To capture this integrated information, one can partition the world state $\bms^t$ into $m$ parts $\bmM_1^t, \bmM_2^t, \cdots, \bmM_m^t$. Accordingly, this partition $p_k \in \mathcal{P}$ (the set of all partitions of the world done in $k$ ways) is defined such that $\cup_i \bmM_i^t = \bms^t$ and $\bmM_i^t \cap \bmM_j^t = \emptyset$. Here, $\bms^t$ represents the set of \emph{world states} $\{\bms^1,\ldots,\bms^{T-1}\}$, whose cause state is the initial state $\bms^0$ and the effect state is the goal state $\bms^T$. 
Hence, the integrated information of a world model given by its irreducibility over its minimum partition $p_k \in \mathcal{P}$ can be defined as follows:
\begin{equation}
\begin{array}{l}
\mbI_{\Phi} = \mbI_{\Phi}^{p_k^*}, \\
\textrm{s.t.}\,\,\, p_k^* = \argmin\limits_{p_k} \frac{\mbI_{\Phi}^{p_k}}{\max\limits_{p_i\in\mathcal{P}}\mbI_{\Phi}^{p_i}}, 
\end{array}
\label{eq_Iphi}
\end{equation}
where $\mbI_{\phi}^{p_k} = \min (\mbI_{\Phi,c}^{p_k} ,\mbI_{\Phi,e}^{p_k} )$, having $\mbI_{\Phi,j}^{p_k} = \mbI_j(\bms_i^{t-1}|\bms_i^t) - \sum\limits_k \mbI_j(\bmM_k^{t-1}|\bmM_k^{t})  \forall j \in \{c,e\}$.
It is worthwhile noting that the normalization above is over the maximum possible value that $\mbI_{\Phi}^{p_k} $ can take for any partition. 
\end{itemize}

\begin{figure}
	\centering
	\includegraphics[width=\linewidth]{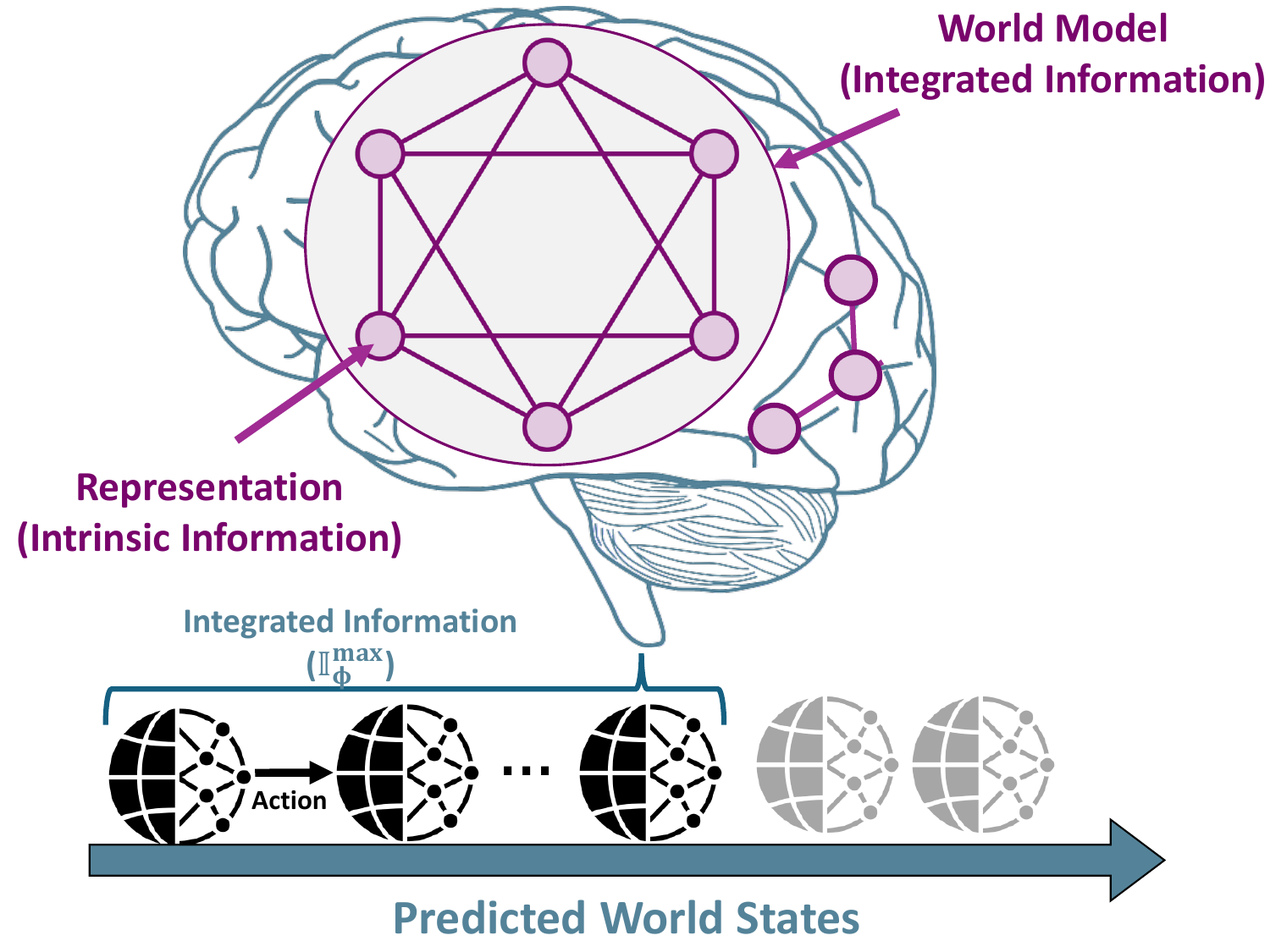}
	\caption{\centering \small{Illustrative figure showing the integration of information between representations in the telecom brain and the role of \ac{IIT} in intent-driven planning.}}
	\label{Intent_Driven_Planning}
 \vspace{-0.3cm}
\end{figure}

Thus, in order to capture the corresponding integrated information $\mathbb{I}_{\Phi}^{\textrm{max}}$, we must find the optimal partitioning $p_k^*$ that can maximize the value of the information in \eqref{eq_Iphi}. Here, $\Phi$ refers to the level of consciousness essential for the integration of information in the telecom brain~\cite{seth2022theories}. Hence, 
$\mathbb{I}_{\Phi}^{\textrm{max}}$ represents the amount of information conveyed about the world and measures the potential of the telecom brain to generate conscious experiences. Henceforth, we can use this metric in intent-driven planning to reflect the depth to which the telecom brain can plan ahead of time. As a result, the integrated information can further be leveraged as a relative measure that captures the number of planning steps that the telecom brain can perform, as shown in Fig.~\ref{Intent_Driven_Planning}. Reflecting on this metric to determine the number of planning steps that the network can perform, the telecom brain must choose the actions that minimize its cost at every step or as a moving average over all steps accordingly.

\subsubsection{Objective-driven planning}

In contrast to intent-driven planning, objective-driven planning primarily considers an end-goal that an \ac{AGI}-native network must achieve.
Here, the telecom brain must determine the action steps for the network to converge towards an objective with minimum cost (or maximum reward). 
However, the telecom brain of an autonomous \ac{AGI}-native network must perfectly plan its actions upon monitoring the convergence of the network towards this objective. One possible way to do this is through \emph{hierarchical planning}~\cite{LeCun2022OR}. 
In particular, hierarchical planning is a problem-solving approach that involves organizing goals into a structured hierarchy of sub-goals and actions. This hierarchical structure enables the decomposition of complex tasks into smaller tasks, making it easier to plan and execute actions efficiently.
Thereby, hierarchical planning considers planning at different levels of abstraction. Notably, the telecom brain can plan its actions over longer terms with higher orders of abstract representations~\cite{knoblock1994automatically}. This long-term planning can then guide short-term planning at lower orders of abstraction. Planning at these lower levels involves determining the intermediate goals as well as the granular steps of actions that must be taken by the telecom brain.
Here, the goal or objective can be specified by human intervention or by the telecom brain. For instance, an objective in an \ac{AGI}-native network can be to: ``Reduce the power consumption in the network by $5\%$". Next, we will discuss how this example can be solved through objective-driven planning.
To do so, we will explain how an \ac{AGI}-native network can acquire abstract representations at different hierarchical levels. Subsequently, we will articulate how these hierarchical representations can be leveraged in hierarchical planning.

The features of each abstract representation in the \ac{HD} space can be categorized in a hierarchical manner according to their concept levels. One promising approach to extract these features in a structured hierarchical manner is through object-centric representation learning~\cite{ChenArxiv2021,TangCVPR2023}. Hence, features of these representations can be categorized into three main hierarchical concepts: a) extrinsic concepts, b) dynamic concepts, and c) intrinsic concepts, described as follows:
\begin{itemize}
    \item \textbf{Extrinsic concepts:} Extrinsic concepts encompass the features situated at the lowest level of abstraction, such as the location of the real-world element. Effectively, these concepts are surface-level attributes, and the perception module can directly encode these contexts at the lowest level of abstraction. Nevertheless, it is essential to note here that the lowest level of abstraction provides granular steps in terms of the future prediction of these features. In fact, it is challenging to predict how these features may change on the long term and are therefore explicitly confined to short-term predictions and planning.
    
    \item \textbf{Dynamic concepts:} Dynamic concepts deal with a higher level of abstraction provided by dynamic concepts that are concealed within temporal and spatial characteristics. Unlike extrinsic concepts, dynamic concepts are suitable to carry out predictions on a longer term. 

    \item \textbf{Intrinsic concepts:} These concepts reside at the highest level of abstraction. Intrinsic concepts include those characteristics that are consistently static for long periods of time. Basically, they include the basic defining characteristics of real-world elements that would possibly indicate a change in the core of the element if they were to be modified. Thus, intrinsic concepts are resilient to change and much more suitable for long-term predictions.
\end{itemize}


As such, an \ac{AGI}-native network can form hierarchical representations of real-world elements at multiple levels of abstraction by categorizing the features of real-world elements into these concepts. These abstract representations include those of real-world elements such as humans and assets, in addition to the \ac{RAN}-\ac{DT} and core-\ac{DT} along with their elements such as beams and \acp{RIS}. We next discuss how these hierarchical orders of abstract representations are leveraged for hierarchical planning.

After discriminating the features of abstract representations according to their concept levels, the network must leverage these abstract representations, at different hierarchical levels, to plan its actions.
This can facilitate hierarchical planning in emerging \ac{AI} frameworks such as the objective-driven \ac{AI} scheme proposed by Y.~LeCun~\cite{singh2023review, LeCun2022OR}.
For instance, consider that the network is instructed to go from its current state ``$A$" to another state ``$B$" with the goal of reducing the power consumption of the network by $5\%$.
At a higher level of abstraction, the action is described as a straightforward ``Optimize the network to move from $A$ to $B$." However, when examining the lower levels, the telecom brain must break down the goal into sub-goals and smaller tasks. Consequently, the telecom brain must choose the optimal actions at each sub-goal. These sub-goals can involve actions like optimizing the resource allocation and beamforming schemes that ultimately enable the network to converge from $A$ to $B$ over a series of steps.

\begin{figure*}
	\centering
	\includegraphics[width=0.9\linewidth]{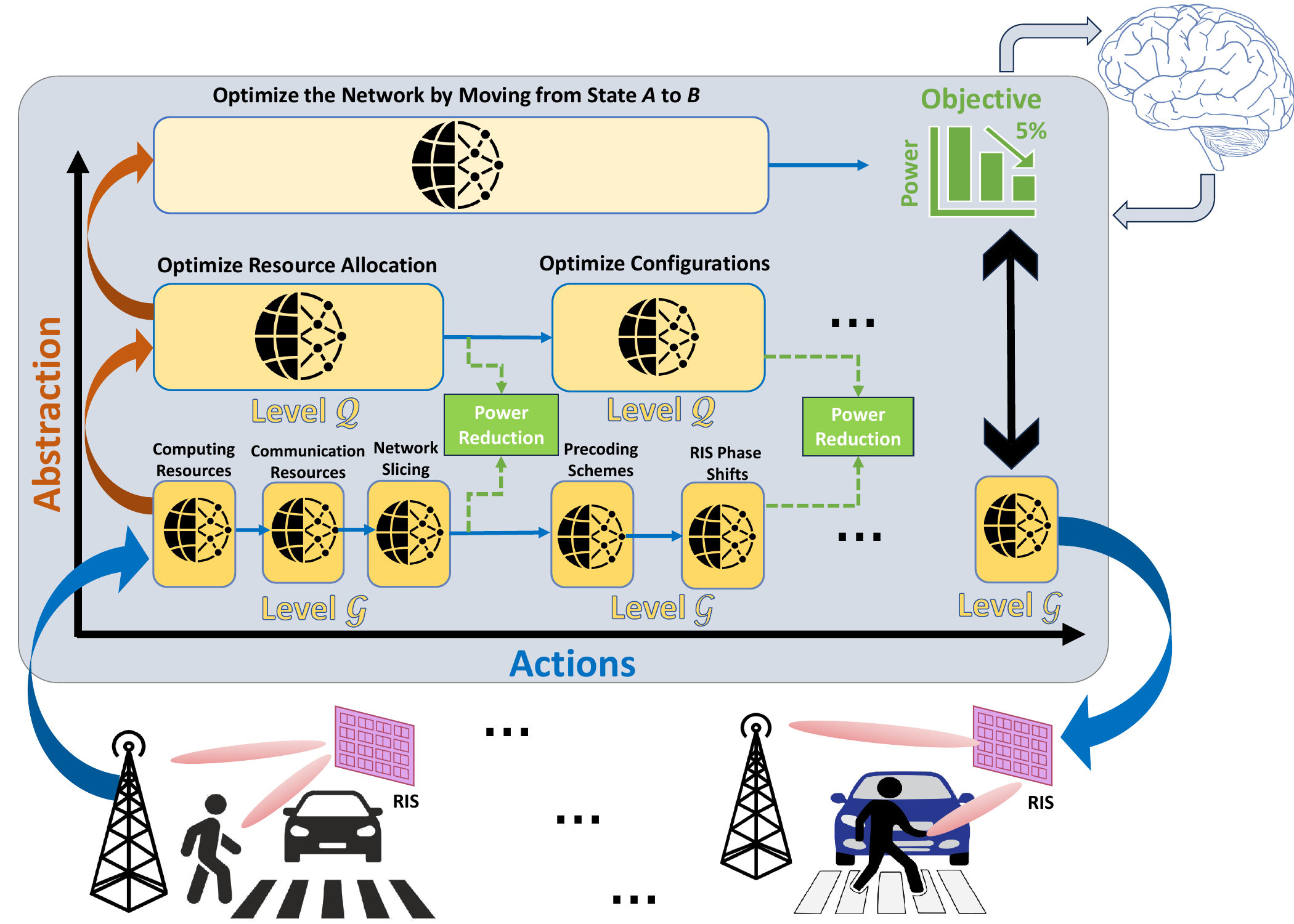}
	\caption{\centering \small{Illustrative figure showcasing an example of objective-driven planning in AGI-native networks.}}
	\label{Objective_DrivenPlanning}
\end{figure*}

As shown in Fig.~\ref{Objective_DrivenPlanning}, the basis for objective-driven planning involves the ability of the network to group adjacent tasks at a lower abstraction level $\mG$ into clusters. Each one of these clusters is represented by a single node in another, higher abstraction level $\mathcal{Q}$. For instance, optimizing the resource network configurations at level $\mathcal{Q}$ can be clustered into sub-goals involving optimizing the precoding scheme at the \ac{BS} and \ac{RIS} phase shifts at level $\mG$.
Accordingly, as $\mathcal{Q}$ is more abstracted, this can facilitate a longer term and more efficient form of planning. Hence, when the network is oriented to move from state $A$ to a state $B$ in $\mG$, the telecom brain can initially plan at a high level of abstraction in $\mathcal{Q}$. Subsequently, this is translated into actions at a lower level of abstraction within $\mG$.
Significantly, upon identifying the high-level path in $\mathcal{Q}$, the agent should exclusively plan within the current cluster in $\mG$. In other words, it only needs to consider its transition from one step to the other, so as to reach the same high level abstract goal with minimal costs. This process repeats until reaching the end goal state in the final cluster.
This hierarchical structure in planning enables the agent to initiate progress toward the goal without calculating the full path in $\mG$. Instead, the agent can follow the high-level plan in $\mathcal{Q}$ and refine it gradually in $\mG$ during execution. Effectively, this hierarchical approach can be recursively applied to higher levels of hierarchies, where higher levels of abstraction continue to be clustered, culminating in a single node at the top of the hierarchy representing the original orientation towards the goal.

\textbf{Open Problems.} While both intent-driven and objective-driven methods provide key strategies for the telecom brain to plan its actions, there still exists different challenges that \ac{AGI}-native networks must overcome so as to proficiently determine their actions, as follows:
\begin{itemize}
    \item \textbf{Design of telecom brain costs and metrics:} Both intent-driven and objective-driven planning methods require minimizing the cost (maximizing the reward) so as to determine the optimal actions of the telecom brain. Namely, these rewards encompass the intrinsic \ac{QoNE} along with the extrinsic \ac{QoPE}, \ac{QoDE}, and \ac{QoVE}. Hence, defining these \ac{QoE} metrics is of substantial importance. Along those lines, our work in~\cite{chaccour2023joint} was the first attempt to define the \ac{QoPE} in terms of the uplink rate, downlink rate, and the \ac{E2E} delay of an \ac{XR} experience. Nevertheless, rigorously determining the rest of the parameters is indeed challenging. However, we can see that the \ac{QoVE} can include the synchronization between avatars and \ac{XR} users. In addition, the \ac{QoNE} can include metrics related to the sustainability, spectral efficiency, and resource utilization in the network. Furthermore, there is a critical need         to design a novel formulation that maps between \ac{IIT} and the number of planning steps to facilitate intent-driven planning.

    \item \textbf{Exploring and executing new actions:}
    While planning has focused on determining the optimal actions of the \ac{AGI}-native network, the pool of actions for the telecom brain is not limited to a closed set of actions. Hence, it is imperative to ask how \ac{AGI}-native networks can innovate to determine new sets of actions that reflect real intelligence. These actions are important when dealing with unforeseen scenarios that may require the network to think outside the box. One solution for such innovation could be through \emph{compositional generalization}~\cite{ito2022compositional}.
    In particular, compositional generalization refers to the capability of generating novel combinations of familiar elementary concepts or actions.
    In \ac{AGI}-native networks, compositional generalization can possibly be defined using the concept of deductive logic. In particular, a deductive logic of actions $c$ represents the conjunction of $N$ actions, i.e.,  $c^{(N)} = (c_1 \land c_2 \cdots \land c_N)$. Here, the telecom brain can choose a novel action $c^{(N)}$ that is a combination of individual actions $c_i$.
    While actions within an \ac{AGI}-native network should be determined to satisfy objectives and intents, dealing with unforeseen scenarios can require a glimpse of novelty in actions. However, this requires \ac{AGI}-native networks to capture broad analogies between novel situations and generalize concepts among the maximum possible number of situations. In this case, an \ac{AGI}-native networks can form relations between situations to deduce such new actions. 

    \item \textbf{Thinking fast and slow:} As the telecom brain architecture presents new opportunities for cognitive abilities in communication networks, its main functionality focuses on the slow, analytical mode of thinking (see Section~\ref{perception}). Nevertheless, humans are not constantly in a deep thinking mode. In particular, humans transition to this mode only when they require focus, logical reasoning, and dealing with critical scenarios. That said, humans effortlessly rely on their fast, intuitive mode of thinking to respond to typical tasks. In other words, humans balance between their fast and slow modes of thinking to take actions~\cite{kahneman2011thinking}. Similarly, an \ac{AGI}-native network must proficiently leverage both modes to take its actions. Typically, these actions range from those requiring continuous, real-time configurations such as resource allocations, to actions that require advanced thinking such as dealing with unforeseen scenarios facing autonomous agents. To incorporate the fast mode of thinking, one can typically rely on some of the \ac{AI}-native infrastructure incorporated into 6G networks. This mode includes solutions encompassing \acp{NN}, auto-encoders, meta-learning, etc. that an \ac{AGI}-native network can build on to further advance fast thinking. Henceforth, it is critical to harmonize the interoperability of both systems for thinking in the telecom brain.

\end{itemize}

In the slow mode of thinking, a major part in planning the actions in unforeseen scenarios comes after carrying analogies with previous instances of real-world elements. Hence, it is necessary for the telecom brain to store and manage the corresponding representations in its memory space for direct analogy. Next, we explain the role of the memory in the telecom brain and what cognitive (reasoning) abilities can it enable for \ac{AGI}-native networks.
\vspace{-0.2cm}

\subsection{Memory}
\label{memory}
To engage in analogical reasoning, the telecom brain requires two memory components: 1) \emph{item memory} that stores the representations learned from the data, and 2) \emph{associative memory} that allows the retrieval of stored information based on similarity or associative relationships to the perceived representations. This memory structure is beneficial for tasks in which recognizing abstract representations and retrieving relevant information are crucial, such as in the case when the network must deal with unforeseen instances. To perform analogical reasoning, it is necessary to carefully relate real-world events, representations, and features to each other in a way that mostly makes sense. In particular, \ac{HD} vectors can capture semantic relationships between entities. Hence, similarity measures in \ac{HD} spaces  such as cosine similarity can be used to quantify the relatedness or similarity between different vectors, aiding in analogical reasoning. 

We further explain how the \ac{HD} representations stored in associative memory can be leveraged in the following simplistic example. Consider $f^{(0)}(\bmz_k)$ and $f^{(0)}(\bmz_j)$ to be the \ac{HD} vector representations corresponding to dissimilar symbols $\bmz_k$ and $\bmz_j$ at the lowest level (level $0$) of abstraction. As discussed in Section~\ref{HD causal world model}, a bundling operator $\bigoplus$ can be used to combine feature vectors to initiate different levels of abstraction. Accordingly, at a higher level of abstraction (level $1$) with less features, we can consider $f^{(0)}(\bmz_k)$ and $f^{(0)}(\bmz_j)$ to be semantically similar. For simplicity, we partition the symbols into two sets, $\mathcal{X}_1$ and $\mathcal{X}_2$, representing dissimilar symbols at abstraction level $1$. The resulting level $1$ representation for all the semantically similar symbols in the set $\mX_r$, where $r\in \{1,2\}$ can be written as: $f^{(1)}(\mX_r) = \bigoplus\limits_{i} f^{(0)}(\bmz_i)$.
Further, we propose to compute the level $0$ representation $f^{(0)}(\bmz):\mX \rightarrow \mH_d$ as follows:
\begin{equation}
\begin{aligned}
   f^{(0)\,\ast}(\bmz) =& \argmax\limits_{f^{(0)}} \frac{1}{\abs{\mX_{1}}}\sum\limits_{\bmz_i \in \mX_{1}}\rho(\bigoplus\limits_{i} f^{(0)}(\bmz), \bigoplus\limits_{i} f^{(0)}(\bmz_i)),\\
   \mbox{s.t.}\,\,\, &\frac{1}{\abs{\mX_{2}}}\sum\limits_{\bmz_i \in \mX_{2}}\rho(\bigoplus\limits_{i} f^{(0)}(\bmz), \bigoplus\limits_{i} f^{(0)}(\bmz_i)) \geq \epsilon,
\label{eq_level1_abstraction}
\end{aligned}
\end{equation}
where $\rho(\bma,\bmb) = \frac{\langle \bma,\bmb\rangle}{\abs{\bma}\abs{\bmb}}$ is the cosine similarity. The constraint in \eqref{eq_level1_abstraction} captures the learned level-$1$ abstraction for $\bmz$ that should be far away from the symbols in $\mX_2$. The objective of \eqref{eq_level1_abstraction} is to compute an HD encoding that brings together objects $\bmz$ with similar semantics at a higher layer while ensuring that they are distant from dissimilar objects. The value of $\epsilon$ is contingent upon the desired reliability for conducting analogical reasoning at the first level of abstraction and the dimensionality of the \ac{HD} vectors as discussed in \cite{ThomasJAIR2024}.

Having defined our \ac{AGI}-native network's components, we next discuss some of the use cases that it will engender.

\section{Use Cases and Experiences in \ac{AGI}-Native Wireless Networks}

Realizing the telecom brain discussed in Section~\ref{Telecomverse} will bring forth unprecedented levels of general intelligence into the wireless network. In addition, the impact of \ac{AGI} will extend to bring forth new use cases and experiences for humans and autonomous agents. Essentially, these use cases include \acp{DT} with analogical reasoning and cognitive avatars with resilient, synchronized experiences. The use cases also include brain-level metaverse experiences such as holographic teleportation with \ac{ToM} (see Fig.~\ref{evolution}). In this section, we provide preliminary expositions of these use cases and highlight their challenges and opportunities. We also note that \ac{AGI}-native networks may pave the way towards a broader set of applications that we cannot yet identify at this early stage.

\vspace{-0.15cm}
\subsection{Analogical reasoning for next-generation \acp{DT} and networks}

One of the crucial pillars of \ac{AGI}-native networks is their ability to deal with unrecognized real-world elements and events through common sense. This will involve the ability of the telecom brain to relate these new elements and events from the real-world to similar elements and situations stored in the memory through analogy. Hence, analogical reasoning enables the telecom brain to identify and proficiently approach these new cases. As discussed in the example of Fig.~\ref{Two_Examples}, this crucial ability also extends to guide \ac{DT}-enabled autonomous agents in their corner cases. For instance, once an \ac{AGI}-native network identifies an unforeseen element (to the autonomous agent) in the world as an obstacle, it can guide the autonomous agent to move away from it. As such, analogical reasoning becomes an indispensable component for both the network and its autonomous agents.

In order to recognize these new elements, the telecom brain must draw parallels with its elements from the memory. This is of notable importance for the telecom brain while planning its actions, since it can face a multitude of unforeseen elements in the real-world. Hence, the telecom brain will need to interact with these new elements to guarantee reaching its optimal cost. As the world model allows perceiving the elements through \ac{HD} vectors of semantic content, one possible approach for analogical reasoning can be to capture the semantic similarity between the representations. Therefore, the key for reliable inferences in analogical reasoning relies on perfectly perceiving and identifying real-world elements.

In essence, analogical reasoning is a fundamental facet of human cognition that involves a sequential process to identify similarities~\cite{GentnerEHB2012}. In particular, the telecom brain performs analogical reasoning through a mechanism that involves the world model, memory, and perception modules. This mechanism involves the semantic similarity between \ac{HD} vectors and is subdivided into the following processes: 
\begin{itemize}
\item \textbf{Retrieval:}  
Upon perception of an element whose identical (i.e., semantically similar) representation is available in the memory, the network can recall the previous, short-term situations that include such element. In this case, this representation must fall into the semantic space of an element. Furthermore, these retrieved situations include the state of the worlds at those particular instants and their associated costs, while recalling causal relations between this particular element and other elements in the world. 
According to the costs recorded through previous interactions with such element, the telecom brain can plan its actions by either exploration or exploitation. 
This is determined by the level of confidence of the telecom brain upon dealing with this particular element (representation). For instance, if, for every encounter with a given object, the telecom brain chooses the same action repeatedly and is satisfied with the cost, then the confidence levels would lead to exploitation rather than exploration. If the telecom brain does not recognize this representation, it will proceed to a mapping phase (explained next) by considering it as a newly identified object that it needs to learn how to deal with. This is beneficial for the telecom brain as it must provide real-time planning of its actions for an \ac{AGI}-native network. That said, the planning process can be interrupted by every new element that the telecom brain must identify.

\item \textbf{Mapping:}
If the perceived representation does not fall into a certain semantic space, the telecom brain recognizes this element as a new instance. Consequently, the telecom brain attempts to approach this element by mapping its representation to one of the nearest semantic spaces. Herein, we highlight the role of the generalizable abstract representations in identifying these elements. Incorporating generalizability in learning abstract representations increases the size of the semantic space corresponding to each element while bringing representations sharing a common structured entity closer together (see Section~\ref{perception}). Consequently, the possibility that a perceived element falls within the semantic space of a specific or similar element increases. Meanwhile, neglecting this generalizability will reduce the semantic space of each representation. Hence, recognizing known elements becomes non-trivial. Henceforth, generalizability plays a role in providing swift predictions of future states, in contrast to continuously identifying new elements that can hinder real-time predictions.
Essentially, identifying new elements involves aligning the perceived representation to the nearest representations from different semantic spaces. One way to achieve this alignment can be through an attention mechanism. For instance, consider two vector representations $\bmz_t$ and $\bmz_d$ from different semantic spaces. Initially, these representations are mapped to different subspaces such that: $\bmQ = \bmW_1^T \bmz_t$, $\bmK =\bmW_2^T \bmz_d$, $\bmV = \bmz_d$, where $\bmW_1$ and $\bmW_2$ are mapping matrices. Subsequently, attention scores are computed as $\bmA = \text{softmax}(Q^TK/\sqrt{\sigma})$, and the encoded representation $\bmV^T\bmA$ reflects the similarity between $\bmz_t$ and $\bmz_d$. Accordingly, the representation is mapped to the semantic space with the highest attention score. Hence, the telecom brain extracts the preceding encounters with this specific representation from the short-term memory. Nonetheless, based on the attention score that reflects the confidence in mapping this representation, the telecom brain should initiate the interaction with this new element through an exploration-exploitation strategy rather than just equating it to this specific representation.

\end{itemize}


\begin{figure}
\centerline{\includegraphics[width=\linewidth]{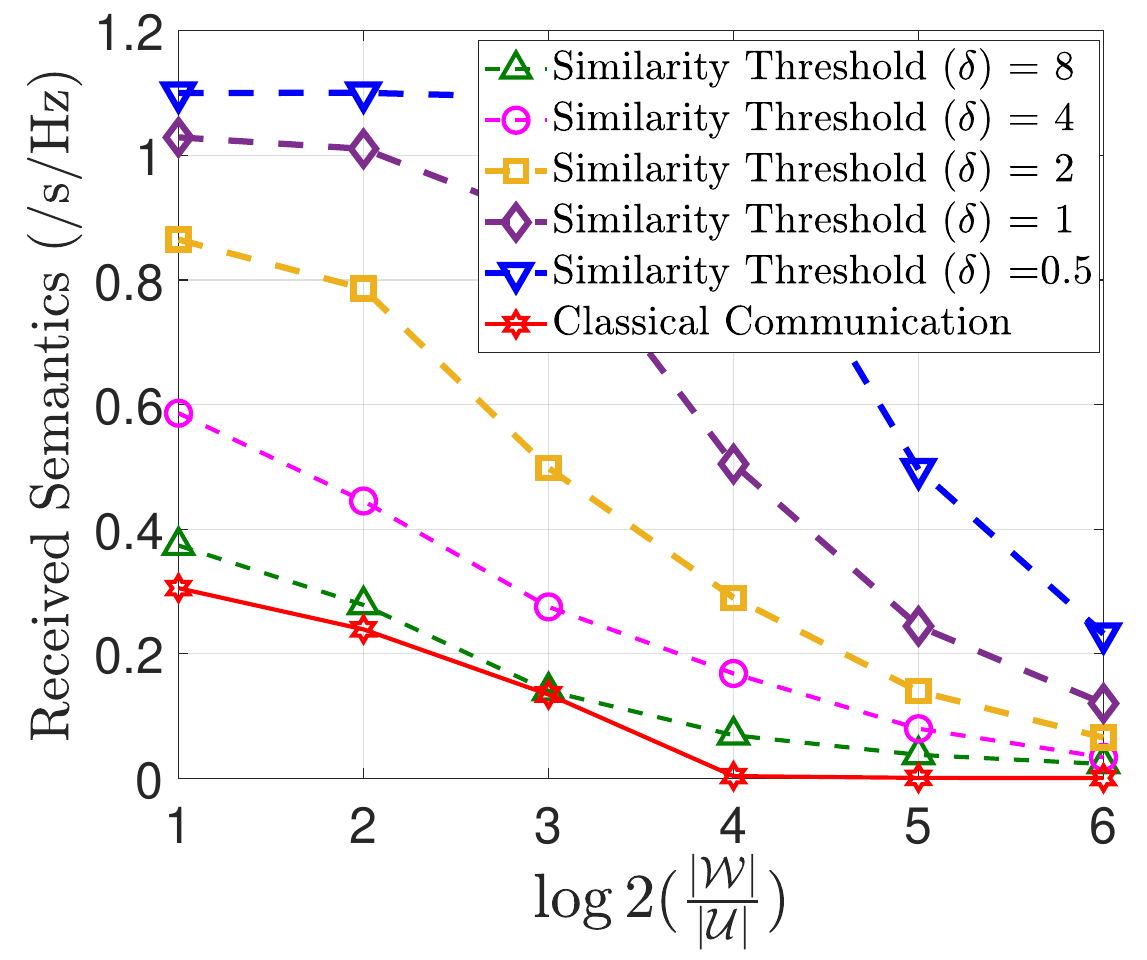}}
\label{EmLang_SemRate}\vspace{-1mm}
\caption{\small Received semantics as a function of reduced semantic representation space $\abs{\mathcal{U}}$ and thresholds, while having $\abs{\mathcal{W}}=256$~\cite{ChristoTWCArxiv2022}.}
\label{TaskAg_Relia_Semantics_2}
 \vspace{-2mm}
 \end{figure}

\begin{figure*}
	\centering
	\includegraphics[width=0.78\linewidth]{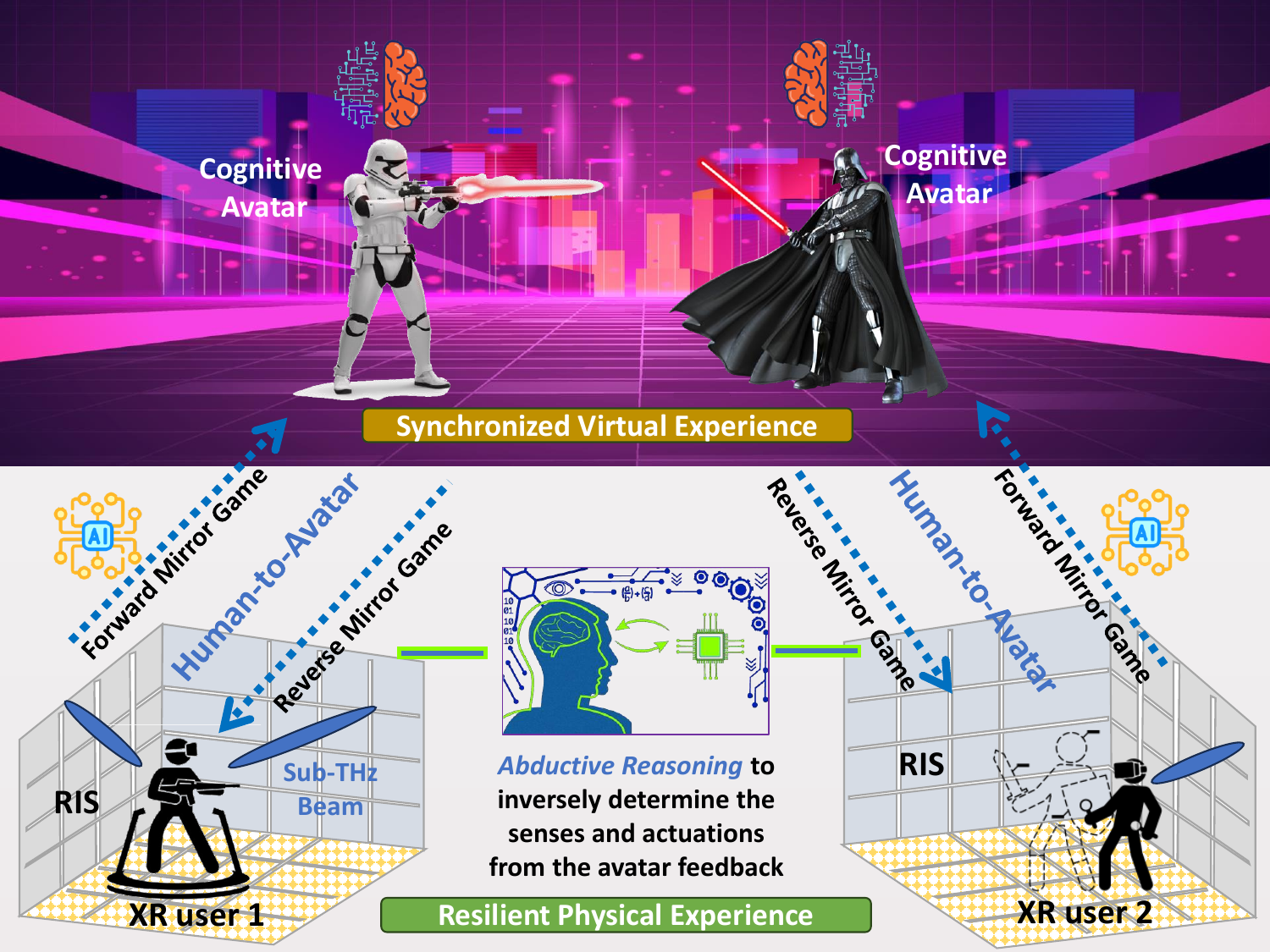}
	\caption{\centering \small{Illustrative figure showing the mirror game between \ac{XR} users and cognitive avatars.}}
	\label{Avatar}
 \vspace{-0.2cm}
\end{figure*}

\textbf{Open Problems.} Specifying the border between retrieval and mapping is a major challenge. In essence, this borderline is contingent on the design of the semantic space in the telecom brain. The design of this space considers two factors: i) the number of representations $\abs{\mathcal{U}}$ in the semantic space $\mW$ and, ii)~the threshold $\delta$ that reflects the semantic space surrounding each representation.
Here, the joint design of $\delta$ and $\abs{\mathcal{U}}$ influences the retrieval and mapping processes to faithfully perceive the real-world.
In fact, our previous work in~\cite{ChristoTWCArxiv2022} studied the impact of the representation space $\mU$ and the threshold $\delta$ on the semantic rate of a communication system. As shown in Fig.~\ref{TaskAg_Relia_Semantics_2}, the reduction in the size of the representation space $\mU$ minimizes the semantic rate, even for low values of $\delta$. This exemplifies an important challenge in analogical reasoning that relates to mapping different elements that exist near each other in the semantic space with a reduced set of representations $\mU$.
In addition, our results in~\cite{ChristoTWCArxiv2022} show that a tradeoff exists between the cardinality measure $\abs{\mathcal{U}}$ and $\delta$ so that the same semantic rate is achieved. This tradeoff provides flexibility in terms of the design of the semantic space so as to ensure the highest confidence in the mapping scores of the recognized elements, while conserving the overall semantics existing in the system.
Therefore, the design of analogical reasoning frameworks in \ac{AGI}-native networks necessitates an in-depth analysis according to the specific setting or application in the real world.

\vspace{-0.13cm}
\subsection{Resilient and synchronized experiences for cognitive avatars}

Realizing the affinity between \ac{XR} users and their avatars depends on the harmonization of the physical and virtual world experiences. Nevertheless, ensuring a seamless mirroring between the physical and virtual realms has multiple requirements.
On the one hand, avatars must authentically embody their corresponding \ac{XR} users, in terms of senses, actuations, and movements, to attain a seamless virtual experience (in terms of \ac{QoVE}). On the other hand, \ac{XR} users require a reliable \ac{QoPE}, which can be expressed in terms of rate, reliability, and latency, to perfectly immerse them into the virtual world.

In essence, fostering this embodiment between \ac{XR} users and avatars requires envisioning it as a harmonious duality, as shown in Fig.~\ref{Avatar}. For instance, the avatar should replicate the sensory and tracking information (e.g., position, movements, etc.) of the \ac{XR} user while interacting in the virtual world. Simultaneously, the avatar must accurately reflect the incoming feedback (from other avatars or virtual objects) from the virtual world to the \ac{XR} user. That said, this duality requires achieving the highest degrees of synchronization while minimizing the mismatch in accuracy and precision between the \ac{XR} user and the avatar.
Nevertheless, achieving this duality is not feasible by considering a mere blind carbon copy approach for avatars. This is due to the fact that the resulting avatars would lack the essential capabilities to initiate a responsive interaction back to their \ac{XR} user. Consequently, the absence of the ability to have back-and-forth interactions between the human user and its avatar prevents executing their corresponding interactions as a complete duality~\cite{lam2022human}.

To effectively address this challenge, avatars should become cognizant of their corresponding \ac{XR} users' actions, by comprehensively understanding the underlying logic stemming from the sensory inputs that initiated them. Accordingly, avatars should transcend being a reactive entity and become a dynamic, \ac{AI}-driven system. To achieve this transformation, these avatars must capture the unique \emph{kinematic fingerprint} of the \ac{XR} user, represented by the mapping between the sensory inputs and corresponding actions~\cite{lombardi2021using}.
Hence, by leveraging this knowledge (i.e., fingerprint), the avatar can reason and execute the action impinging from peer avatars (and virtual elements). 
In this case, the avatar inversely determines the senses and actuations that the user would most likely have experienced due to this action. Subsequently, the avatar feeds back the corresponding senses and actuations to the user. 
Thus, to inversely reach the senses and actuations, an \ac{AI}-driven avatar should be equipped with cognitive \emph{abductive reasoning} capabilities, thereby becoming a cognitive avatar. In fact, we have proposed to model this duality as a bi-directional mirror game in our previous work~\cite{hashash2023seven}.

Furthermore, as cognitive avatars are \ac{AI} models, they face significant challenges when deployed over a wireless network. On the one hand, such avatars must reside at the network edge to reduce the synchronization mismatch with the \ac{XR} user. On the other hand, the avatars must still migrate and interact in the virtual world (at the cloud or another edge), which can make this mismatch more pronounced. Hence, the optimal placement of cognitive avatars becomes a bottleneck for synchronization over networks.
Furthermore, another impediment lies in the ability of the network to ensure an uninterrupted immersive physical experience for \ac{XR} and metaverse users. This is mainly due to the susceptibility of narrow beams to \ac{LoS} blockage  particularly when they operate at high frequency bands (e.g., \ac{mmWave} or sub-\ac{THz}).
Effectively, addressing these challenges requires achieving the following:
\begin{itemize}
    \item Reliably guaranteeing a high \ac{QoPE} for \ac{XR} users upon mitigating \ac{LoS} blockages, and 
    \item Reducing the synchronization mismatch between \ac{XR} users and their avatars to sustain an adequate \ac{QoVE}.
\end{itemize}

Here, we note that wireless networking for \ac{XR} has been studied extensively over the past few years \cite{chaccour2022can, chehimi2023roadmap, hu2020cellular}. However, those prior works do not particularly address the requirements of avatars over wireless networks. In particular, these works do not look into the synchronization aspect and virtual experience of avatars with their \ac{XR} users. In fact, the scope of the prior art is largely limited to the problem of enabling the network to meet the low latency and high rate demands of \ac{XR} applications. While this can be indeed crucial for an immersive physical experience, the state-of-art solutions in~\cite{chaccour2022can, chehimi2023roadmap, hu2020cellular} do not inherently guarantee the reliability of this physical experience. 
In contrast, it is necessary to maintain a reliable, immersive physical experience and synchronized virtual experience for cognitive avatars with their \ac{XR} users. This plays a critical role in embodying \ac{XR} users in their avatars and achieving the \ac{E2E} duality between them.

A possible solution to address these limitations can be presented with an \ac{AGI}-native network. In essence, an \ac{AGI}-native network can leverage its common sense abilities to sustain an adequate \ac{QoPE} for its \ac{XR} users. In particular, the telecom brain can predict the possible future states of the world, as shown in Fig.~\ref{world model}. As such, the telecom brain can foresee whether the \ac{XR} user would suffer from any blockage of the \ac{LoS} beam through intuitive physics. For instance, consider the basic example of Fig.~\ref{Avatar} in which the telecom brain can predict the possible blockage of a sub-\ac{THz} beam from the \ac{RIS} by an obstacle that eventually prevents the establishment of a \ac{LoS} connection between the network and the \ac{XR} user. Clearly, this blockage can reduce the \ac{QoPE}, even for this simple example.

To mitigate this issue, one possible approach is to perform a beam handover so as to  sustain a \ac{LoS} link that preserves the \ac{QoPE}. Unlike other methods that initiate a beam handover once the \ac{QoPE} degrades due to sudden blockages, an \ac{AGI}-native network can anticipate the blockage and proactively initiate a multi-beam handover~\cite{xue2024survey}. Given that the handover process can introduce latency (e.g., to establish the new connection), causing temporary interruptions or delays in communication, this can negatively impact the \ac{QoPE}~\cite{chaccour2023joint}. Hence, an \ac{AGI}-native network, with its proactive abilities, can seek to minimize the duration of such handovers (optimally reaching zero) and the corresponding \ac{QoPE} degradation.
In this case, the \ac{XR} user can be assigned another beam to facilitate a continuous \ac{LoS} which guarantees that their immersive physical experience is uninterrupted. This is in contrast to other handover methods with relatively prolonged handover times that prevent the \ac{QoPE} from returning back to the values necessary for an immersive experience. 
Therefore, an \ac{AGI}-native network ensures a continuous physical experience that is \emph{resilient} to \ac{LoS} blockages and \ac{QoPE} degradation.

Here, the concept of a resilient experience is defined as the ability to mitigate any degradation in the \ac{QoE}, by a swift return to guaranteed levels~\cite{reifert2023comeback}.  
In particular, resilience in our setting means that the physical experience does not suffer from \ac{LoS} beam blockages, whereby anticipating blockage scenarios can initiate proactive handovers to mitigate any interruption in the immersive experience. This, in turn, sustains an adequate \ac{QoPE} reliably within its required levels for a continuous immersive avatar experience.
In contrast to reliability and robustness, resilience is important here because \ac{XR} users are susceptible to frequent \ac{LoS} blockages of their narrow beams at \ac{mmWave} or \ac{THz} frequencies. Hence, these blockage limitations can initiate frequent handovers that can interrupt the immersive experience. Consequently, such interruptions can prevent the \ac{AGI}-native network from sustaining the desired \ac{QoPE} levels. Obviously, the proposed \ac{AGI}-based approach is therefore more reliable than the aforementioned conventional approaches.

In other words, an \ac{AGI}-native network can continuously guarantee a \ac{LoS} for \ac{XR} users as it anticipates their future states and possible blockage through intuitive physics.
Accordingly, this framework requires real-time sensing of the real-world along with high rate low latency communications, simultaneously.
One possible way to achieve this functionality is through a joint sensing and communications framework at sub-\ac{THz} frequencies. On the one hand, communication at sub-\ac{THz} bands promises to provide the necessary data rates and latency requirements for \ac{XR} and metaverse use cases. On the other hand, sensing at the sub-\ac{THz} bands provides a major opportunity to capture the situational awareness that maps the physical environmental into the digital world.
In fact, our previous work~\cite{chaccour2023joint} showed that by leveraging a non-autoregressive multi-resolution generative \ac{AI} framework integrated with an adversarial transformer in such a joint system can outperform other benchmarks such as those of beamtracking in providing a resilient physical experience. As sensing in the sub-\ac{THz} regime can be largely sparse, a multi-resolution generative \ac{AI} framework is adopted to compensate for any missing sensing values. Basically, an \ac{AGI}-native network with a world model can perform similarly to a generative \ac{AI} framework as it can fill in the missing blanks, as described in Sections~\ref{common sense in AI native} and~\ref{HD causal world model}.
In addition, the adversarial transformer enables predicting future situational awareness information that can leveraged for detect blockage and future beam allocation. Evidently, this mimics a key functionality of our envisioned world model.

\textbf{Open Problems.} While an \ac{AGI}-native network can ensure an uninterrupted, immersive physical experience, one important open problem is to design new approaches for reducing the mismatch between the \ac{XR} user and the cognitive avatar so as to provide a synchronized virtual experience. In this regard, we propose to design \ac{AI}-driven cognitive avatars as \emph{foundation models}. These models can be pre-trained over a huge corpus of data that encompasses the tracking and sensory information with the corresponding actions of \ac{XR} users. Accordingly, we propose leveraging the captured kinematic fingerprint of each \ac{XR} user to fine-tune the foundation model to each \ac{XR} user. 
In this case, the foundation model can be placed in the virtual world over the network (at the cloud or at another edge). Thus, each \ac{XR} user must fine-tune this model according to their unique kinematic fingerprint prior to participating in the virtual world. Considering the universal scope of the \ac{XR} users, such foundation models must be open source, whereby all humans can participate in their training process.
To address the synchronization challenge that results from placing the avatar model in the virtual world, an \ac{AGI}-native network must go beyond predicting the future states of the \ac{XR} user through intuitive physics. Here, the \ac{AGI}-native network must predict the future sensory information of the \ac{XR} user with more granular details. Hence, such sensory information expands the scope beyond the framework of joint sensing and communications that is limited to predicting the six degrees-of-freedom of the \ac{XR} user, as shown in our work in~\cite{chaccour2023joint}. Such sensory information can include the specific location of the arms, legs, etc. 
This predicted sensory information can then be leveraged to generate the corresponding actions proactively in the virtual world, by using the foundation model. 
After the interaction in the virtual world, the avatar can determine the feedback to the \ac{XR} user via its reasoning capabilities. Accordingly, this feedback can then be reflected from the avatar to the \ac{XR} user. In essence, this proactive mechanism promises to close the synchronization gaps between the \ac{XR} user and their cognitive avatar. As such, it is imperative for an \ac{AGI}-native to further leverage principles of physics (e.g., laws of motion) to reliably predict the sensory information of the \ac{XR} user. In essence, the prediction of the sensory information is built on the premise that an \ac{AGI}-native can also understand the behavior of the \ac{XR} user. Henceforth, it is crucial to develop novel physics-aware frameworks in \ac{AGI}-native networks that allow them to faithfully predict the sensory information of \ac{XR} and metaverse users while also minimizing the synchronization mismatch.

\vspace{-0.2cm}

\subsection{Brain-level metaverse experiences: Holographic teleportation with \ac{ToM}}
\label{Holo_ToM}



Live metaverse experiences such as \emph{holographic teleportation} provide means to bridge the physical gap between entities residing at different geo-spatial settings. 
Holographic teleportation is based on transmitting descriptive representations of objects and events~\cite{chaccour2022seven}. Essentially, the teleportation of real-world elements and objects requires a merger of digital and virtual worlds to spatially transfer holographic entities over the network~\cite{hashash2023seven}.
In this scenario, relying on classical communications to perfectly describe and transmit large amounts of data in an attempt to construct such elements can fail to meet the stringent \ac{E2E} synchronization delays of this process. Naturally, this can degrade the overall \ac{QoVE} of the metaverse end-user. Alternatively, going beyond classical communications, holographic teleportation should capitalize on capturing detailed real-time abstract representations of objects and events, such as those captured by the telecom brain (see Section~\ref{perception}). These abstract representations are then transmitted from one location to the other for reassembly and generation as holograms.

\begin{figure}
	\centering
	\includegraphics[width=0.95\linewidth]{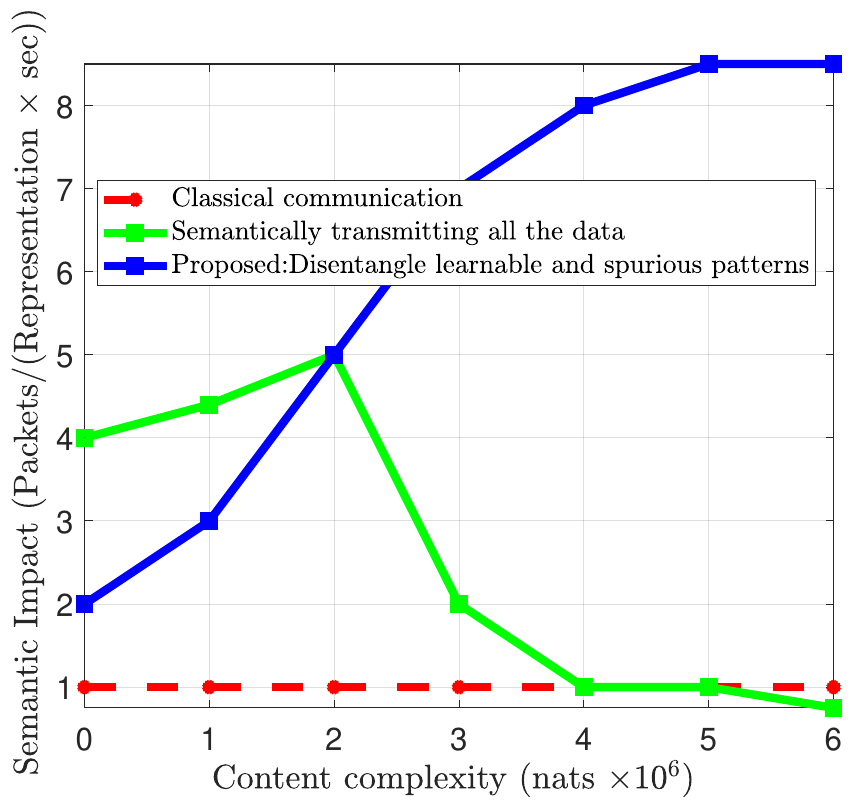}
	\caption{\centering \small{Semantic impact vs. complexity of the transmitted content~\cite{chaccour2022disentangling}.}}
	\label{disentangling}
 \vspace{-0.3cm}
\end{figure}

In general, the telecom brain can capture representations of real-world elements and facilitate their teleportation in an efficient manner over the \ac{AGI}-native network. For example, leveraging our data disentangling approach from Section~\ref{perception} provides a promising method to capture representations and transmit them over the network. As shown in Fig.~\ref{disentangling}, our previous work in~\cite{chaccour2022disentangling} proves that we can efficiently represent rich, complex data (such as that of holograms) for transmission over the network. This includes transmitting the learnable data (as representations) semantically, while continuing to send the spurious data  through conventional classical communications\footnote{Here, we transmit the abstract representation along with the remaining spurious data over the network to faithfully generate the real-world objects.}.
In fact, our approach showcases superiority in terms of semantic impact~\cite{chaccour2022less} over two benchmarks: a) transmitting the data using classical communication and, b)~transmitting all the data semantically. Here, the semantic impact is a metric that captures the number of packets
that would have been needed to be transmitted during a certain time interval to regenerate the semantic content element. Hence, we are able to rigorously and efficiently represent such holograms even when the underlying complexity for such objects or events increases, without jeopardizing the quality of the hologram and the overall \ac{QoVE}.

Nonetheless, one crucial requirement to attain seamless holographic teleportation in such scenarios is based on the correct regeneration of the objects and events at the \ac{Rx} side (i.e., end-user). For instance, any error in reconstruction can lead to a degradation in the \ac{QoVE} of the metaverse end-user. In other words, the constructed elements at the \ac{Rx} should be semantically similar to those at the telecom brain (which here acts as a \ac{Tx} essentially) to achieve a reliable teleportation. 
In particular, consider an element $\boldsymbol{n}$ (i.e., object or event) conveying the abstract representation $\bmz$ that has a semantic message space $\mathcal{C}$.
To be semantically similar, the constructed element $\widehat{\boldsymbol{n}}$ at the \ac{Rx} with an abstract representation $\widehat{\bmz}$ must belong to the same semantic message space $\mathcal{C}$. Moreover, the semantic message space corresponding to $\widehat{\bmz}$ can be defined as the Euclidean space over which the semantic information conveyed by $\bmzh$ is the same within a ball of radius $\delta$ \cite{ChristoTWCArxiv2022}. Thus, to achieve a reliable teleportation, the following condition must be satisfied:
\begin{equation}
E(\boldsymbol{n},{\widehat{\boldsymbol{n}}})  \leq \delta,\,\,  \mbox{s.t.}\,\, E(\bmz,\widehat{\bmz}) = 0,
\end{equation}
where $E(\boldsymbol{n},\widehat{\boldsymbol{n}}) = \norm{\boldsymbol{n}-\widehat{\boldsymbol{n}}}^2$ and  $E(\bmz,\widehat{\bmz}) = \norm{\bmz-\widehat{\bmz}}^2$.

One of the major errors during reconstruction of real-world elements from their representations can be related to the causal models acquired at the \ac{Rx} for re-generation. Here, the acquired causal models incorporate the causal relationships  (e.g., in the form of an \ac{SCM}, \ac{GNN}, etc.) and parameters extracted from the data to describe the underlying elements and events. In fact, reasoning-based transmitters and receivers extract and interpret messages based on their different beliefs and knowledge, whereby this knowledge is captured within the \ac{Tx}/\ac{Rx} parameters. Hence, a mismatch in causal models at the \ac{Tx} and \ac{Rx} can degrade the reconstruction process. Therefore, the \ac{Tx} and \ac{Rx} should be aligned in terms of their parameters.

One possible approach to address this alignment concern can be through the \emph{intuitive psychology} abilities pertaining to common sense in an \ac{AGI}-native network.
In particular, the telecom brain can leverage the psychological concept of \emph{\ac{ToM}} to estimate the concealed mental states of other elements~\cite{ZhangCS2012}. In general, \ac{ToM} can be defined as follows.
\begin{definition}
    \emph{\ac{ToM}} is defined as the cognitive ability of the brain to attribute mental states to others and to oneself, which may not necessarily be in agreement with each other. In essence, these mental states can refer to the different beliefs, emotions, or intentions~\cite{carlson2013theory}.
\end{definition}

\begin{figure}
    \centerline{\includegraphics[width=\linewidth]{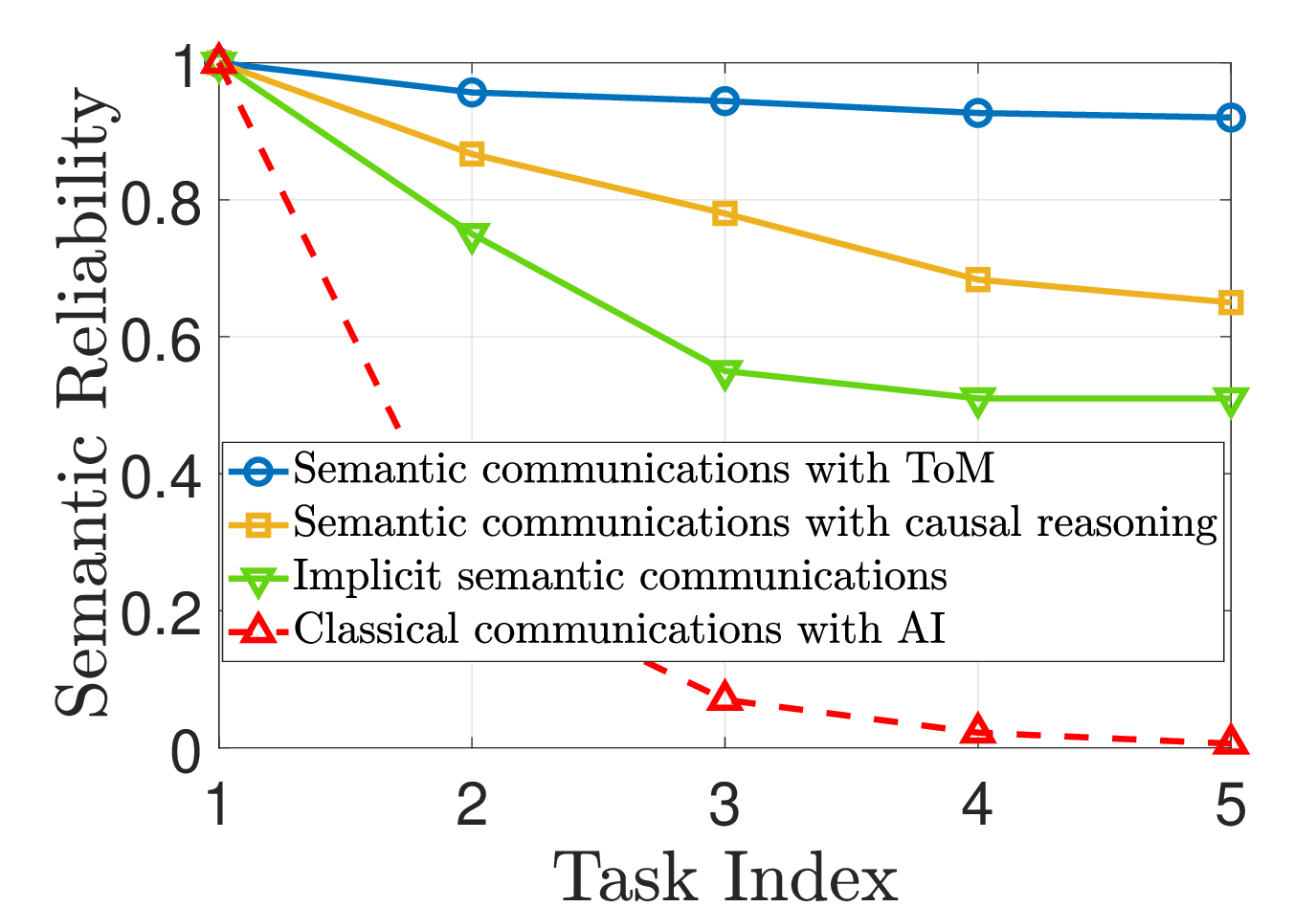}}
    \caption{\small Semantic reliability vs task index (task complexity)~\cite{thomas2023reasoning}.}
    \label{TaskAg_Relia_Semantics}
 \vspace{-3mm}
 \end{figure}

Since an \ac{AGI}-native network operates with common sense, it can possibly reason the mental states of different users in the network (see Fig.~\ref{evolution}). A ``mental state'' in an \ac{AGI}-native network is essentially the causal knowledge and corresponding attained models of the end-user. 
In our case, the \ac{Tx} (i.e., telecom brain) can estimate the mental states (i.e., a priori causal knowledge and models) of the \ac{Rx} side prior to the rounds of communication that take place~\cite{HoTCS2022}. Subsequently, the \ac{Tx} can dynamically adapt its parameters and the corresponding representations based on the feedback measures of semantic effectiveness from the \ac{Rx}~\cite{thomas2023reasoning}. In this case, the telecom brain tries to understand the causal model of the \ac{Rx} and, then, align its parameters with it for reliable reconstruction. 
Effectively, supporting the \ac{Tx} with \ac{ToM} abilities can reduce the number of iterations needed with the \ac{Rx} to achieve the same semantic reliability.
Here, semantic reliability measures the ability to achieve semantic similarity between the \ac{Tx} and \ac{Rx}.
As shown in Fig.~\ref{TaskAg_Relia_Semantics}, our work in~\cite{thomas2023reasoning} proves that the semantic reliability achieved with \ac{ToM} reasoning can outperform several benchmarks that include causal reasoning and implicit semantic communications (i.e., semantic communication with imitation learning-based implicit reasoning). In fact, our results in~\cite{thomas2023reasoning} show that \ac{ToM} can become more effective in achieving semantic reliability for constructing elements at the \ac{Rx} as the complexity of causal relationships (complexity increases with task index) increases. Notably, \ac{ToM} can become crucial for the telecom brain that seeks to acquire complex abstract representations (see Fig.~\ref{Perception_Disentangling_Causal_Generalizable}) and can possibly leverage them to enable applications such as holographic teleportation. Therefore, \ac{ToM} can present a promising ability for \ac{AGI}-native networks to further enhance reliable communication over the network through common sense.

\textbf{Open Problems.} 
Although \ac{ToM} can play an crucial role in an \ac{AGI}-native network, scaling \ac{ToM} with multiple receivers can be challenging. For instance, consider an example of holographic teleportation that supports the omnipresence of the teleported element at multiple receivers.
Clearly, the telecom brain will have to transmit its representations to receivers that have different causal models. Here, the telecom brain has to adapt its parameters to compromise between the different causal models at the \acp{Rx}. Effectively, this compromise could jeopardize the semantic reliability of the \ac{AGI}-native network and degrade the \ac{QoVE} of the metaverse user. In other words, this results in a miscommunication between the \ac{Tx} and the receivers. 
One possible interpretation of this issue lies in the underlying communication model, which typically treats individual communication links independently, overlooking the possibility that multiple receivers may have varying beliefs when interpreting the same message.
Here, we can potentially look at the use of the framework of \emph{mass communication theory}~\cite{mcquail2010mcquail} that considers more complex communication models with multiple receivers. One candidate from this theory is the Westley and Maclean communication model. This model considers multiple receivers which inherently have different beliefs or experiences that influence how communication messages are interpreted. In addition, the \ac{Tx} can also have its own beliefs. Clearly, this majorly agrees with the model of communication in an \ac{AGI}-native network with receivers having different causal models and knowledge, as in the example of holographic teleportation. In this case, the different beliefs can be incorporated into the communication model, facilitating a rigorous system design that reflects the underlying communication. Thus, it is crucial to design new metrics that can be optimized in this system to ensure the message is largely conveyed by the different receivers. For instance, this system can benefit from a collective channel capacity between the \ac{Tx} and the \ac{Rx} instead of separate communication capacities.

\vspace{-0.3cm}

\section{Conclusion and Recommendations}
In this paper, we have proposed a novel vision of \ac{AGI}-native networks that could constitute the next frontier of the wireless network evolution. This vision advocates for a paradigm shift from the traditional, unsustainable path of network evolution that is ultimately limited by various physical constraints of the wireless enablers, towards an \ac{AI}-driven path guided by next-generation \ac{AGI} technologies. In essence, we have shown that enabling next-generation of networks that can meet the perpetual demands of emerging applications like the metaverse, \acp{DT}, or holographic teleportation, will require equipping the network with cognitive abilities, namely, common sense. We have identified the pillars of common sense, and showcased how it can be implemented using a native cognitive telecom brain architecture having three essential modules: perception, world model, and action-planning, and complemented by proper memory and cost modules. 
This architecture enables future wireless networks to operate at \ac{AGI} levels and augments autonomous agents with common sense. Towards this end, we have presented a novel method for capturing generalizable abstract representations of the real-world through optimizing the complexity of these representations while balancing between causality and generalizability. Moreover, we have studied how transforming these representations into an \ac{HD} vector space facilitates building a world model that is compatible with the inuitive physics operations pertaining to common sense and the real, physical world. In addition, we have discussed how this world model can enable efficient analogical reasoning. Furthermore, we have proposed two action-planning methods, namely, intent-driven and objective-driven planning, that enable an \ac{AGI}-native network to plan its actions. 
In addition, we have shown how an \ac{AGI}-native network plays a crucial role in enabling a new set of use cases including cognitive avatars and holographic teleportation.

We finally conclude by offering concrete recommendations to pave the way for the emergence of \ac{AGI}-native wireless networks:

\begin{itemize}
    
    

    \vspace{-0.1cm}
    \item \textbf{Physics-based digital world models:} It is evident that world models bring forth common sense into \ac{AGI}-native wireless systems. Nevertheless, ensuring the proper foundations to implement the intuitive physics in world models is necessary. Hence, there is a need to push more towards designing physics-based world models and perfectly representing basic physical actions in such world.

    \item \textbf{Role of \acp{LLM} and generative \ac{AI} in an \ac{AGI}-native network}: \acp{LLM} and more broadly generative AI have become a major discussion point for current 6G efforts. In the proposed vision, LLMs and generative AI can have a few roles. For instance, there could be a need for translating human input into machine-understandable intents, and those inputs can be done through LLM prompts for instance. Those prompts are then translated into proper data points for our telecom brain. Moreover, generative AI can be used to generate synthetic data points out of the DT for potentially further training our system on additional inputs, thereby refining proactive learning. These generative capabilities can also be useful at the receiver in the context of a semantic communication network whereby the receiver can locally generate content in absence of the communication link. LLMs and generative AI can also have a role in human-centric applications, whereby they can be used in the design of foundation models for cognitive avatar applications. However, LLMs and generative AI, by themselves are clearly not the appropriate path for fully-fledged network automation and for instilling rigorous intelligence into wireless systems. 
    Instead, an AGI-native network, with the components defined specifically in Fig.~\ref{Telecom_Brain}, is the necessary path towards autonomy and AI-nativeness for wireless systems.
    In other words, LLMs are ``fluent'' AI systems that know a vast amount of information. However, \emph{fluent AI systems are not necessarily intelligent} as they lack cognitive functions like reasoning and common sense. In contrast, towards building the next-generation of AI-native networks, we need to put more emphasis on the design of AI systems that have strong reasoning,  planning, and common sense abilities. 
    Therefore, we recommend shifting focus from generative and large \ac{AI} models towards \ac{AI} designs that can rigorously understand  how the world works, and that can help further develop the building blocks in Fig.~\ref{Telecom_Brain}. In this regard, it is also necessary to design a more intelligent \ac{AI} that focuses more on the world and its physics, and less on the massive sizes of models and data.
    
    \item \textbf{Open interfaces in \ac{AGI}-native wireless systems:} 
    The design of the Open \ac{RAN} (ORAN) architecture is built on the premise of disaggregating software from hardware components and introducing the radio intelligent controller (RIC) functionality. Essentially, the openness of the \ac{RAN} interfaces plays a role in achieving this premise, and ORAN can be a suitable architecture for incorporating \ac{AGI}-native networks. Moreover, as \ac{AGI}-native systems anticipate the emergence of the telecom brain, the intelligence in the network can become fully disaggregated. Consequently, there is a need to adopt open interfaces beyond the \ac{RAN}, particularly, in the core network. For instance, a fully open network is crucial for the core-\ac{DT} and \ac{RAN}-\ac{DT} to interact with each other as well as with the elements of the world, so that the \ac{AGI}-native network can plan its actions. In addition, these open interfaces must enable the interoperability of \acp{DT} from different sources on the network. Hence, the design of a fully open network becomes a vital cornerstone for \ac{AGI}-native wireless systems. Here, we also note that the presence of an AGI-native network is also useful to orchestrate and manage the various functionalities of the existing Open RAN architecture. Indeed, deploying \ac{AGI}-native infrastructure over ORAN systems is an important subject for future investigation.

    \item \textbf{Synergies between fundamental mathematics and AGI}: In order to design \ac{AGI}-native wireless networks, it is necessary to exploit rigorous, fundamental tools from mathematics. Those tools include frameworks such as category theory, that is useful for transforming symbols in the telecom brain into \ac{HD} vectors, persistent homology that allows efficient grouping of similar symbols, as well as notions like integrated information, that allows quantifying the number of intent-driven planning steps. Here, we expect that building reliable and explainable \ac{AGI}-native wireless networks requires exploring synergies with fundamental mathematical frameworks in order to ``open'' the black-box, and be able to concretely define some of the \ac{AGI} and common sense abilities. Close collaboration between \ac{AI}, wireless, and mathematics researchers is therefore important. 

    \item \textbf{Design of efficient guardrails for autonomous AI:} As wireless systems continue to become autonomous, they will need to be guided throughout their learning, planning, and execution phases. Thus, this will require incorporating proper guardrails for \ac{AI} systems to remain safe and steerable with human intervention. Therefore, an indispensable step towards \ac{AGI}-native networks is to design efficient guardrails that can safely guide the network without constraining it from learning and acquiring knowledge. These guardrails should satisfy the guidelines posed by standardization bodies such as the Third Generation Partnership Project (3GPP) for wireless communications and networking applications. 
\end{itemize}
\vspace{-0.15cm}


\bibliographystyle{IEEEtran}
\def\baselinestretch{0.91}
\bibliography{bibliography}

\end{document}